\documentclass[nohyperref]{article}

\usepackage{microtype}
\usepackage{graphicx}
\usepackage{subfigure}
\usepackage{booktabs} 

\usepackage{hyperref}

\usepackage[accepted]{icml2022}

\usepackage{amsmath}
\usepackage{amssymb}
\usepackage{mathtools}
\usepackage{amsthm}

\usepackage{subfigure}
\usepackage{comment}
\usepackage{caption}
\usepackage{multirow}
\usepackage[multiple,bottom]{footmisc}
\usepackage[outline]{contour}
\usepackage{tablefootnote}

\usepackage{threeparttable}
\renewcommand{\algorithmiccomment}[1]{$\triangleright$ #1}
\usepackage[capitalize,noabbrev]{cleveref}

\theoremstyle{plain}

\theoremstyle{definition}

\theoremstyle{remark}

\usepackage[font=small,skip=0pt]{caption}
\usepackage[belowskip=0pt,aboveskip=0pt]{caption}

\usepackage[textsize=tiny]{todonotes}

\icmltitlerunning{Robust Contrastive Active Learning with Feature-guided Query Strategies}

\begin{document}
\twocolumn[
\icmltitle{Robust Contrastive Active Learning with Feature-guided Query Strategies}


\icmlsetsymbol{equal}{*}

\begin{icmlauthorlist}
\icmlauthor{Ranganath Krishnan}{ilo}
\icmlauthor{Nilesh Ahuja}{ilc}
\icmlauthor{Alok Sinha}{ici}
\icmlauthor{Mahesh Subedar}{ilo}
\icmlauthor{Omesh Tickoo}{ilo}
\icmlauthor{Ravi Iyer}{ilo}
\end{icmlauthorlist}

\icmlaffiliation{ilo}{Intel Labs, Oregon, USA}
\icmlaffiliation{ilc}{Intel Labs, California, USA}
\icmlaffiliation{ici}{Intel Corporation, Bangalore, India}

\icmlcorrespondingauthor{Ranganath Krishnan}{ranganath.krishnan@intel.com}

\icmlkeywords{Deep Active Learning, Contrastive Learning, Robustness}

\vskip 0.3in
]



\printAffiliationsAndNotice{}  

\begin{abstract}
We introduce supervised contrastive active learning (SCAL) and propose efficient query strategies in active learning based on the feature similarity (\textit{featuresim}) and principal component analysis based feature-reconstruction error (\textit{fre}) to select informative data samples with diverse feature representations. We demonstrate our proposed method achieves state-of-the-art accuracy, model calibration and reduces sampling bias in an active learning setup for balanced and imbalanced datasets on image classification tasks. We also evaluate robustness of model to distributional shift derived from different query strategies in active learning setting. Using extensive experiments, we show that our proposed approach outperforms high performing compute-intensive methods by a big margin resulting in 9.9\% lower mean corruption error, 7.2\% lower expected calibration error under dataset shift and 8.9\% higher AUROC for out-of-distribution detection. 
\end{abstract}

\section{Introduction}
\label{sec:introduction}

Supervised deep learning relies on a large amount of labeled data for training the models. Data annotation is often very expensive and time-consuming.  It is challenging to obtain labels for large-scale datasets in complex tasks such as medical diagnostics~\cite{irvin2019chexpert} that require specific expertise for data labeling, and semantic segmentation~\cite{cordts2016cityscapes} requiring pixel-wise labeling. Active learning~\cite{settles2009active,lewis1994sequential} enables cost-efficient labeling and is a promising solution that allows the model to choose the most informative data samples from which it can learn while requesting a human annotator to label the carefully selected data based on some query strategy. Deep active learning has been widely studied recently \cite{gal2017deep, shen2017deep, sener2018active, beluch2018power, ducoffe2018adversarial, yoo2019learning, kirsch2019batchbald},  but there remain many open research problems which we enumerate below. 

First, existing research in active learning has mainly focused on improving the model accuracy as samples are acquired, but accuracy alone is not indicative of the robustness of the trained model. It has been shown that once deep neural networks are trained, they face challenges in real-world conditions such as dataset shift and out-of-distribution data \cite{ovadia2019can, krishnan2020improving}. Dataset shifts \cite{quionero2009dataset} arise due to the non-stationary environments in the real-world as the observed data evolve from training data distribution and models can encounter novel scenarios~\cite{hendrycks17baseline}. Models should be well-calibrated and robust under such shifts in order to be deployed in safety-critical applications. This is particularly important in an active-learning setting since we need to ensure that the model remains robust despite being trained with fewer samples.  

Moreover, in real-world applications, the collected datasets often follow long-tailed distribution where the number of samples for different classes are highly imbalanced~\cite{liu2019large}. We show that models trained with existing active-learning methods show poor robustness when the datasets are highly imbalanced. Even when the datasets are balanced, active learning introduces sampling bias \cite{dasgupta2008hierarchical, dasgupta2011two, farquhar2021statistical} owing to the heuristic nature of sample selection. Sampling bias in deep neural network training can cause undesired behavior with respect to fairness, robustness and trustworthiness when deployed in real-world situations \cite{buolamwini2018gender,bhatt2020uncertainty}. 

We investigate and address these problems in this paper, and propose a simple yet effective active-learning method with novel query strategies leveraging the feature representation learnt through contrastive learning paradigm. Recent advancements in contrastive learning \cite{chen2020simple, chen2020big, he2020momentum, chen2020improved} have resulted in state-of-the-art performance in unsupervised representation learning. We extend the contrastive loss to active learning in a supervised setting to obtain well-clustered feature representations. We devise the query strategies that harness the properties of contrastive loss to select data samples with diverse features while maintaining a balanced representation of samples from each class in order to mitigate the sampling bias. 
Our motivation is to combine the benefits of uncertainty and diversity-based approaches to select most informative and diverse samples. Unlike other sample-selection methods such as CoreSet \cite{sener2018active} and Bayesian active learning by disagreement (BALD) \cite{gal2017deep} which are compute intensive resulting in very high query time at every iteration \cite{shui2020deep}, our method is fast and computationally inexpensive.

With these innovations -- efficient sample-selection strategy based on feature-similarity and principal component analysis (PCA) based feature-reconstruction error acquisition scores  harnessed with contrastive learning -- we show that our approach outperforms in terms of model robustness, model calibration, and accuracy with far fewer labeled samples than existing high-performing active learning methods including CoreSet~\cite{sener2018active}, Learning Loss~\cite{yoo2019learning} and BALD~\cite{gal2017deep} for balanced and imbalanced datasets on image classification tasks. To the best of our knowledge, this is the first work to evaluate the robustness of models to distributional shift derived from different query strategies in active learning setting. Also, the work presented in this paper is the first to leverage the supervised contrastive learning approach in active learning setup.




In summary, our \textbf{main contributions} include: 

\begin{itemize}
\item We propose two novel query strategies for active learning based on the feature-similarity (\textit{featuresim}) and PCA-based feature-reconstruction error (\textit{fre}) scoring functions harnessed with supervised contrastive active learning (SCAL), which we build upon sound contrastive learning scheme~\cite{khosla2020supervised}.
\item We evaluate the model calibration and robustness in active learning setup, demonstrating our method yields well-calibrated models and clearly outperforms existing methods in robustness to dataset shift and out-of-distribution data.
\item We demonstrate that the proposed method is computationally efficient in selecting diverse and informative data samples in active learning, reduces the sampling bias and improves active learning performance in both balanced and long-tailed imbalanced dataset scenarios.

\end{itemize}

\section{Background and Problem setup}
\label{sec:background}

Active learning aims to learn from a small set of informative data samples, which are acquired from a huge unlabeled dataset, thus minimizing the data annotation cost.
The acquired data samples are labeled by an oracle (e.g. human annotator), which are used for training the model. In this framework, the models are allowed to select the data from which they can learn based on a query strategy. 
The query strategy evaluates the informativeness of data samples; some of the commonly used strategies include the  uncertainty-based~\cite{lewis1994sequential, tong2001support} and diversity-based~\cite{brinker2003incorporating} approaches.
For example, the data samples with higher uncertainty estimates are considered to be most useful in uncertainty-based query strategy. We refer to~\cite{settles2009active} for an overview of earlier works in active learning and~\cite{ren2021survey} for survey of recent deep active learning methods. 
\begin{figure}[t]
\begin{center}
\centerline{\includegraphics[width=\columnwidth]{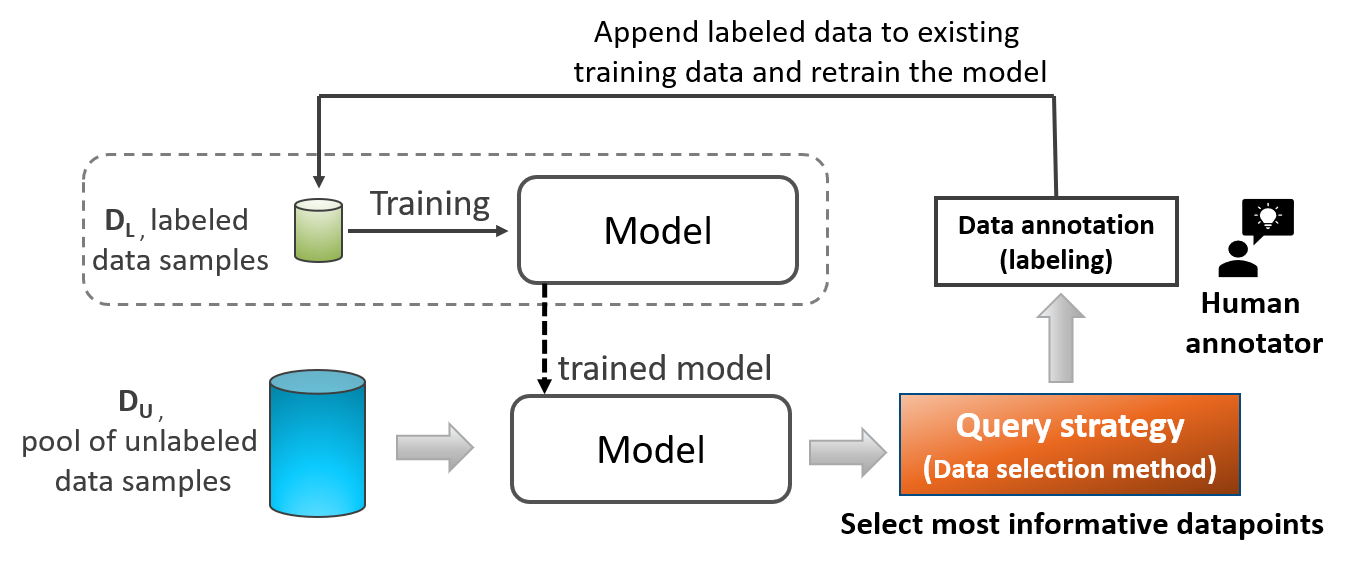}}
\caption{Active Learning setup}
\label{fig:al}
\end{center}
\end{figure}
Contrastive learning~\cite{chen2020simple,chen2020big,he2020momentum,chen2020improved} requires positive and negative samples in a mini-batch to bring the positive samples closer in the feature space while pushing the negative samples further apart. In the self-supervised setting, positive samples are selected by applying data augmentation and the rest of the samples are assumed to be negative examples. 
~\citet{khosla2020supervised} propose a generalization to contrastive loss by extending to a fully-supervised learning setting by using the label information. \citet{tack2020csi} used contrastive learning for novelty detection. 
Relying on the theoretical guarantees for robust performance of the learned representations \cite{saunshi2019theoretical,wang2020understanding,graf2021dissecting,tosh2021contrastive} with contrastive learning, we intuitively expect to select informative samples for active learning from the carefully designed query strategy based on feature representations. 

\subsection{Problem formulation:}
\label{sec:problem_setup}
Let $f_\theta(\cdot)$ represent a deep neural network with model parameters $\theta$ to be trained for a multi-class classification problem with a limited and fixed data labeling budget $B$.
Let $\mathcal{D}_{\mathrm{U}}=\left\{{x_{u}}\right\}_{u=1}^{N}$ represent an initial large pool of unlabeled data. Initially a small set of M samples are randomly sampled from $\mathcal{D}_{\mathrm{U}}$ and annotated by an oracle (human annotator) to obtain initial labeled set $\mathcal{D}_\mathrm{L}^{1}=\left\{\left({x_\ell}, {y_\ell}\right)\right\}_{\ell=1}^{M}$, where 
${y_{\ell}} \in \small{\{c_k}\}_{k=1}^{K}$ is the ground-truth class label with K-classes. 
The labeled M samples are removed from the unlabeled pool and model is trained with $\mathcal{D}_\mathrm{L}^{1}$. The next batch of $\mathrm{M}$ samples for the next training iteration of the active-learning are chosen from the remaining unlabeled data $\mathcal{D}_\mathrm{U}^1 = \mathcal{D}_\mathrm{U} \setminus \mathcal{D}_\mathrm{L}^{1}$. These are chosen based on an acquisition function $\mathcal{Q}(\mathcal{D}, f_{\theta}, \mathcal{S})$ that evaluates the samples from dataset $\mathcal{D}$ on the network $f_{\theta}$ and determines the indices $I_\mathrm{M}$ of the M samples that yield the best scores on a scoring function $\mathcal{S}$. $\mathcal{S}$ depends on the particular method used and typically indicates the informativeness of a sample to be acquired to train $f_{\theta}$. The query function $\mathcal{Q}$ returns, therefore, the most informative samples for the next active learning iteration. These samples are annotated by an oracle and added to existing labeled set $\mathcal{D}_{\mathrm{L}}^1$ to create $\mathcal{D}_{\mathrm{L}}^2=\mathcal{D}_{\mathrm{L}}^1 \cup \left\{\left({x_\ell}, {y_\ell}\right)\right\}_{\ell \in I_\mathrm{M}}$, and simultaneously are removed from the unlabeled set to obtain $\mathcal{D}_\mathrm{U}^2 = \mathcal{D}_\mathrm{U} \setminus \mathcal{D}_{\mathrm{L}}^2$. Thus, a sequence of [$\mathcal{D}_{\mathrm{L}}^1, \mathcal{D}_{\mathrm{L}}^2, \mathcal{D}_{\mathrm{L}}^3, ....$] labeled sets of size [M, 2M, 3M, ....] samples and corresponding trained models [$f_{\theta}^{1}, f_{\theta}^{2}, f_{\theta}^{3}, .... $] are obtained from every iteration of active learning. This cycle is repeated for $T$ iterations, each time acquiring $M$ samples for labeling until the query budget is reached as shown in Fig.~\ref{fig:al}.

\section{Proposed Method}
\label{sec:methods}
In this section, we present our proposed method supervised contrastive active learning (\textbf{SCAL}) with the query strategies based on the \textit{feature-similarity} (\textbf{\textit{featuresim}}) and \textit{feature-reconstruction error} (\textbf{\textit{fre}}) scores. We follow the problem formulation and notations described in Section~\ref{sec:problem_setup}.
Suppose a batch of $M$ samples has to be selected from the unlabeled set as described in Section \ref{sec:problem_setup}. Ideally, we would like to choose an equal number of samples from each of the $K$ classes, yet with diverse feature representations. Since true class-labels, $y$ are not available for the samples in the unlabeled pool, we use the predicted class labels, $\hat{y}$. If  $\hat{y}=c_k$, then we calculate a score between the observed feature and cluster of previously labeled features from class $k$ only. This allows us to select $M/K$ samples per predicted class which leads to a balanced sample selection, as demonstrated empirically in subsection \ref{sec:results}.
\subsection{Supervised Contrastive Active Learning}
\label{subsec:scal}
We extend the contrastive loss proposed in~\cite{khosla2020supervised} to active learning in a supervised setting. At every active learning iteration, the model is trained with newly acquired labeled data using the loss function given by Equation~\eqref{eqn:supconloss}.
\begin{equation}
    \begin{small}
    \label{eqn:supconloss}
    \mathcal{L}_{con}=\sum_{i \in I} \frac{-1}{|P(i)|} \sum_{p \in P(i)} \log \frac{\exp \left({z}_{i} \cdot {z}_{p} / \lambda \right)}{\sum_{\mathrm{n} \in \eta(i)} \exp \left({z}_{i} \cdot {z}_{\mathrm{n}} / \lambda \right)}
    \end{small}
\end{equation}
${z}$ is the output features from projection head of the neural network, $\lambda$ is scalar temperature parameter, $i \in I$ is the index of sample in augmented batch, $P(i)$ is the set of all positives (similar examples from the same class) corresponding to the index $i$ and $|P(i)|$ is its cardinality and $\eta(i) \in I \setminus \{i\}$. 

\begin{algorithm}[t]
\begin{small}
  \caption{SCAL (\textit{featuresim}) pseudocode}
  \label{alg:scal_featuresim}
   
  \textbf{Input:} Neural network model $f_\theta(x)$, unlabelled data $\mathcal{D}_{\mathrm{U}}$, number of iterations $T=B/$M where, $B$ is query budget and M is sample acquisition size per iteration).
\begin{algorithmic}[1]
  \STATE \textbf{Initialize:} $\mathcal{D}_L^0 \leftarrow \phi$, $Z_L \leftarrow \phi$
  \STATE Query labels for small subset of M samples drawn uniformly at random from $\mathcal{D}_{\mathrm{U}}$ to get initial labeled dataset, $\mathcal{D}_\mathrm{L}^{1}$.
  \STATE Train model $f_{\theta}^{1}$ on $\mathcal{D}_\mathrm{L}^{1}$ by minimizing $\mathcal{L}_{con}$ 
  \FOR{$t=1,2,...,T$:}
      \STATE $\triangleright$ Get features from the model $f_\theta^{t}$
      \STATE $Z_L^t \leftarrow \{z|z \leftarrow f_\theta^{t}(x)$, $\forall x \in \mathcal{D}_\mathrm{L}^{t} \setminus \mathcal{D}_\mathrm{L}^{t-1}\}$
      \STATE $Z_L \leftarrow Z_L \cup Z_L^t$
      \STATE $Z_U \leftarrow \{z|z \leftarrow f_\theta^{t}(x)$, $\forall x \in \mathcal{D}_\mathrm{U} \setminus \mathcal{D}_\mathrm{L}^{t}\}$      
      \FOR[Iterate over all classes]{$k \in \{1, \dots, K\}$:}  
      \STATE $S_k \leftarrow \{\mathcal{S}_{featuresim}(x,k)|x\in \mathcal{D}_\mathrm{U} \setminus \mathcal{D}_\mathrm{L}^{t}\}$
      \STATE $\triangleright$ Select indices of $M/K$ samples with minimum $S_k$
      \STATE $I_k \leftarrow \{\text{argsort}(S_k)\}_{1\dots{M/K}}$
      \ENDFOR
  \STATE $I_M \leftarrow \bigcup_{k=1}^{K} I_k$
  \STATE $\mathcal{D}_\mathrm{L}^{t+1} \leftarrow \mathcal{D}_\mathrm{L}^{t} \cup \left\{\left({x_\ell}, {y_\ell}\right)\right\}_{\ell \in I_\mathrm{M}} $
  \STATE Train a model $f_\theta^{t+1}$ on $\mathcal{D}_\mathrm{L}^{t+1}$ by minimizing $\mathcal{L}_{con}$ 
  \ENDFOR
  \STATE \textbf{return} final model $f_\theta^{T+1}$
\end{algorithmic}
\end{small}
\end{algorithm}

\begin{figure}[t]
	\small
	\begin{subfigure}
		\centering
		\captionsetup{
			justification=centering}
		\includegraphics[scale=0.4]{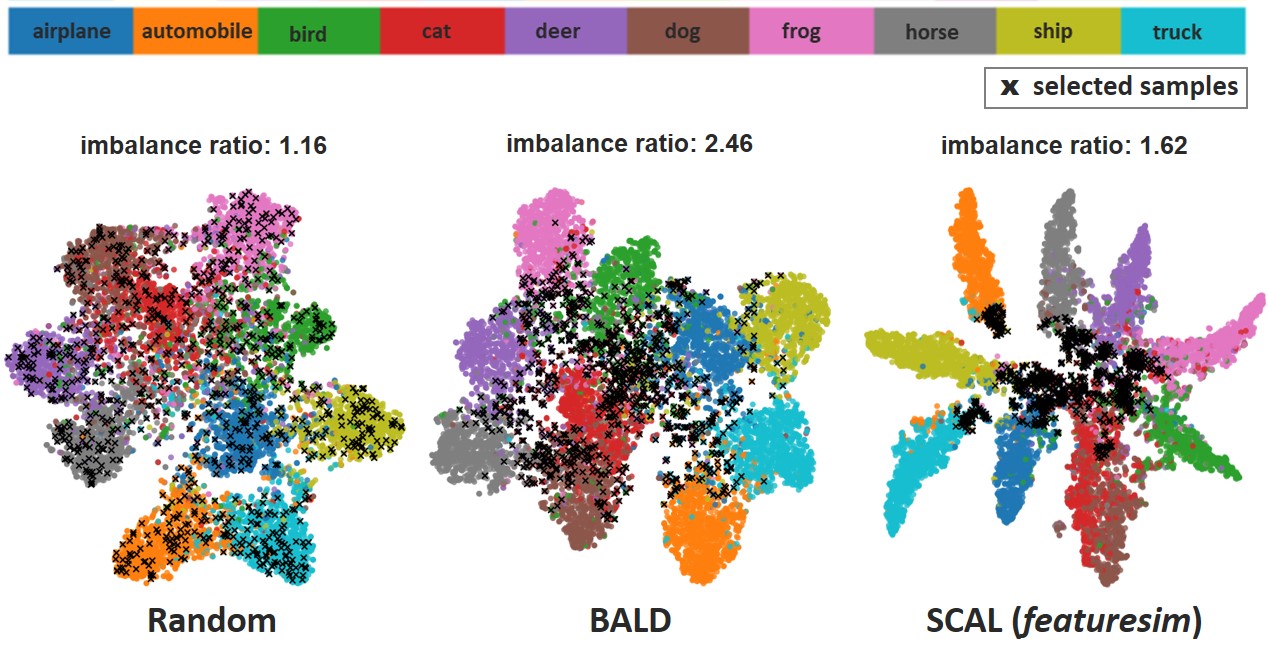}
	\end{subfigure}
	\caption{\small tSNE embedding of CIFAR10/ResNet-18 showing selected samples with different query methods at fourth iteration in active learning. Our proposed SCAL method selects balanced, diverse and informative samples (samples in-between clusters and from edge of clusters) from each class. Imbalance ratio indicates ratio of samples from most frequent and least frequent class.}
	\label{fig:tsneplot}
\end{figure}

The guiding principle of contrastive loss is to pull together clusters of samples from the same class (positives) and push apart the clusters of distinct classes (negatives) in the feature embedding space (tSNE visualization shown in Fig.~\ref{fig:tsneplot}), with theoretical guarantees for optimal learned feature representations~\cite{saunshi2019theoretical,graf2021dissecting,tosh2021contrastive}. As labels are acquired in an active learning setting at each iteration, the model training can benefit from accurately drawing the negatives for the anchor from different classes and drawing the positives from the same class using label information, not just relying on the data augmentations of the anchor. This helps supervised contrastive loss to result in more robust clustering of the feature representation~\cite{khosla2020supervised}. Within such a scheme, the use of simple and inexpensive distance-based scores in the feature space suffices for sample-selection without the need for expensive algorithms such as k-Center-Greedy used in Coreset. We take advantage of this class-based clustering to propose a strategy for selecting a set of diverse, unbiased and informative samples from the unlabeled pool. Specifically, we propose two simple and inexpensive scoring functions: a feature-similarity (\textit{featuresim}) score, which measures the similarity between a feature and a cluster, and a PCA based feature-reconstruction error (\textit{fre}) score, which measures the distance of a feature from a cluster. Both of these query strategies are described next. We refer to this method as supervised contrastive active learning (SCAL).

\subsubsection{\textbf{SCAL (\textit{featuresim})}}
Contrastive loss maximizes the cosine similarity between similar samples and minimize the similarity of distinct feature vectors. 
The \textit{featuresim} score leverages this property by computing cosine similarity between the sample from unlabeled pool and corresponding training features embedding belonging to the same class as the prediction, weighted by the norm of the feature vector $\|{z}\|$.
The query function $\mathcal{Q}$ is intended to obtain $M/K$ samples from each cluster in feature embedding space based on the feature similarity score as given by Eq. \eqref{eqn:sim_score}. Here, $l$ is an index over the labeled set and we only consider those samples for which the true label $y_l=\hat{y}$, where $\hat{y}=c_k$ is the predicted label for the input ${x}$.
\begin{equation}
\begin{aligned}
\label{eqn:sim_score}
   \mathcal{S}_{featuresim}({x}, k) := \max _{\ell} \frac{z\left({x}_{\ell} |  {y}_{\ell}=c_{k}\right)}{ \|z\left({x}_{\ell} |  {y}_{\ell}=c_{k}\right)\|} \cdot z({x} | \hat{y}=c_{k}) \\
\end{aligned}
\end{equation}

We refer to this as SCAL (\textit{featuresim}), where the query function selects the samples with the least \textit{featuresim} scores (distinct samples from the currently labeled data in feature embedding space) at each active learning iteration. 

\subsubsection{\textbf{SCAL (\textit{fre})}}
\label{subsec:dfm}
The other score that we propose for measuring distance between an observed feature, $z$, and the cluster of features from a specific class from the labeled dataset, $\mathcal{D}_L^t$, is the \emph{feature reconstruction error}. To calculate this, we simplify the approach proposed in \citet{ahuja2019probabilistic} for detecting out-of-distribution samples, which involved modeling class-conditional probability distributions to the deep-features, $z$, of a DNN. Prior to learning the distribution, they reduced the dimension of the feature space by applying a set of class-conditional PCA (principal component analysis) transforms, $\left\{\mathcal{T}_k\right\}_{k=1}^K$. In our application, this means estimating $\mathcal{T}_k$ from the subset of the labeled data that belongs to class $c_k$, i.e. from $\{(x,y)\in \mathcal{D}_L^t | y=c_k\}$.
To calculate \textit{fre}, therefore, we first select the transform, $\mathcal{T}_k$, that corresponds to the predicted class label $\hat{y}=c_k$ for the current input $x$, and  then calculate the norm of the difference between the original feature vector and the pre-image of its reduced embedding:

\begin{figure*}[t]
		\centering     
		\subfigure[Imbalanced-CIFAR10]{\includegraphics[width=0.64\columnwidth]{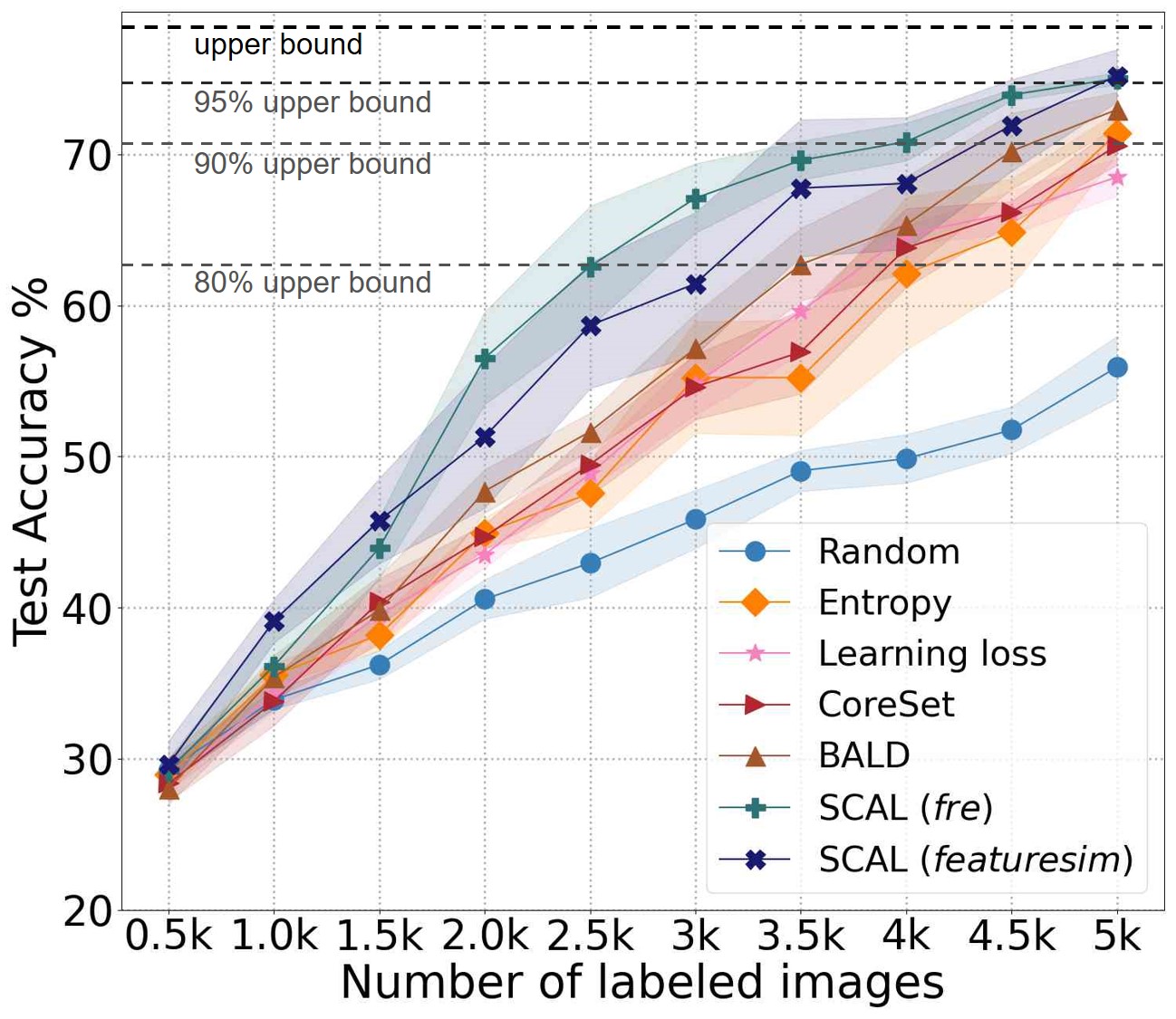}}
		\subfigure[CIFAR10]{\includegraphics[width=0.64\columnwidth]{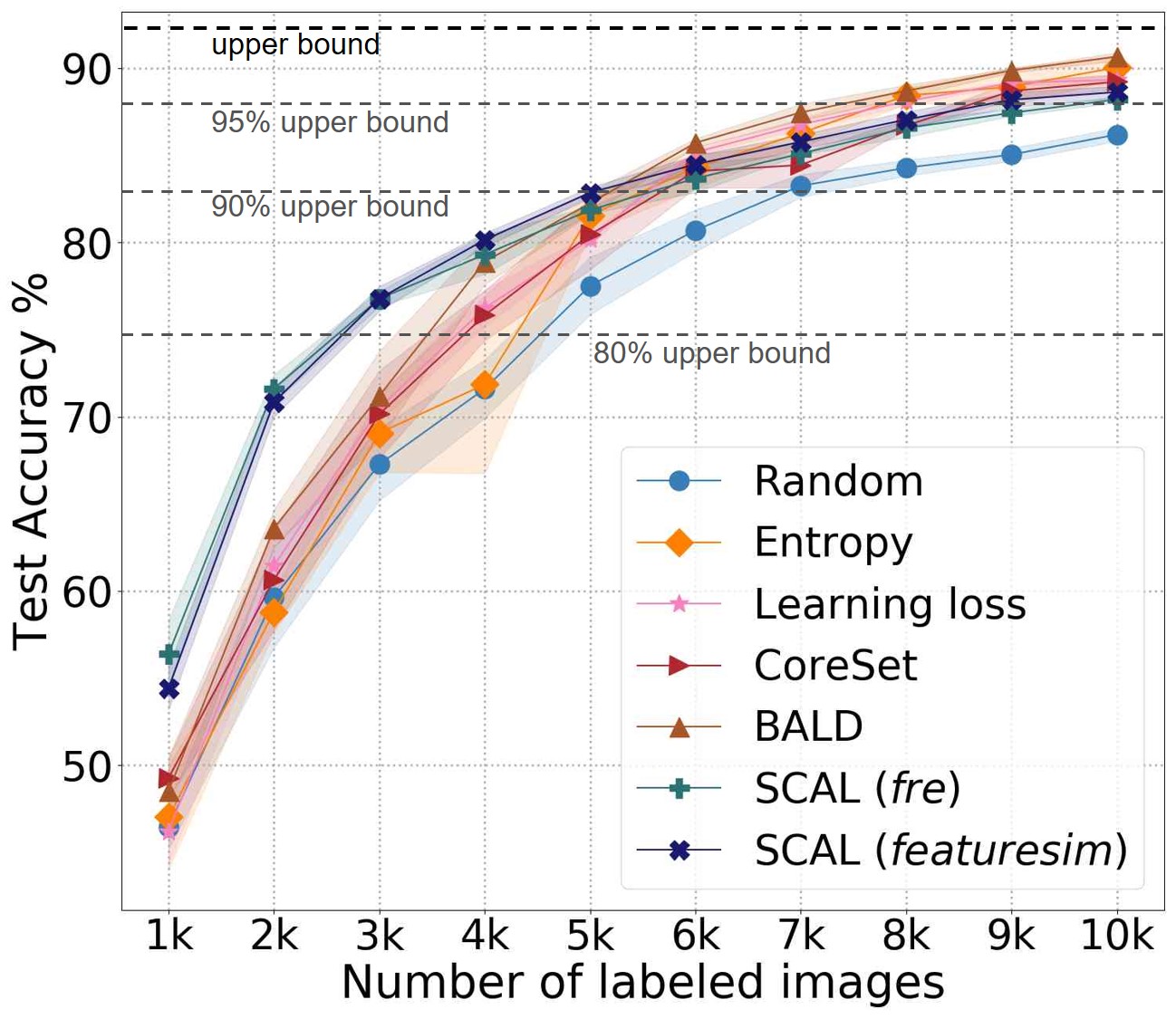}}
		\subfigure[SVHN]{\includegraphics[width=0.64\columnwidth]{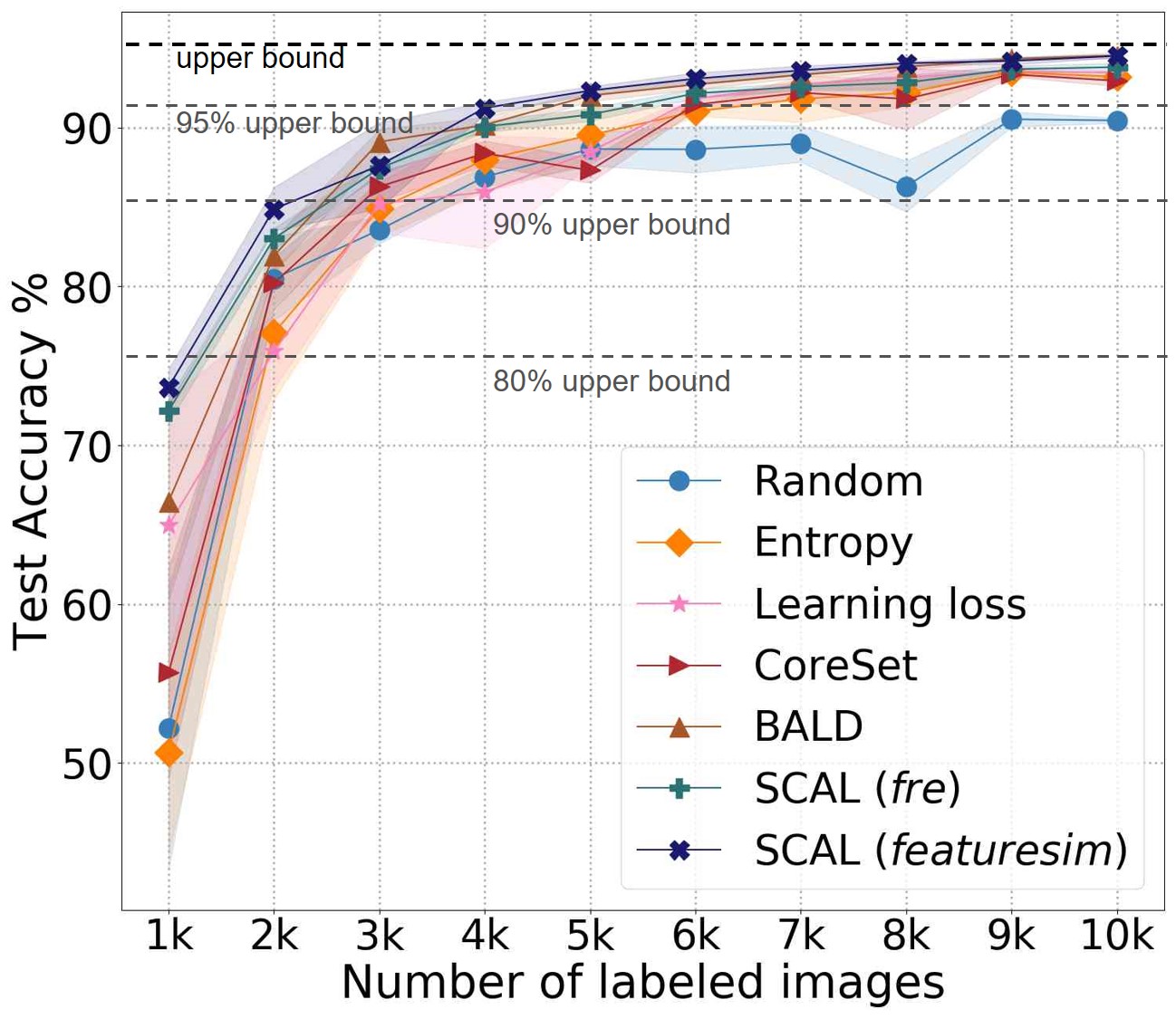}}
		\caption{\footnotesize Test Accuracy$\uparrow$ as a function of acquired data samples with different query methods in active learning for Imbalanced-CIFAR10, CIFAR10 and SVHN datasets. The shading for this and subsequent plots show std-dev from 5 independent trials for each method. SCAL achieves accuracy comparable to other methods for balanced datasets (CIFAR10) and clearly outperforms on imbalanced datasets. The upper bound indicates the maximum accuracy achieved when model is trained with entire dataset.}
\label{fig:acc}		
\end{figure*}

\begin{equation}
\label{eqn:fre_score}
    \mathcal{S}_{fre}({x}, k) := \|z({x} | \hat{y}=c_{k})-(\mathcal{T}_k^{\dagger} \circ \mathcal{T}_k) z({x} | \hat{y}=c_{k})\|
\end{equation}

Here, $\mathcal{T}_k$, is the forward PCA transformation, and $\mathcal{T}_k^{\dagger}$ is its Moore-Penrose pseudo-inverse \cite{golub1996matrix}.
Intuitively, query strategy select samples that are most distant from the feature space of the currently labeled data. 

Both \textit{featuresim} and \textit{fre} are simple yet effective query strategies for sample selection in active learning. While \textit{fre} method does not require storage of all the labelled features, \textit{featuresim} does not require the recomputation of the PCA transforms at each active learning iteration.

\section{Experiments and Results}
\label{sec:experiments}
\subsection{Experimental setup}
We perform a thorough empirical evaluation on image classification tasks with balanced and imbalanced datasets. 


We use CIFAR-10~\cite{krizhevsky2009learning} for balanced, long-tailed Imbalanced-CIFAR10~\cite{cao2019learning} and Street View House Numbers (SVHN)~\cite{netzer2011reading} for imbalanced setup (dataset details are provided in Appendix~\ref{appdx:experiments}). 
We use CIFAR10-C~\cite{hendrycks2019benchmarking} to evaluate robustness of models derived from different query methods to distributional shift. CIFAR10-C comprises 80 variations of dataset shift resulting from 16 different types of image corruptions and perturbations at 5 different levels of intensities for each dataset shift type.

We compare our proposed SCAL(\textit{featuresim}) and SCAL(\textit{fre}) with the state-of-the-art methods including Learning loss~\cite{yoo2019learning}, CoreSet~\cite{sener2018active} and Bayesian active learning by disagreement (BALD)~\cite{gal2017deep} along with baseline Random and Entropy query methods. These query strategies are described in Appendix~\ref{appdx:query}.

We use ResNet-18~\cite{he2016deep} model architecture for all the methods and datasets under study. We use the same hyperparameters for all the models for a fair comparison. The implementation details and the  hyperparameters are provided in Appendix~\ref{appdx:hyperparam} and \ref{appdx:implementaion}.

As in a typical active learning setup, we assume there are no labels available initially in the training set. The initial unlabeled set $\mathcal{D}_{\mathrm{U}}$ has 50K samples for CIFAR10, 14K samples for Imbalanced-CIFAR10 and 73.2K samples for SVHN. We set sample acquisition size $\mathrm{M}=1000$ for CIFAR10 and SVHN datasets, and $M=500$ for Imbalanced-CIFAR10. As described in Section \ref{sec:problem_setup}, the models are trained iteratively for $T=10$ iterations with a sequence of labeled data $\left[\mathcal{D}_{\mathrm{L}}^{1}, \mathcal{D}_{\mathrm{L}}^{2}, \dots, \right]$ that are acquired through the query strategy from respective methods. 
After every iteration, we evaluate the models with independent labeled test samples. For each method, results from 5 independent trials is presented.

\begin{figure*}[h]
		\centering     
		\subfigure[Imbalanced-CIFAR10]{\includegraphics[width=0.64\columnwidth]{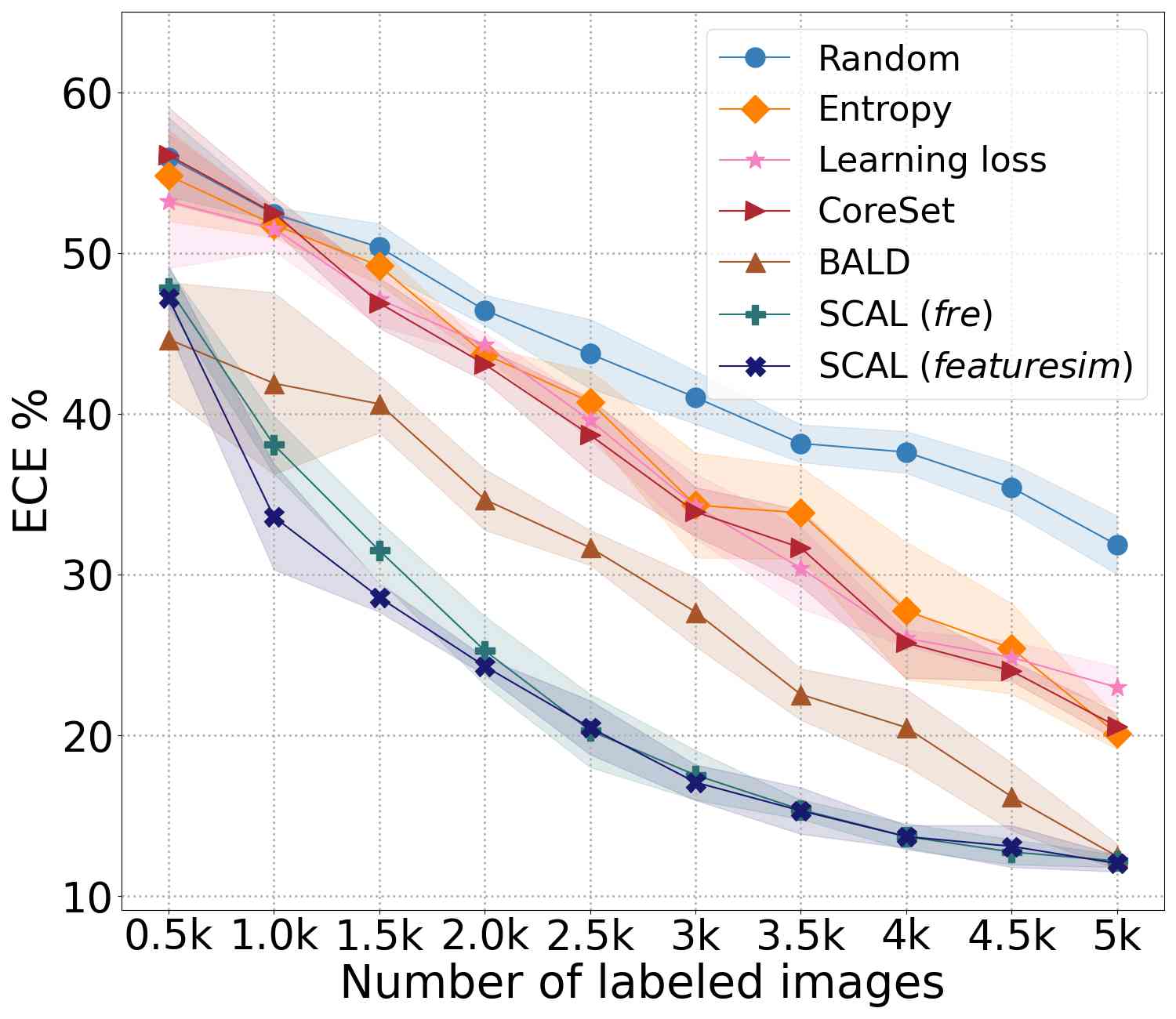}}
		\subfigure[CIFAR10]{\includegraphics[width=0.64\columnwidth]{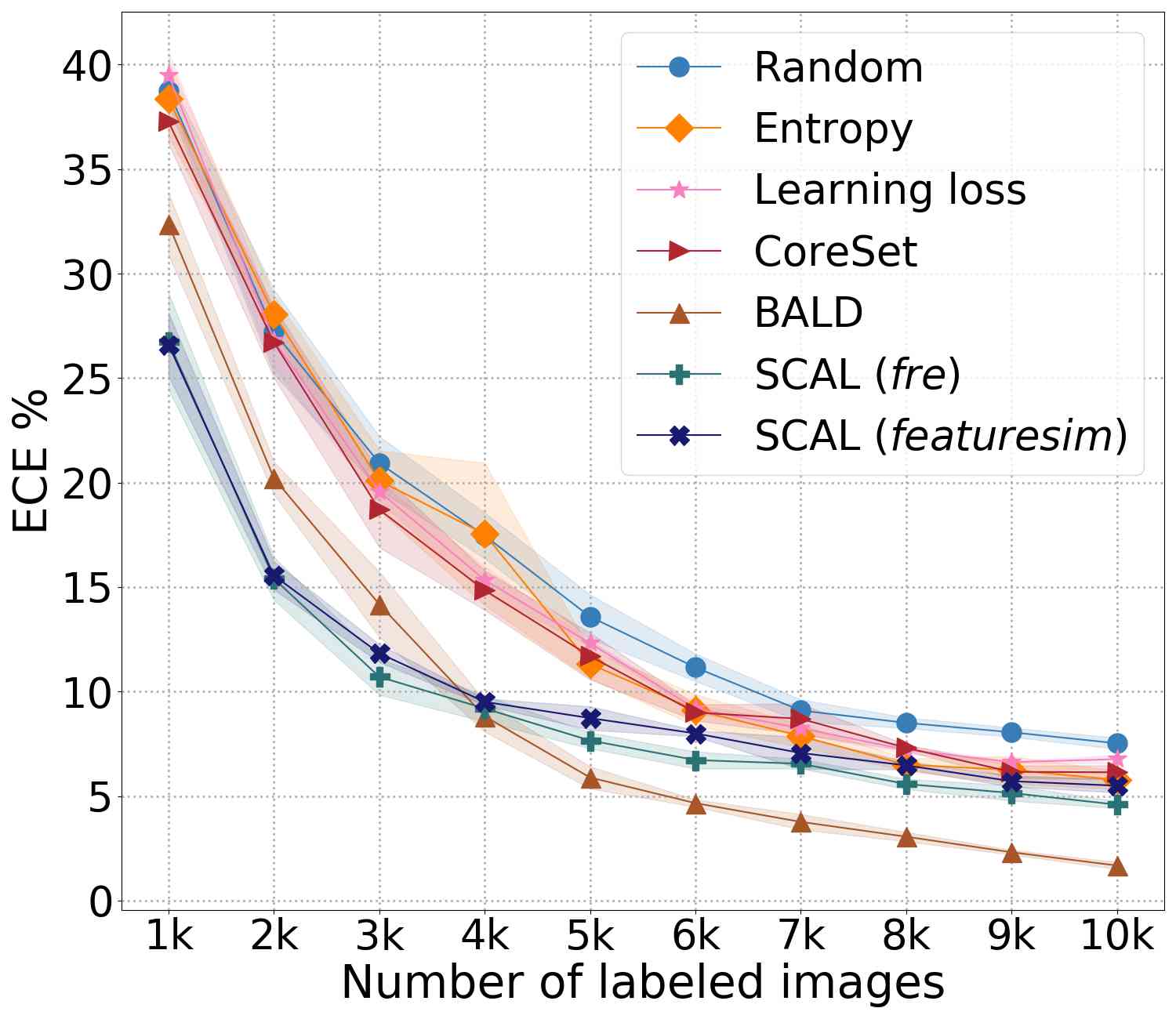}}
		\subfigure[SVHN]{\includegraphics[width=0.64\columnwidth]{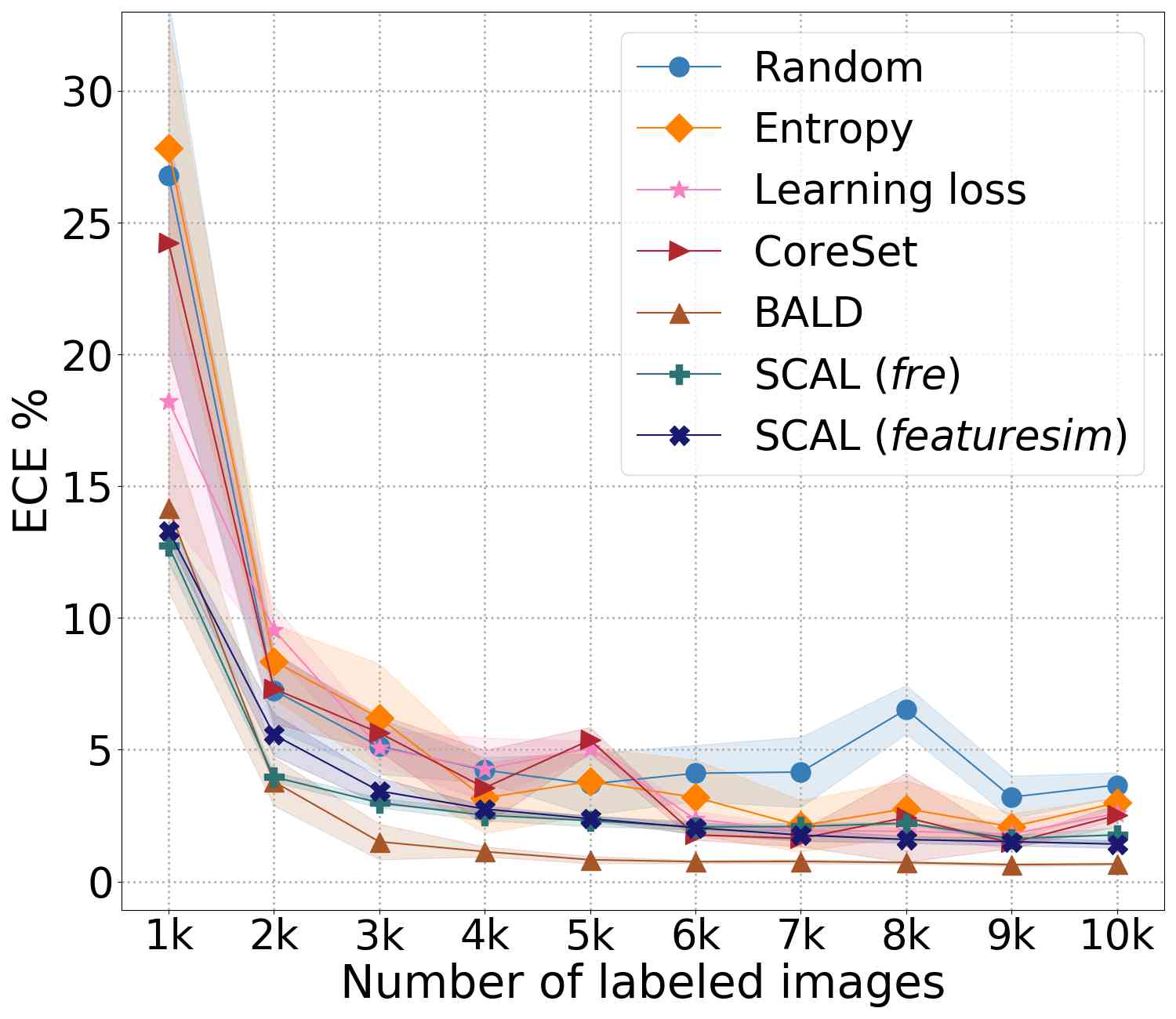}}
		\caption{\small \footnotesize Expected Calibration Error (ECE)$\downarrow$ as a function of acquired data samples with different query methods on three different datasets. Lower ECE is better, indicating the model is well-calibrated. SCAL yields lowest ECE on long-tailed Imbalanced-CIFAR-10 while BALD yields the lowest ECE for balanced datasets, but it is computationally expensive as noted in Table\eqref{tab:query-table}.}
\label{fig:ece}		
\end{figure*}

\begin{figure*}[t]
		\centering     
		\subfigure[Dataset shift: mCE {$\downarrow$}]{\includegraphics[width=0.64\columnwidth]{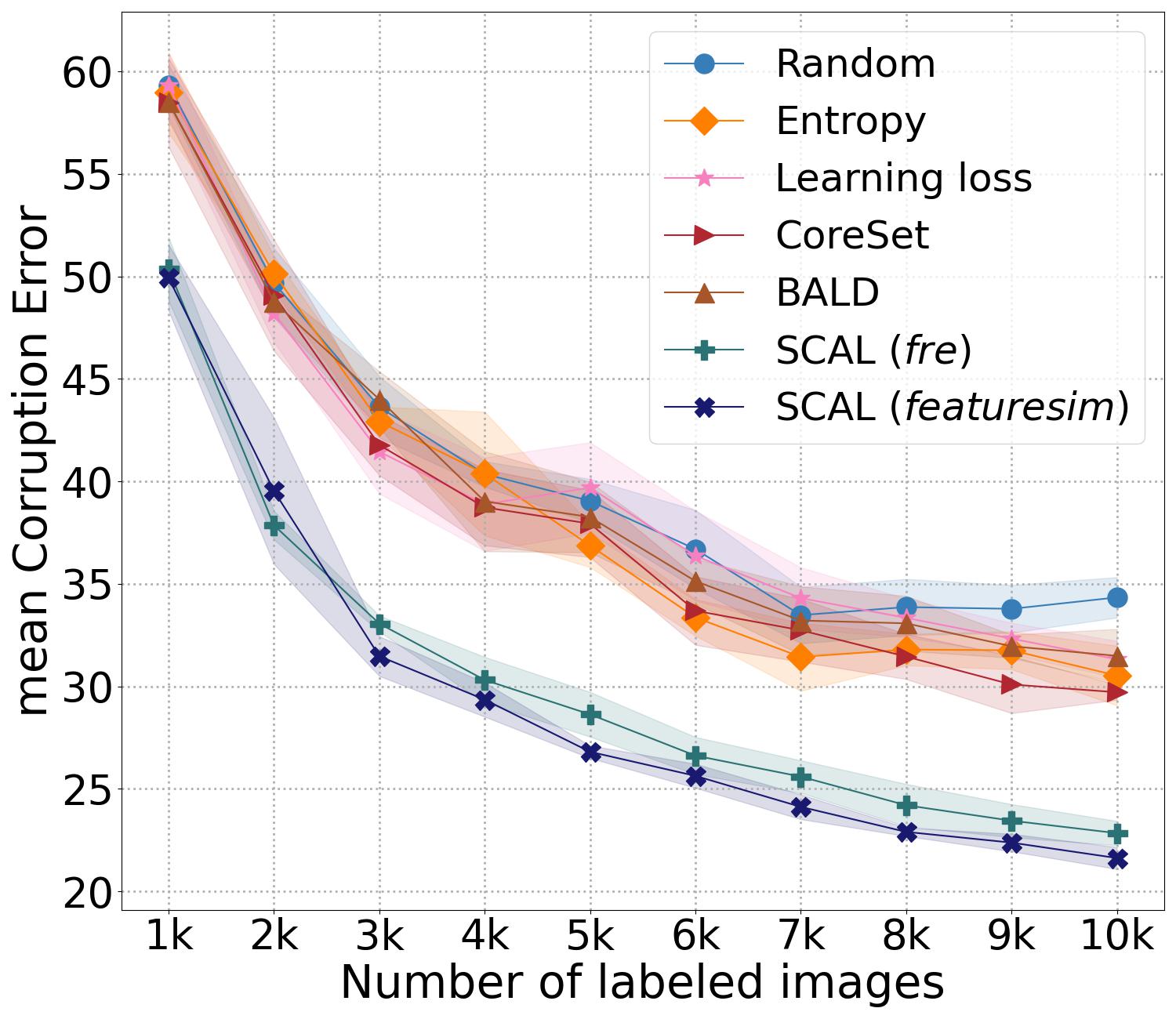}}
		\subfigure[Dataset shift: ECE {$\downarrow$}]{\includegraphics[width=0.64\columnwidth]{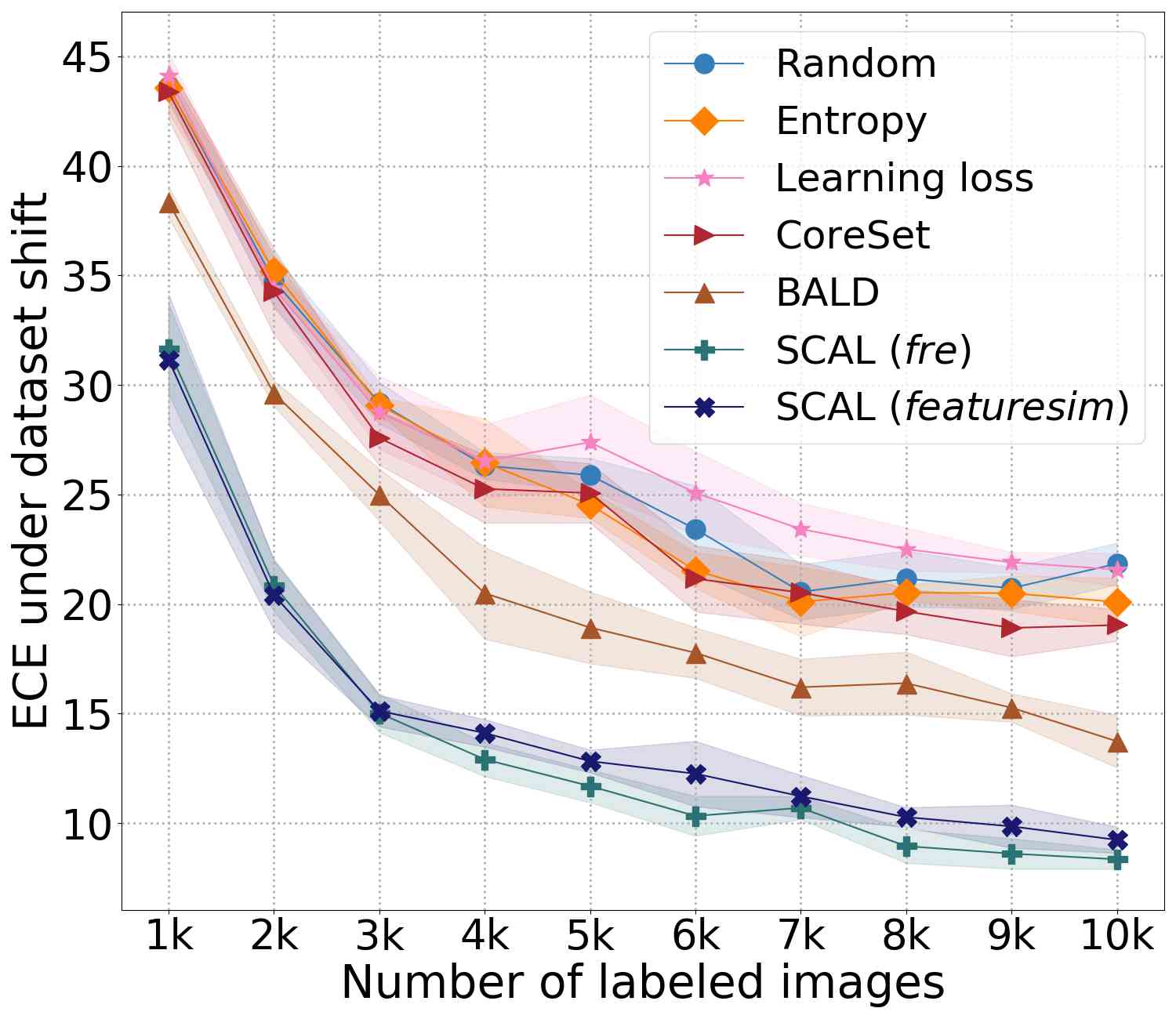}}
		\subfigure[ out-of-distribution: AUROC {$\uparrow$} ]{\includegraphics[width=0.64\columnwidth]{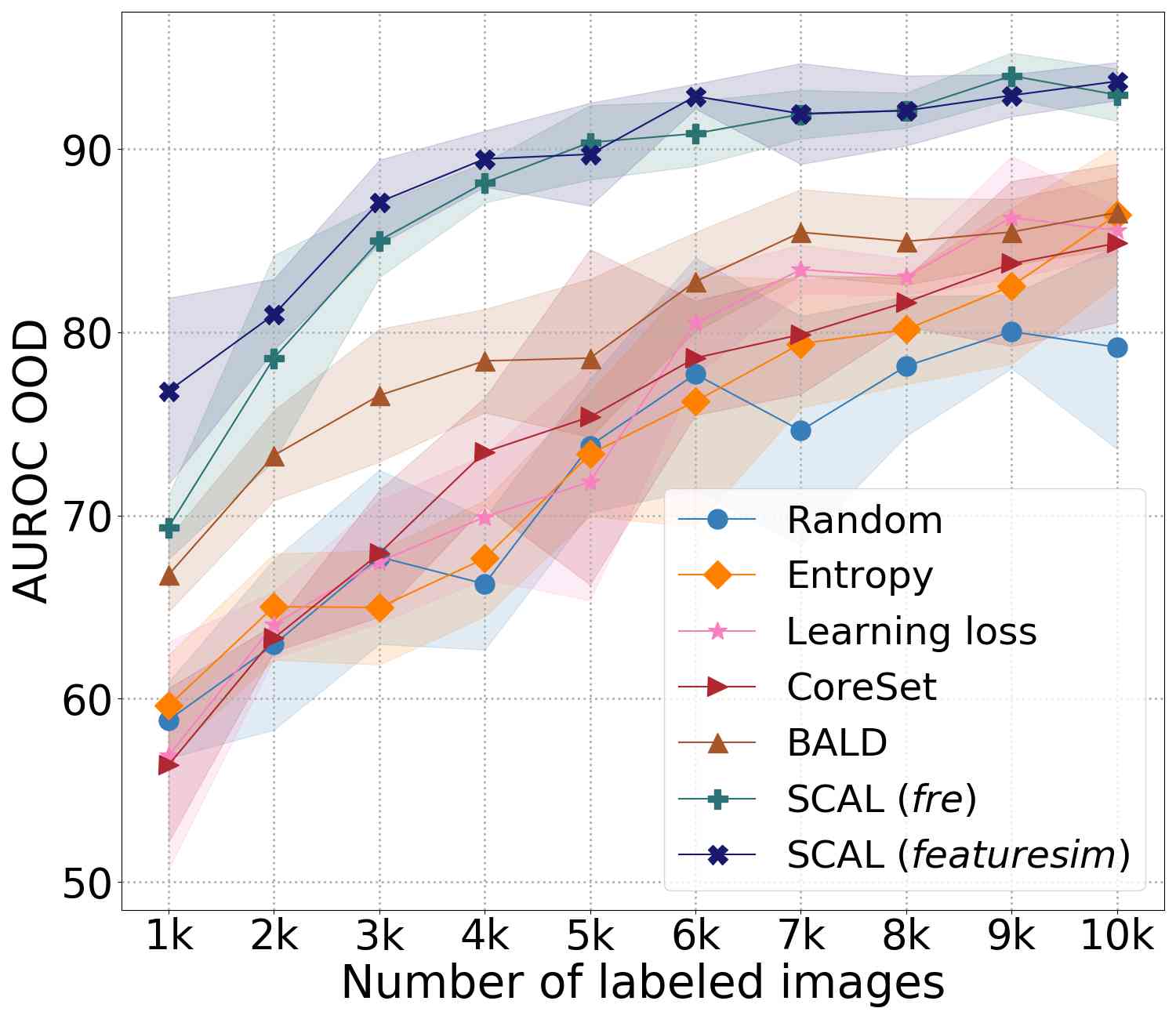}}
		\caption{\small \footnotesize Robustness to  dataset shift (CIFAR10-C with 16 different shift types at 5 different corruption intensity levels) and out-of-distribution (OOD) data (in-distribution: CIFAR-10, OOD: SVHN). (a) mean corruption error: SCAL is more robust to corruptions (dataset shift) due to better learnt feature representation as shown in tSNE embedding in Fig.~\ref{fig:tsneshiftplot}. (b) ECE on corrupted data: SCAL yield reliable confidence under dataset shift as reflected by lower ECE. (c) SCAL outperforms in OOD detection with higher AUROC. 
		}
\label{fig:robustness}
\end{figure*}

\subsection{Evaluation metrics}
Existing works in active learning have mainly focused on evaluating model accuracy as trained from fewer data samples, but accuracy alone is not indicative of the model performance. Models need to provide reliable and calibrated confidence measures in addition to providing accurate predictions. Further models need to be robust to distributional shift as the observed data may shift from the training data distribution in real-world, this is particularly important in an active-learning setting since model need to be robust despite being trained with fewer samples.
We evaluate all these aspects on model performance, calibration and robustness in active learning setup with various widely adopted evaluation metrics{\footnote[1]{Arrows next to each metric indicate lower($\downarrow$) or higher($\uparrow$) value is better.}} as listed below:
\begin{itemize}
\item Model performance and efficiency of query strategy are evaluated using test accuracy($\uparrow$), sampling bias($\downarrow$) and query time($\downarrow$).
\item Model calibration and robustness to dataset shift is evaluated using expected calibration error (ECE)($\downarrow$)~\cite{naeini2015obtaining}, mean corruption error (mCE)($\downarrow$)~\cite{hendrycks2019benchmarking} and proper scoring rules including negative log-likelihood (NLL)($\downarrow$) and Brier score($\downarrow$)~\cite{brier1950verification}. We study the robustness to dataset shift using CIFAR10 perturbed with 16 different shift types at 5 different intensity levels for each datashift type (such as Gaussian blur, brightness, contrast, etc.)~\cite{hendrycks2019benchmarking}.
\item Robustness to out-of-distribution (OOD) data is evaluated using \textit{area under the receiver operating characteristic curve} (AUROC)($\uparrow$)~\cite{davis2006relationship}
\end{itemize}

We propose the following sampling bias score which measures the class imbalance in the acquired data:
\begin{equation}
\label{eq:bias}
\text{Sampling Bias} = 1-\frac{\mathcal{H}_{D_L}}{\mathcal{H}_{balanced}}
\end{equation}
Here, $\mathcal{H}_{D_L}$ is the entropy of the sample distribution over the labeled dataset defined as:
$\mathcal{H}_{D_L} = - \sum_{k=1}^K \left({M_k}/{M}\right) \log \left({M_k}/{M}\right)$,
where, $M_k$ is the number of samples from class $c_k$, and $M=\sum_k M_k$ is the total number of samples. $\mathcal{H}_{balanced}$ is the entropy of a balanced sample distribution for which all $M_k$ are equal. 


\begin{table}
\begin{threeparttable}
\caption{\footnotesize \textbf{Query complexity and time {$\downarrow$}} for computing the scores to select informative samples from a from the unlabeled pool at each iteration. Here, $K=10$ are the number of classes; $\tau=50$ are the number of forward passes; $D=512$ is the feature dimension; $L=80$ is the reduced feature dimension following PCA (for SCAL-fre); and, finally, $M$ are the number of labeled samples. $F$ indicates the computation in one forward pass of the Resnet18, and $F_{\textit{LP}}$ indicates one forward pass of the loss prediction module. The average query time (unit relative to SCAL) is for computing the scores to select 1k samples from a subset of 10k unlabelled samples with ResNet-18/CIFAR-10.}
\label{tab:query-table}
\begin{center}
\begin{footnotesize}
\begin{tabular}{lcc}
\toprule
Query Method & {\begin{tabular}{c} Computational \\Complexity \end{tabular}} & {\begin{tabular}{c} Avg. Query \\Time unit {$\downarrow$} \end{tabular}}\\
\midrule
Entropy  & $F + \mathcal{O}(K)$  & 0.73 \\
Learning Loss & {\begin{tabular}{c} $F + F_{\scriptscriptstyle{LP}}+ \mathcal{O}(K)$ \end{tabular}} & 1.32\\
CoreSet & $F +CS$ {\tnote{\S}} & 11.06\\
BALD & $\tau \cdot F + \mathcal{O}(\tau K) $  &  26.05 {\tnote{\textdagger}}\\
SCAL (\textit{fre}) & $F + \mathcal{O}(KDL)$  & 1.0 \\
SCAL (\textit{featuresim}) & $F + \mathcal{O}(MD)$ & 1.0 \\
\bottomrule
\end{tabular}

\begin{tablenotes}
        \small
        \item[\S] $CS:$ CoreSet is NP-hard
        \item[\textdagger] $\tau=50$ stochastic forward passes (Monte Carlo Dropout) 
\end{tablenotes}
\end{footnotesize}
\end{center}
\end{threeparttable}
\end{table}

\subsection{Results}
\label{sec:results}
\textbf{Accuracy:} Fig.~\ref{fig:acc} shows all the methods improve accuracy as more training samples are acquired and perform better than the random selection of samples. SCAL method outperforms in the imbalanced setup (Imbalanced-CIFAR10 and SVHN) and in the initial active learning iterations for balanced setup (CIFAR10), even with fewer samples. On an average over 10 iterations, SCAL yields 5.7\%, 14.9\% and 6.2\% higher accuracy compared to random selection and 1.2\%, 5.1\% and 1.1\% higher accuracy compared to BALD (next best performing method), when evaluated on CIFAR10, Imbalanced-CIFAR10 and SVHN datasets respectively.



The upper bound in Fig.~\ref{fig:acc} indicates the maximum accuracy achieved when model is trained with entire dataset (Imbalanced-CIFAR10: 79.53\%, CIFAR10: 93.04\%, SVHN: 95.93\%). To achieve 80\% of upper bound accuracy, SCAL method required 39\% lesser labeled samples compared to entropy based selection and 28.5\% lesser labeled samples compared to BALD on Imbalanced-CIFAR10. Similarly, to achieve maximum upper bound accuracy SCAL required 48\% lesser labeled samples as compared to random selection and 27.8\% lesser labeled samples compared to BALD, requiring an oracle to label much fewer data samples to be annotated to achieve same accuracy. We provide the full learning curves in Appendix Fig.~\ref{fig:ablation_full_learning_curve}. 



\begin{figure}[!t]
	\small
	\begin{subfigure}
		\centering
		\captionsetup{
			justification=centering}
		\includegraphics[scale=0.42]{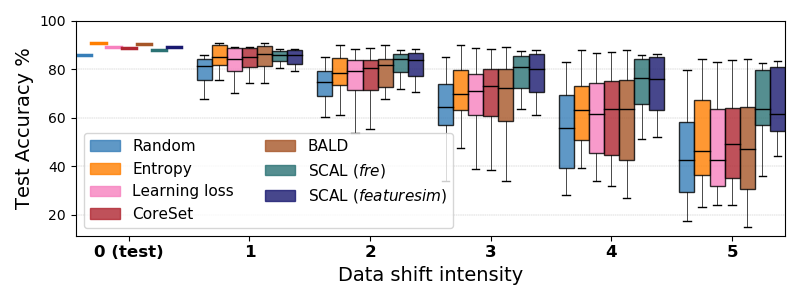}
	\end{subfigure}
	\begin{subfigure}
		\centering
		\captionsetup{
			justification=centering}
		\includegraphics[scale=0.42]{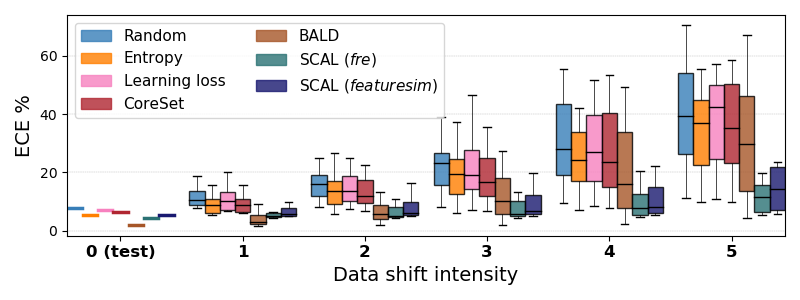}
	\end{subfigure}
	\begin{subfigure}
		\centering
		\captionsetup{
			justification=centering}
		\includegraphics[scale=0.42]{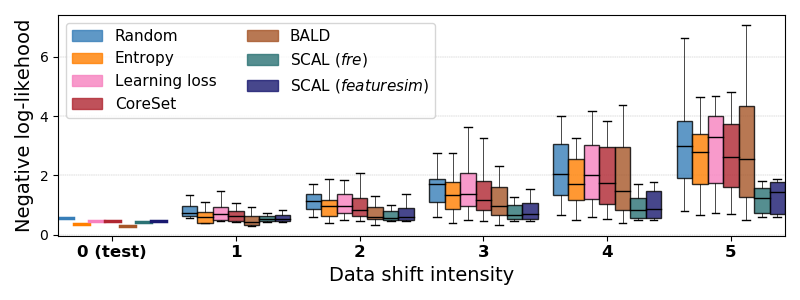}
	\end{subfigure}
	\caption{\small Robustness to dataset shift comparison of models derived with different query methods in active learning setting. Evaluation using Accuracy$\uparrow$, ECE$\downarrow$ and Negative log-likelihood (NLL)$\downarrow$ on CIFAR10-C at different levels of shift intensities (1-5). At each data shift intensity level, the boxplot summarizes the results across 16 different shift types showing the min, max and quartiles. Our proposed \textbf{SCAL} consistently yields higher accuracy, lower ECE and NLL even with increased dataset shift intensity, demonstrating better robustness compared to other methods.}
	\label{fig:boxplots}
\end{figure}

\begin{figure}[!t]
	\small
	\begin{subfigure}
		\centering
		\captionsetup{
			justification=centering}
		\includegraphics[scale=0.36]{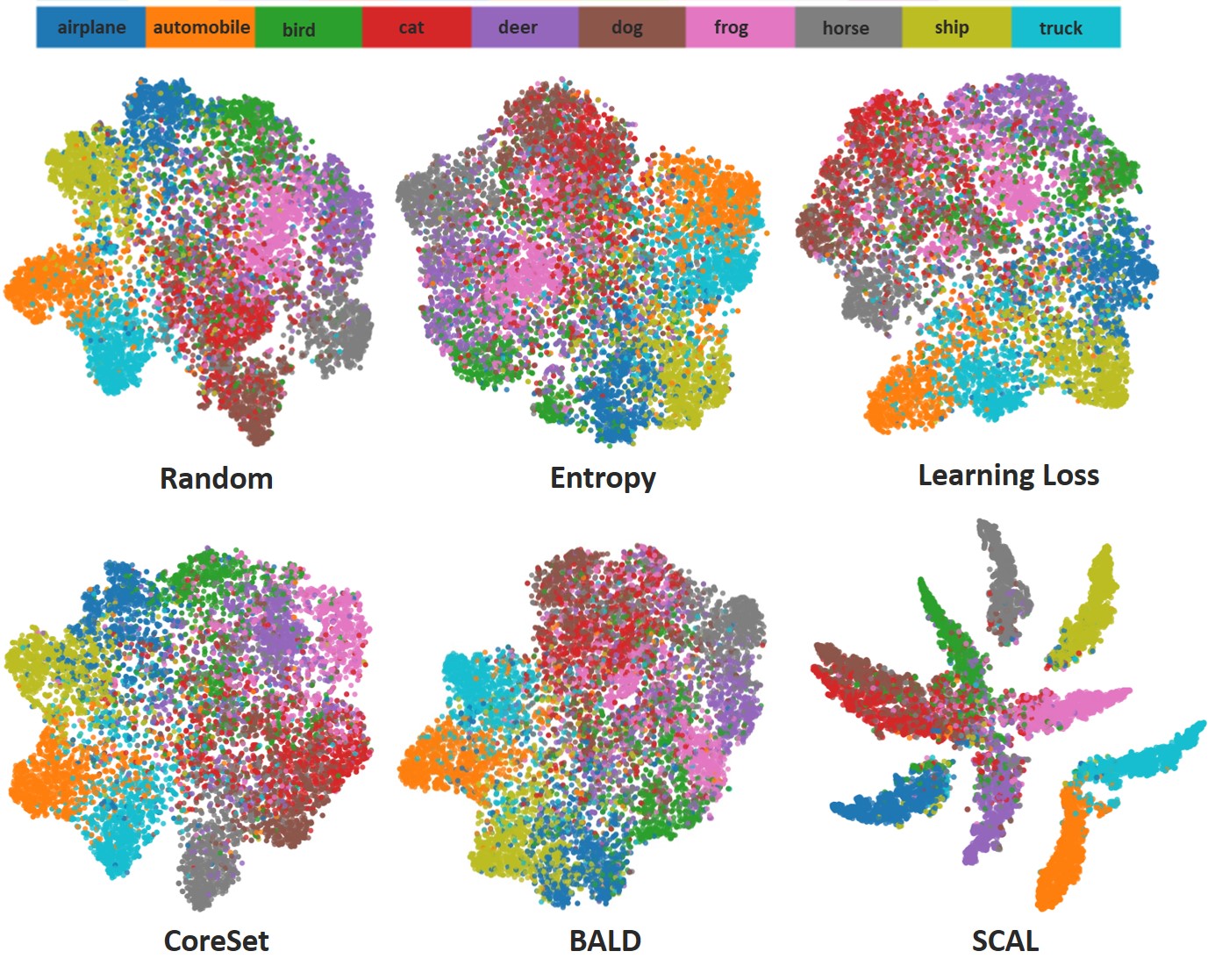}
	\end{subfigure}
	\vskip -0.15in
	\caption{\small t-SNE of CIFAR10-C (dataset shift: Gaussian blur corruption, intensity level 3) feature distribution from models obtained with different query methods at fourth iteration in active learning setting. Note that the models are trained with clean CIFAR10.
}
	\label{fig:tsneshiftplot}
\end{figure}

\begin{figure*}[!ht]
		\centering     
		\subfigure[Imbalanced-CIFAR10]{\includegraphics[width=0.65\columnwidth]{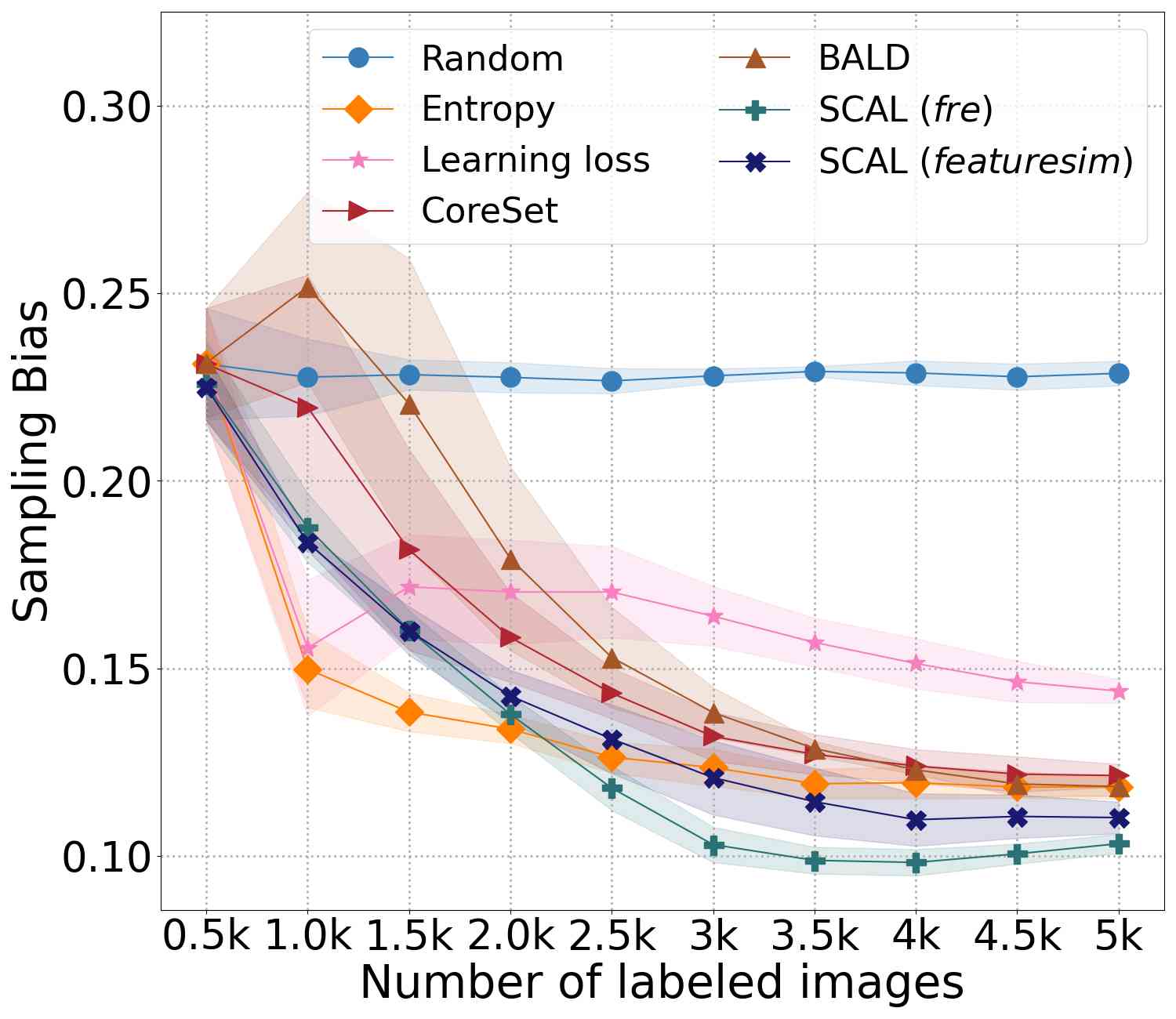}}
		\subfigure[CIFAR10]{\includegraphics[width=0.65\columnwidth]{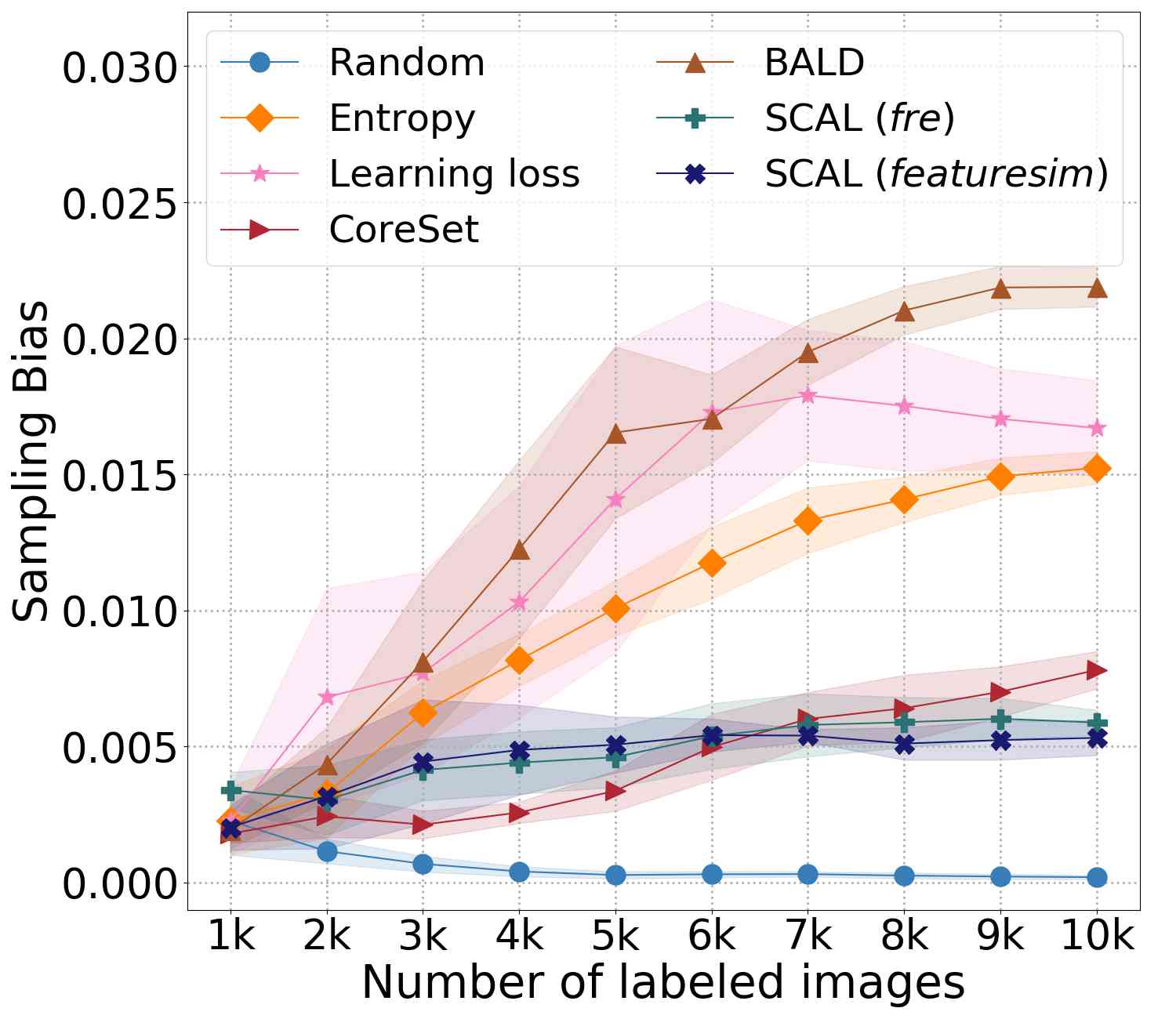}}
		\subfigure[SVHN]{\includegraphics[width=0.65\columnwidth]{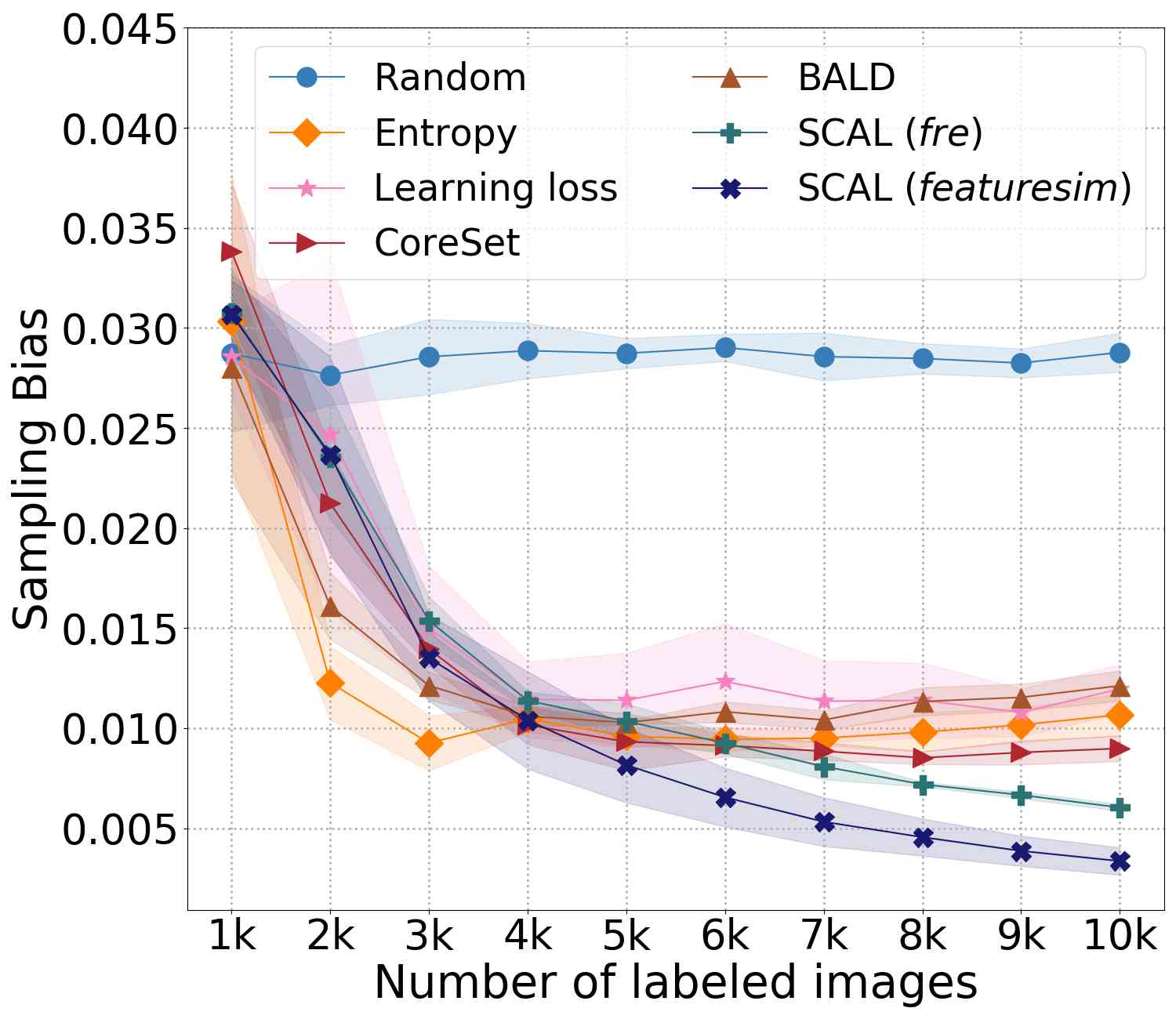}}
		\caption{\small Sampling bias $\downarrow$ evaluation of different query methods on three different datasets. It is to be noted that Imbalanced-CIFAR10 and SVHN has an inherent bias in the dataset with significant imbalance in the number of classes, while CIFAR10 is a balanced dataset. SCAL has lower sampling bias learning from both balanced and imbalanced datasets.}
		\vskip -0.2in
\label{fig:bias}
\end{figure*}
\textbf{Model Calibration:} Fig.~\ref{fig:ece} SCAL models are well calibrated yielding on an average 6\%, 30.8\% and 3.4\% lower ECE compared to random selection over 10 iterations. Fig.~\ref{fig:ece} shows that SCAL yields lowest ECE on long-tailed imbalanced CIFAR10 while BALD provides lower ECE for fairly balanced datasets at the expense of significantly higher computational cost.

\begin{table}[!b]
\caption{\small Robustness to distributional shift after 10th active learning iteration (models trained only with 20\% of total available samples). OOD evaluation with SVHN and dataset shift evaluation with CIFAR10-C for models trained on CIFAR10.} 
\label{tab:ood_datashift_metrics}
\begin{center}
\begin{small}
\begin{tabular}{ccccccccccccc}
\toprule
\multirow{ 2}{*}{Methods} &  AUROC \textbf{{$\uparrow$}} & mCE \textbf{{$\downarrow$}} &  ECE \textbf{{$\downarrow$}} \\
						  &  (OOD) &  (datashift) &  (datashift) \\\midrule
Random &  79.18 & 34.35 &  21.84 \\
Entropy &  86.41 & 30.52 &  20.09 \\
Learning Loss &  85.55 &  31.32 &  21.57 \\
CoreSet &  84.85 & 29.71 &  19.04 \\
BALD &  86.53 & 31.50 &  13.73 \\
SCAL (\textit{fre}) &  \textbf{92.95} & \textbf{22.84} &  \textbf{8.35} \\
SCAL (\textit{featuresim}) &  \textbf{93.69} & \textbf{21.63} &  \textbf{9.24} \\
\bottomrule
\end{tabular}
\end{small}
\end{center}
\end{table}

\textbf{Robustness to dataset shift:} Fig.~\ref{fig:robustness}~(a)~\&~(b) shows models with SCAL method outperforms all the other methods by a bigger margin yielding 9.9\% lower mean corruption error and 7.2\% lower expected calibration error under dataset shift, on an average over 10 iterations. The boxplots in Fig.~\ref{fig:boxplots} summarize the results across all shift types and intensity levels of the final model derived from active learning iterations. Even at increased intensity of dataset shift, SCAL provides higher accuracy, lower ECE and NLL demonstrating robust and well-calibrated models in active learning setting. The superior robustness from SCAL method is attributed to the powerful feature representations learnt from the selected samples. Fig.~\ref{fig:tsneshiftplot} shows the feature embeddings of SCAL method are well clustered and well separated between classes even under dataset shift, as compared to other methods. 

\textbf{Robustness to OOD data:} We compare the OOD detection performance for a model trained on CIFAR10 and evaluated on SVHN. 
For each method, the same scoring function that was used to rank and select the samples in active learning iteration is used to compute the AUROC for OOD detection. As shown in Fig.~\ref{fig:robustness} (c), and in Table~\ref{tab:ood_datashift_metrics}, SCAL method yields consistently higher AUROC compared to other state-of-the-arts methods. In Table~\ref{tab:ood_datashift_metrics}, we provide robustness metrics for the final model after the 10th iteration.

\textbf{Query complexity:} In Table~\ref{tab:query-table}, we compare the query complexity and time for all the methods. All measurements were captured on the same compute machine using the same experimental settings. SCAL is 26x faster than BALD (which involves multiple foward passes), and 11x faster than Coreset (which involves solving the NP-hard k-Center problem). For all other methods, the times are comparable since the query time is dominated by the forward-pass of the model during inference, while the actual score calculation is only a small fraction.

\textbf{Sampling bias:} The plots in Fig.~\ref{fig:bias} shows SCAL with $~\textit{featuresim}$ and $\textit{fre}$ query methods help in minimizing the sampling bias for both balanced and imbalanced datasets. Imbalanced-CIFAR10 and SVHN are  imbalanced datasets with inherent bias between classes, which is reflected in the random selection of the samples at every iteration. 

\begin{figure}[!b]
\vskip -0.2in
    \centering
    \centerline{\includegraphics[width=0.9\columnwidth]{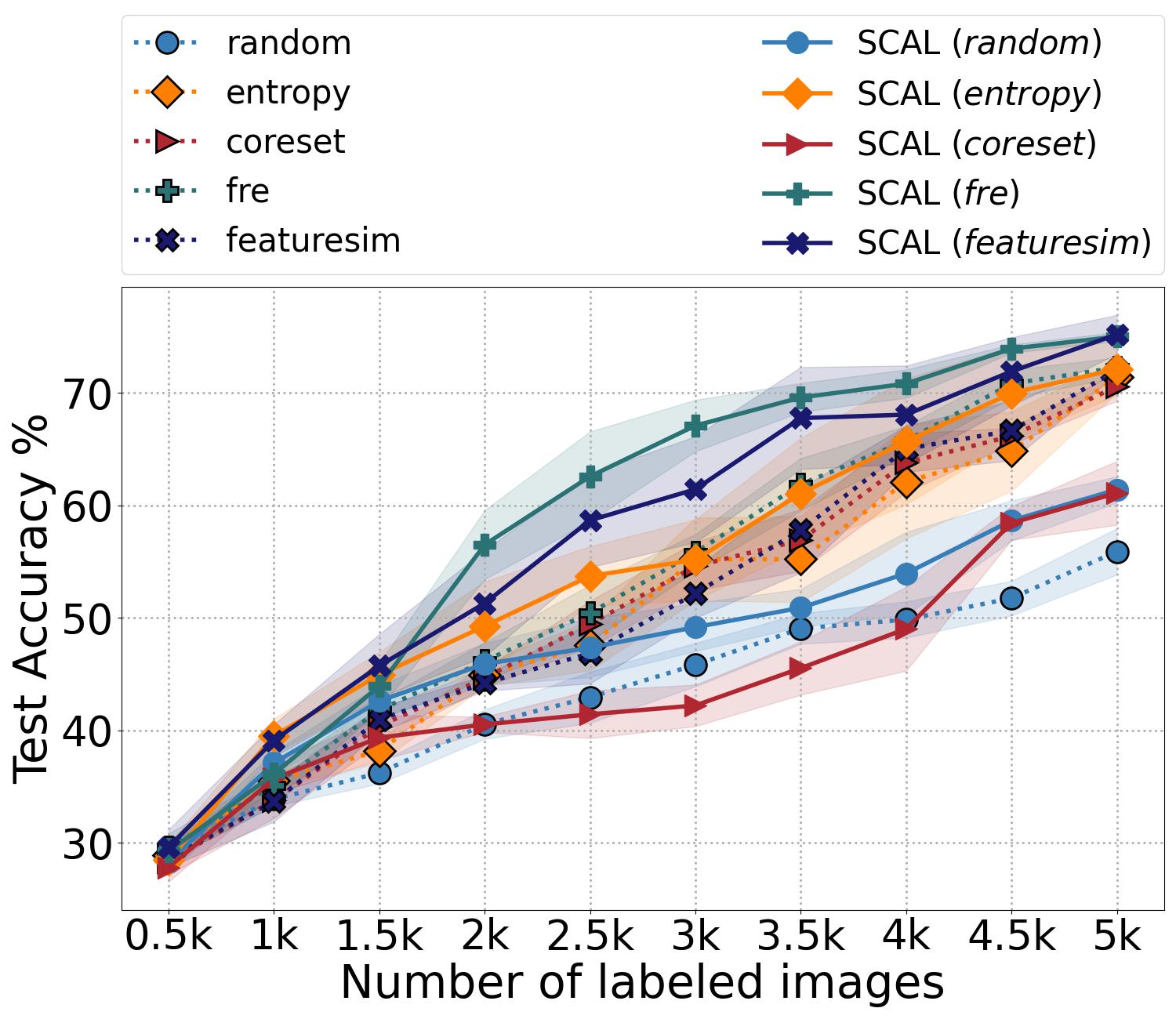}}
    \caption{Ablation study on Imbalanced-CIFAR10: The effect of query functions and training the model with cross-entropy vs contrastive loss in active learning setting.}
\label{fig:loss_query_ablation}
\end{figure}

\textbf{Qualitative Analysis using tSNE:} We study the feature representations obtained from different methods with t-SNE~\cite{van2008visualizing} embeddings. Fig.~\ref{fig:tsneplot} shows the feature embeddings from different methods at 4th iteration of the active learning. The plots visualize the data samples corresponding to 10 classes with 10 different colors. The samples shown with black markers are the most informative samples selected by the query strategies of corresponding methods. We notice that the features of SCAL method are well clustered even at 4th iteration and show lower class imbalance ratio~\cite{cao2019learning}. This signifies~\textit{featuresim} selected balanced samples yet diverse and informative samples (samples in-between clusters and from edge of clusters) from each clusters. Further qualitative analysis on sample selection exploration is presented in Appendix~\ref{appdx:tsne}.

\textbf{Ablation studies:} We performed an ablation study to understand the effect of proposed query functions $\textit{featuresim}$ and $\textit{fre}$, and training loss function in active learning setting. We compared the query methods including Random, Entropy, CoreSet, $\textit{featuresim}$ and $\textit{fre}$ while training the models with cross-entropy loss and contrastive loss separately as shown in Fig.~\ref{fig:loss_query_ablation}. Detailed results from our ablation studies on CIFAR10 and Imbalanced-CIFAR10 is provided in Appendix~\ref{appdx:ablation}. We notice contrastive loss helps in learning better feature representation and improving robustness under distributional shift, while the proposed query functions help in selecting unbiased and diverse samples that guides the contrastive loss to learn feature representations from the most informative samples. 
CoreSet approach which performs well with cross-entropy loss, suffers when trained with contrastive loss resulting in higher sampling bias and lower accuracy. We find that combining contrastive loss and proposed query functions ($\textit{featuresim}$ and $\textit{fre}$) together benefits in active learning setting, as observed in Fig.~\ref{fig:loss_query_ablation}.

\section{Conclusion}
\label{sec:discussion}
We introduced supervised contrastive active learning by proposing computationally inexpensive yet effective query methods to select diverse and informative data samples in active learning. The proposed methods yield well-calibrated models while requiring far fewer labeled samples to attain state-of-the-art accuracy in both imbalanced and balanced dataset setup.
We compared the robustness of models to distributional shift derived from various state-of-the-art query methods in the active learning setting. The supervised contrastive active learning outperforms existing high-performing methods by a big margin in robustness to dataset shift and out-of-distribution. Further, our query strategy reduces sampling bias from both balanced and long-tailed imbalanced datasets. We envision the proposed active learning method can help in building efficient, robust, fair and trustworthy models with much lesser labeled data.

\bibliography{scal}

\begin{thebibliography}{54}
\providecommand{\natexlab}[1]{#1}
\providecommand{\url}[1]{\texttt{#1}}
\expandafter\ifx\csname urlstyle\endcsname\relax
  \providecommand{\doi}[1]{doi: #1}\else
  \providecommand{\doi}{doi: \begingroup \urlstyle{rm}\Url}\fi

\bibitem[Ahuja et~al.(2019)Ahuja, Ndiour, Kalyanpur, and
  Tickoo]{ahuja2019probabilistic}
Ahuja, N.~A., Ndiour, I., Kalyanpur, T., and Tickoo, O.
\newblock Probabilistic modeling of deep features for out-of-distribution and
  adversarial detection.
\newblock \emph{Fourth workshop on Bayesian Deep Learning at NeurIPS}, 2019.

\bibitem[Beluch et~al.(2018)Beluch, Genewein, N{\"u}rnberger, and
  K{\"o}hler]{beluch2018power}
Beluch, W.~H., Genewein, T., N{\"u}rnberger, A., and K{\"o}hler, J.~M.
\newblock The power of ensembles for active learning in image classification.
\newblock In \emph{Proceedings of the IEEE Conference on Computer Vision and
  Pattern Recognition}, pp.\  9368--9377, 2018.

\bibitem[Bhatt et~al.(2021)Bhatt, Antor{\'a}n, Zhang, Liao, Sattigeri,
  Fogliato, Melan{\c{c}}on, Krishnan, Stanley, Tickoo,
  et~al.]{bhatt2020uncertainty}
Bhatt, U., Antor{\'a}n, J., Zhang, Y., Liao, Q.~V., Sattigeri, P., Fogliato,
  R., Melan{\c{c}}on, G.~G., Krishnan, R., Stanley, J., Tickoo, O., et~al.
\newblock Uncertainty as a form of transparency: Measuring, communicating, and
  using uncertainty.
\newblock \emph{AAAI/ACM Conference on Artificial Intelligence, Ethics, and
  Society (AIES)}, 2021.

\bibitem[Blundell et~al.(2015)Blundell, Cornebise, Kavukcuoglu, and
  Wierstra]{blundell2015weight}
Blundell, C., Cornebise, J., Kavukcuoglu, K., and Wierstra, D.
\newblock Weight uncertainty in neural network.
\newblock In \emph{International Conference on Machine Learning}, pp.\
  1613--1622, 2015.

\bibitem[Brier(1950)]{brier1950verification}
Brier, G.~W.
\newblock Verification of forecasts expressed in terms of probability.
\newblock \emph{Monthly weather review}, 78\penalty0 (1):\penalty0 1--3, 1950.

\bibitem[Brinker(2003)]{brinker2003incorporating}
Brinker, K.
\newblock Incorporating diversity in active learning with support vector
  machines.
\newblock In \emph{Proceedings of the 20th international conference on machine
  learning (ICML-03)}, pp.\  59--66, 2003.

\bibitem[Buolamwini \& Gebru(2018)Buolamwini and Gebru]{buolamwini2018gender}
Buolamwini, J. and Gebru, T.
\newblock Gender shades: Intersectional accuracy disparities in commercial
  gender classification.
\newblock In \emph{Conference on fairness, accountability and transparency},
  pp.\  77--91. PMLR, 2018.

\bibitem[Cao et~al.(2019)Cao, Wei, Gaidon, Arechiga, and Ma]{cao2019learning}
Cao, K., Wei, C., Gaidon, A., Arechiga, N., and Ma, T.
\newblock Learning imbalanced datasets with label-distribution-aware margin
  loss.
\newblock \emph{Advances in Neural Information Processing Systems},
  32:\penalty0 1567--1578, 2019.

\bibitem[Chen et~al.(2020{\natexlab{a}})Chen, Kornblith, Norouzi, and
  Hinton]{chen2020simple}
Chen, T., Kornblith, S., Norouzi, M., and Hinton, G.
\newblock A simple framework for contrastive learning of visual
  representations.
\newblock In \emph{International conference on machine learning}, pp.\
  1597--1607. PMLR, 2020{\natexlab{a}}.

\bibitem[Chen et~al.(2020{\natexlab{b}})Chen, Kornblith, Swersky, Norouzi, and
  Hinton]{chen2020big}
Chen, T., Kornblith, S., Swersky, K., Norouzi, M., and Hinton, G.~E.
\newblock Big self-supervised models are strong semi-supervised learners.
\newblock In \emph{Advances in Neural Information Processing Systems},
  volume~33, 2020{\natexlab{b}}.

\bibitem[Chen et~al.(2020{\natexlab{c}})Chen, Fan, Girshick, and
  He]{chen2020improved}
Chen, X., Fan, H., Girshick, R., and He, K.
\newblock Improved baselines with momentum contrastive learning.
\newblock \emph{arXiv preprint arXiv:2003.04297}, 2020{\natexlab{c}}.

\bibitem[Cordts et~al.(2016)Cordts, Omran, Ramos, Rehfeld, Enzweiler, Benenson,
  Franke, Roth, and Schiele]{cordts2016cityscapes}
Cordts, M., Omran, M., Ramos, S., Rehfeld, T., Enzweiler, M., Benenson, R.,
  Franke, U., Roth, S., and Schiele, B.
\newblock The cityscapes dataset for semantic urban scene understanding.
\newblock In \emph{Proceedings of the IEEE conference on computer vision and
  pattern recognition}, pp.\  3213--3223, 2016.

\bibitem[Dasgupta(2011)]{dasgupta2011two}
Dasgupta, S.
\newblock Two faces of active learning.
\newblock \emph{Theoretical computer science}, 412\penalty0 (19):\penalty0
  1767--1781, 2011.

\bibitem[Dasgupta \& Hsu(2008)Dasgupta and Hsu]{dasgupta2008hierarchical}
Dasgupta, S. and Hsu, D.
\newblock Hierarchical sampling for active learning.
\newblock In \emph{Proceedings of the 25th international conference on Machine
  learning}, pp.\  208--215, 2008.

\bibitem[Davis \& Goadrich(2006)Davis and Goadrich]{davis2006relationship}
Davis, J. and Goadrich, M.
\newblock The relationship between precision-recall and roc curves.
\newblock In \emph{Proceedings of the 23rd international conference on Machine
  learning}, pp.\  233--240, 2006.

\bibitem[Ducoffe \& Precioso(2018)Ducoffe and Precioso]{ducoffe2018adversarial}
Ducoffe, M. and Precioso, F.
\newblock Adversarial active learning for deep networks: a margin based
  approach.
\newblock \emph{Proceedings of the 35th International Conference on Machine
  Learning}, 2018.

\bibitem[Farquhar et~al.(2020)Farquhar, Gal, and
  Rainforth]{farquhar2021statistical}
Farquhar, S., Gal, Y., and Rainforth, T.
\newblock On statistical bias in active learning: How and when to fix it.
\newblock In \emph{International Conference on Learning Representations}, 2020.

\bibitem[Gal \& Ghahramani(2016)Gal and Ghahramani]{gal2016dropout}
Gal, Y. and Ghahramani, Z.
\newblock Dropout as a bayesian approximation: Representing model uncertainty
  in deep learning.
\newblock In \emph{international conference on machine learning}, pp.\
  1050--1059. PMLR, 2016.

\bibitem[Gal et~al.(2017)Gal, Islam, and Ghahramani]{gal2017deep}
Gal, Y., Islam, R., and Ghahramani, Z.
\newblock Deep bayesian active learning with image data.
\newblock In \emph{International Conference on Machine Learning}, pp.\
  1183--1192. PMLR, 2017.

\bibitem[Gneiting \& Raftery(2007)Gneiting and Raftery]{gneiting2007strictly}
Gneiting, T. and Raftery, A.~E.
\newblock Strictly proper scoring rules, prediction, and estimation.
\newblock \emph{Journal of the American statistical Association}, 102\penalty0
  (477):\penalty0 359--378, 2007.

\bibitem[Golub \& Van~Loan(1996)Golub and Van~Loan]{golub1996matrix}
Golub, G.~H. and Van~Loan, C.~F.
\newblock Matrix computations. johns hopkins studies in the mathematical
  sciences, 1996.

\bibitem[Graf et~al.(2021)Graf, Hofer, Niethammer, and
  Kwitt]{graf2021dissecting}
Graf, F., Hofer, C., Niethammer, M., and Kwitt, R.
\newblock Dissecting supervised constrastive learning.
\newblock In \emph{International Conference on Machine Learning}, pp.\
  3821--3830. PMLR, 2021.

\bibitem[He et~al.(2016)He, Zhang, Ren, and Sun]{he2016deep}
He, K., Zhang, X., Ren, S., and Sun, J.
\newblock Deep residual learning for image recognition.
\newblock In \emph{Proceedings of the IEEE conference on computer vision and
  pattern recognition}, pp.\  770--778, 2016.

\bibitem[He et~al.(2020)He, Fan, Wu, Xie, and Girshick]{he2020momentum}
He, K., Fan, H., Wu, Y., Xie, S., and Girshick, R.
\newblock Momentum contrast for unsupervised visual representation learning.
\newblock In \emph{Proceedings of the IEEE/CVF Conference on Computer Vision
  and Pattern Recognition}, pp.\  9729--9738, 2020.

\bibitem[Hendrycks \& Dietterich(2019)Hendrycks and
  Dietterich]{hendrycks2019benchmarking}
Hendrycks, D. and Dietterich, T.
\newblock Benchmarking neural network robustness to common corruptions and
  perturbations.
\newblock In \emph{International Conference on Learning Representations}, 2019.

\bibitem[Hendrycks \& Gimpel(2017)Hendrycks and Gimpel]{hendrycks17baseline}
Hendrycks, D. and Gimpel, K.
\newblock A baseline for detecting misclassified and out-of-distribution
  examples in neural networks.
\newblock \emph{Proceedings of International Conference on Learning
  Representations}, 2017.

\bibitem[Houlsby et~al.(2011)Houlsby, Husz{\'a}r, Ghahramani, and
  Lengyel]{houlsby2011bayesian}
Houlsby, N., Husz{\'a}r, F., Ghahramani, Z., and Lengyel, M.
\newblock Bayesian active learning for classification and preference learning.
\newblock \emph{arXiv preprint arXiv:1112.5745}, 2011.

\bibitem[Irvin et~al.(2019)Irvin, Rajpurkar, Ko, Yu, Ciurea-Ilcus, Chute,
  Marklund, Haghgoo, Ball, Shpanskaya, et~al.]{irvin2019chexpert}
Irvin, J., Rajpurkar, P., Ko, M., Yu, Y., Ciurea-Ilcus, S., Chute, C.,
  Marklund, H., Haghgoo, B., Ball, R., Shpanskaya, K., et~al.
\newblock Chexpert: A large chest radiograph dataset with uncertainty labels
  and expert comparison.
\newblock In \emph{Proceedings of the AAAI Conference on Artificial
  Intelligence}, volume~33, pp.\  590--597, 2019.

\bibitem[Khosla et~al.(2020)Khosla, Teterwak, Wang, Sarna, Tian, Isola,
  Maschinot, Liu, and Krishnan]{khosla2020supervised}
Khosla, P., Teterwak, P., Wang, C., Sarna, A., Tian, Y., Isola, P., Maschinot,
  A., Liu, C., and Krishnan, D.
\newblock Supervised contrastive learning.
\newblock In \emph{Advances in Neural Information Processing Systems},
  volume~33, pp.\  18661--18673, 2020.

\bibitem[Kirsch et~al.(2019)Kirsch, Van~Amersfoort, and
  Gal]{kirsch2019batchbald}
Kirsch, A., Van~Amersfoort, J., and Gal, Y.
\newblock Batchbald: Efficient and diverse batch acquisition for deep bayesian
  active learning.
\newblock \emph{Advances in neural information processing systems},
  32:\penalty0 7026--7037, 2019.

\bibitem[Krishnan \& Tickoo(2020)Krishnan and Tickoo]{krishnan2020improving}
Krishnan, R. and Tickoo, O.
\newblock Improving model calibration with accuracy versus uncertainty
  optimization.
\newblock \emph{Advances in Neural Information Processing Systems}, 33, 2020.

\bibitem[Krishnan et~al.(2020)Krishnan, Subedar, and
  Tickoo]{krishnan2020specifying}
Krishnan, R., Subedar, M., and Tickoo, O.
\newblock Specifying weight priors in bayesian deep neural networks with
  empirical bayes.
\newblock In \emph{Proceedings of the AAAI Conference on Artificial
  Intelligence}, volume~34, pp.\  4477--4484, 2020.

\bibitem[Krizhevsky et~al.(2009)]{krizhevsky2009learning}
Krizhevsky, A. et~al.
\newblock Learning multiple layers of features from tiny images.
\newblock 2009.

\bibitem[Lewis \& Gale(1994)Lewis and Gale]{lewis1994sequential}
Lewis, D.~D. and Gale, W.~A.
\newblock A sequential algorithm for training text classifiers.
\newblock In \emph{SIGIR’94}, pp.\  3--12. Springer, 1994.

\bibitem[Liu et~al.(2019)Liu, Miao, Zhan, Wang, Gong, and Yu]{liu2019large}
Liu, Z., Miao, Z., Zhan, X., Wang, J., Gong, B., and Yu, S.~X.
\newblock Large-scale long-tailed recognition in an open world.
\newblock In \emph{Proceedings of the IEEE/CVF Conference on Computer Vision
  and Pattern Recognition}, pp.\  2537--2546, 2019.

\bibitem[Naeini et~al.(2015)Naeini, Cooper, and
  Hauskrecht]{naeini2015obtaining}
Naeini, M.~P., Cooper, G., and Hauskrecht, M.
\newblock Obtaining well calibrated probabilities using bayesian binning.
\newblock In \emph{Twenty-Ninth AAAI Conference on Artificial Intelligence},
  2015.

\bibitem[Netzer et~al.(2011)Netzer, Wang, Coates, Bissacco, Wu, and
  Ng]{netzer2011reading}
Netzer, Y., Wang, T., Coates, A., Bissacco, A., Wu, B., and Ng, A.~Y.
\newblock Reading digits in natural images with unsupervised feature learning.
\newblock 2011.

\bibitem[Ovadia et~al.(2019)Ovadia, Fertig, Ren, Nado, Sculley, Nowozin,
  Dillon, Lakshminarayanan, and Snoek]{ovadia2019can}
Ovadia, Y., Fertig, E., Ren, J., Nado, Z., Sculley, D., Nowozin, S., Dillon,
  J., Lakshminarayanan, B., and Snoek, J.
\newblock Can you trust your model\textquotesingle s uncertainty? evaluating
  predictive uncertainty under dataset shift.
\newblock In \emph{Advances in Neural Information Processing Systems},
  volume~32, 2019.

\bibitem[Paszke et~al.(2019)Paszke, Gross, Massa, Lerer, Bradbury, Chanan,
  Killeen, Lin, Gimelshein, Antiga, Desmaison, Kopf, Yang, DeVito, Raison,
  Tejani, Chilamkurthy, Steiner, Fang, Bai, and Chintala]{paszke2019pytorch}
Paszke, A., Gross, S., Massa, F., Lerer, A., Bradbury, J., Chanan, G., Killeen,
  T., Lin, Z., Gimelshein, N., Antiga, L., Desmaison, A., Kopf, A., Yang, E.,
  DeVito, Z., Raison, M., Tejani, A., Chilamkurthy, S., Steiner, B., Fang, L.,
  Bai, J., and Chintala, S.
\newblock Pytorch: An imperative style, high-performance deep learning library.
\newblock In \emph{Advances in Neural Information Processing Systems},
  volume~32, 2019.

\bibitem[Quionero-Candela et~al.(2009)Quionero-Candela, Sugiyama, Schwaighofer,
  and Lawrence]{quionero2009dataset}
Quionero-Candela, J., Sugiyama, M., Schwaighofer, A., and Lawrence, N.~D.
\newblock \emph{Dataset shift in machine learning}.
\newblock The MIT Press, 2009.

\bibitem[Ren et~al.(2021)Ren, Xiao, Chang, Huang, Li, Gupta, Chen, and
  Wang]{ren2021survey}
Ren, P., Xiao, Y., Chang, X., Huang, P.-Y., Li, Z., Gupta, B.~B., Chen, X., and
  Wang, X.
\newblock A survey of deep active learning.
\newblock \emph{ACM Computing Surveys (CSUR)}, 54\penalty0 (9):\penalty0 1--40,
  2021.

\bibitem[Saunshi et~al.(2019)Saunshi, Plevrakis, Arora, Khodak, and
  Khandeparkar]{saunshi2019theoretical}
Saunshi, N., Plevrakis, O., Arora, S., Khodak, M., and Khandeparkar, H.
\newblock A theoretical analysis of contrastive unsupervised representation
  learning.
\newblock In \emph{International Conference on Machine Learning}, pp.\
  5628--5637. PMLR, 2019.

\bibitem[Sener \& Savarese(2018)Sener and Savarese]{sener2018active}
Sener, O. and Savarese, S.
\newblock Active learning for convolutional neural networks: A core-set
  approach.
\newblock In \emph{International Conference on Learning Representations}, 2018.

\bibitem[Settles(2009)]{settles2009active}
Settles, B.
\newblock Active learning literature survey.
\newblock 2009.

\bibitem[Shannon(1948)]{shannon1948mathematical}
Shannon, C.~E.
\newblock A mathematical theory of communication.
\newblock \emph{Bell system technical journal}, 27\penalty0 (3):\penalty0
  379--423, 1948.

\bibitem[Shen et~al.(2018)Shen, Yun, Lipton, Kronrod, and
  Anandkumar]{shen2017deep}
Shen, Y., Yun, H., Lipton, Z.~C., Kronrod, Y., and Anandkumar, A.
\newblock Deep active learning for named entity recognition.
\newblock In \emph{International Conference on Learning Representations}, 2018.

\bibitem[Shui et~al.(2020)Shui, Zhou, Gagn{\'e}, and Wang]{shui2020deep}
Shui, C., Zhou, F., Gagn{\'e}, C., and Wang, B.
\newblock Deep active learning: Unified and principled method for query and
  training.
\newblock In \emph{International Conference on Artificial Intelligence and
  Statistics}, pp.\  1308--1318. PMLR, 2020.

\bibitem[Tack et~al.(2020)Tack, Mo, Jeong, and Shin]{tack2020csi}
Tack, J., Mo, S., Jeong, J., and Shin, J.
\newblock Csi: Novelty detection via contrastive learning on distributionally
  shifted instances.
\newblock \emph{Advances in Neural Information Processing Systems},
  33:\penalty0 11839--11852, 2020.

\bibitem[Tong \& Koller(2001)Tong and Koller]{tong2001support}
Tong, S. and Koller, D.
\newblock Support vector machine active learning with applications to text
  classification.
\newblock \emph{Journal of machine learning research}, 2\penalty0
  (Nov):\penalty0 45--66, 2001.

\bibitem[Tosh et~al.(2021)Tosh, Krishnamurthy, and Hsu]{tosh2021contrastive}
Tosh, C., Krishnamurthy, A., and Hsu, D.
\newblock Contrastive learning, multi-view redundancy, and linear models.
\newblock In \emph{Algorithmic Learning Theory}, pp.\  1179--1206. PMLR, 2021.

\bibitem[Van~der Maaten \& Hinton(2008)Van~der Maaten and
  Hinton]{van2008visualizing}
Van~der Maaten, L. and Hinton, G.
\newblock Visualizing data using t-sne.
\newblock \emph{Journal of machine learning research}, 9\penalty0 (11), 2008.

\bibitem[Wang \& Isola(2020)Wang and Isola]{wang2020understanding}
Wang, T. and Isola, P.
\newblock Understanding contrastive representation learning through alignment
  and uniformity on the hypersphere.
\newblock In \emph{International Conference on Machine Learning}, pp.\
  9929--9939. PMLR, 2020.

\bibitem[Yoo \& Kweon(2019)Yoo and Kweon]{yoo2019learning}
Yoo, D. and Kweon, I.~S.
\newblock Learning loss for active learning.
\newblock In \emph{Proceedings of the IEEE/CVF Conference on Computer Vision
  and Pattern Recognition}, pp.\  93--102, 2019.

\bibitem[Zhang et~al.(2019)Zhang, Li, Zhang, Chen, and
  Wilson]{zhang2019cyclical}
Zhang, R., Li, C., Zhang, J., Chen, C., and Wilson, A.~G.
\newblock Cyclical stochastic gradient mcmc for bayesian deep learning.
\newblock In \emph{International Conference on Learning Representations}, 2019.

\end{thebibliography}
\bibliographystyle{icml2022}

\clearpage
\appendix


\section*{\centering \Large Appendix}
\renewcommand{\thefigure}{F\arabic{figure}}
\setcounter{figure}{0}

\section{Experimental details}
\label{appdx:experiments}
\subsection{Datasets}
\begin{itemize}
\item \textbf{CIFAR-10}~\cite{krizhevsky2009learning}: This dataset consists training set of 50,000 examples and a test set of 10,000 examples. Each example is a 32x32 RGB color image, associated with a label from 10 classes. CIFAR-10 is a balanced dataset with 5000 examples and 1000 examples from each class in training and test set respectively.  In our setup, we consider the 50,000 examples part of the unlabelled pool and evaluate the model with 10,000 test examples at every iteration of the active learning cycle.

\item \textbf{Imbalanced-CIFAR10}~\cite{cao2019learning}: Long-tailed version of CIFAR10 is created following an exponential decay in sample sizes across different classes. The imbalance ratio is set to 50, the ratio of sample size in most frequent class to the sample size in least frequent class. The long-tailed Imbalanced-CIFAR10 dataset with imbalance ratio $\rho=50$ has 13996 train samples with the number of samples in each class include: [5000, 3237, 2096, 1357, 878, 568, 368, 238, 154, 100]. This long-tailed data is used as unlabelled set to train models with active learning. The test set has balanced 10,000 samples.

\item \textbf{SVHN}~\cite{netzer2011reading}: 
The Street View House Numbers (SVHN) dataset has house numbers obtained from Google Street View images to recognize the digits (0-9, representing 10 classes). The dataset has 73257 training and 26032 test images. The images are 32x32 color image patches centered around a single digit. In our setup, we consider the 73257 examples part of the unlabelled pool and evaluate the model with 10,000 test examples at every iteration of the active learning cycle. SVHN dataset has class imbalance with the number of samples in each class include: [4948, 13861, 10585, 8497, 7458, 6882, 5727,  5595, 5045, 4659].

\item \textbf{CIFAR10-C}~\cite{hendrycks2019benchmarking}: CIFAR10 test set (10,000 samples) corrupted with 16 different types of perturbations ('brightness', 'contrast', 'defocus blur', 'elastic transform', 'fog', 'frost', 'gaussian blur', 'gaussian noise', 'glass blur', 'impulse noise', 'pixelate', 'saturate', 'shot noise', 'spatter', 'speckle noise', 'zoom blur') and 5 different levels of intensities (1 to 5), totally 80 different variations of test set. We use this data to evaluate the robustness to dataset shift on models trained with clean CIFAR10.

\end{itemize}

\subsection{Model details and hyperparameters}
\label{appdx:hyperparam}
We use ResNet-18~\cite{he2016deep} model architecture for all the methods and datasets under study. We use the same hyperparameters for all the models for a fair comparison. At each iteration in active learning, the models are trained with labeled samples acquired through the query strategy from respective methods. The models are trained with SGD optimizer for 200 epochs with an initial learning rate of 0.1, batch size of 128, momentum of 0.9 and weight decay of 0.0005. As part of the learning rate schedule, the initial learning rate was multiplied by 0.1 at epoch 160. Note that the same batch size was used for training the cross-entropy and contrastive loss models. Also, same data augmentations in training samples are used for cross-entropy and contrastive models (random horizontal flips and random crops for all the methods, an additional random grayscale with probability of 0.2 is used for SCAL), there is no corruptions or color jittering augmentations included during training for a fair evaluation of robustness to dataset shift using CIFAR10-C. We evaluate the models for 10 iterations. At the end of each iteration, the models were evaluated with the independent labeled test set of 10K samples. We used these hyperparameters and protocol for all three datasets in our experiments.

At every iteration $t$ in the active learning, we obtain a random subset $\mathcal{D}_{\mathrm{S}} \subset \mathcal{D}_\mathrm{U}^t$ of 10K samples from the remaining unlabeled sets for CIFAR10 and SVHN, from which $\mathrm{M}$ most informative samples are selected using query strategy. This subset strategy in the unlabeled pool has been suggested in~\cite{beluch2018power,yoo2019learning} to avoid picking overlapping similar samples.  The subset size is set to 8000 for Imbalanced-CIFAR10 as the number of data samples available in the unlabeled pool is lower (13.9k samples). At every iteration, we choose and annotate $\mathrm{M}=1000$ (for CIFAR10 and SVHN datasets) and $M=500$ (for Imbalanced-CIFAR10) additional samples according to the strategy described in Section \ref{sec:methods}.

\subsection{Implementation details}
\label{appdx:implementaion}
We implemented all the models and methods including SCAL (\textit{featuresim}), SCAL (\textit{fre}), BALD, Entropy and Random for our experiments using PyTorch~\cite{paszke2019pytorch} framework. The experiments for CoreSet~\cite{sener2018active} and Learning loss~\cite{yoo2019learning} methods are performed with the implemented PyTorch models by extending the code available from open-source implementations{\footnote[2]{\tiny \url{ https://github.com/google/active-learning/blob/master/sampling_methods/kcenter_greedy.py}},\footnote[3]{\tiny \url{ https://github.com/Mephisto405/Learning-Loss-for-Active-Learning}}}.

The projection head and classifier for the SCAL method follow the same methodology as in~\cite{khosla2020supervised}. The temperature parameter T was set to 0.07 in the contrastive loss.  The features from the neural network (dimension=512 for ResNet-18) is propagated through a projection network consisting of two linear layers to reduce the feature dimension to 128, on which the contrastive loss is computed during training. After training, the projection head is discarded. A linear classifier is trained using cross-entropy loss on the output features of the neural network, and plugged in during test time. The feature-similarity and feature-reconstruction error scores are computed from the feature representations (dimension=512) of ResNet-18 network from the layer before linear classifier.

\begin{figure*}
		\centering     
		\subfigure[ECE {$\downarrow$}]{\includegraphics[width=0.6\columnwidth]{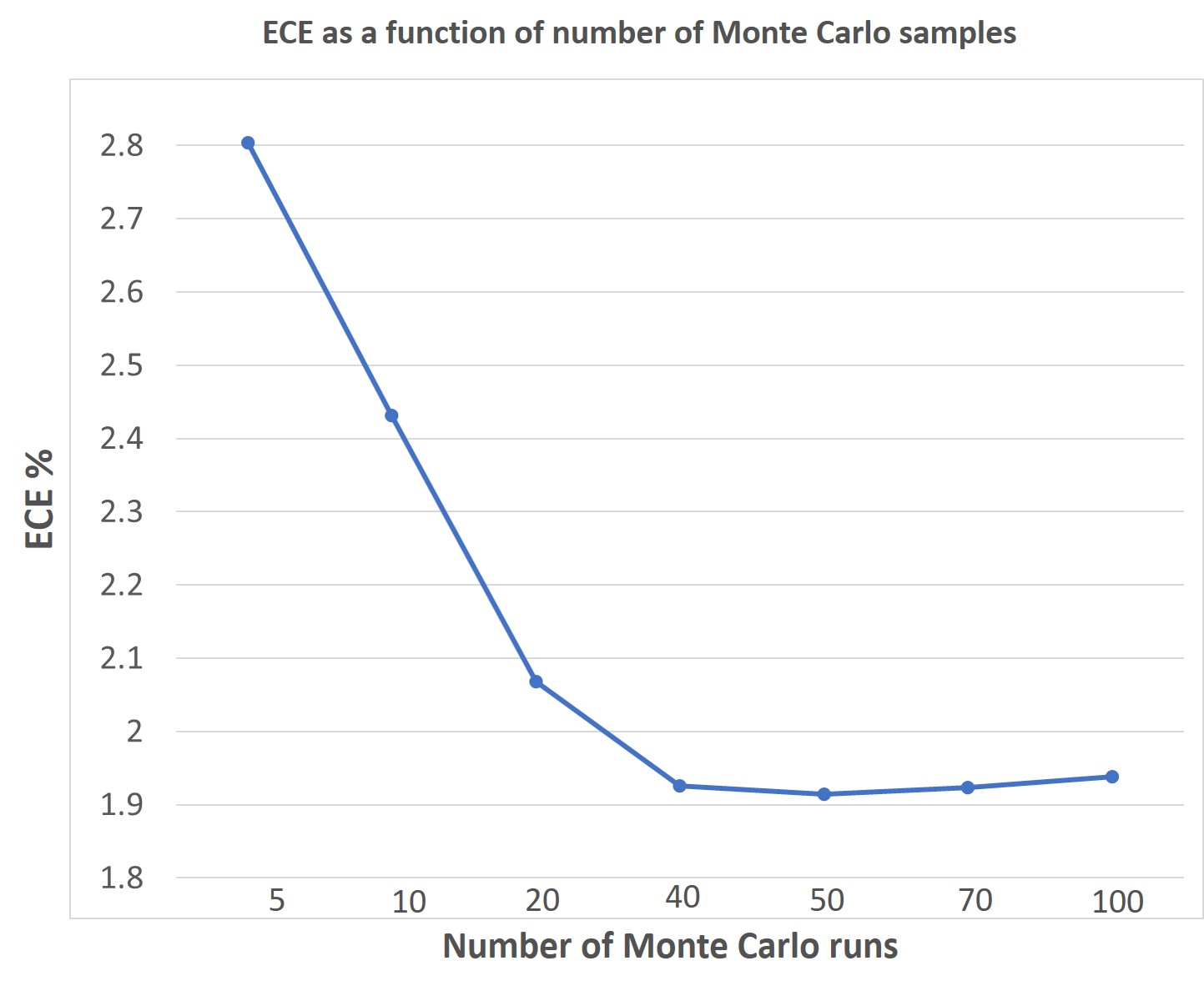}}
		\subfigure[Accuracy {$\uparrow$}]{\includegraphics[width=0.68\columnwidth]{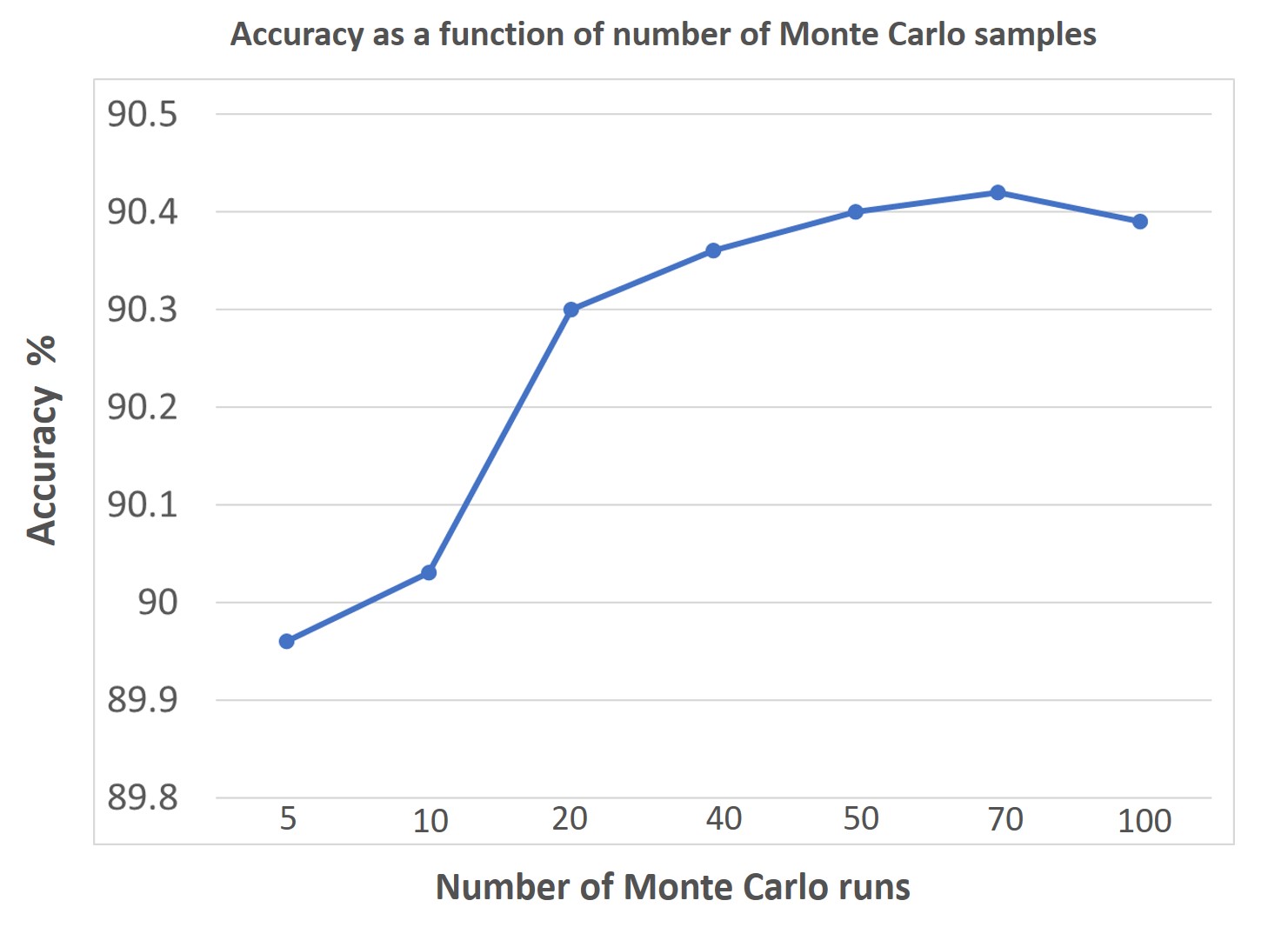}}
		\caption{\small Ablation study (BALD): Accuracy and ECE as a function of number of Monte Carlo runs in BALD method (MC Dropout)}
		\label{fig:ablation_mcdropout}
		\vspace{128in}
\end{figure*}

BALD method is implemented with Monte-Carlo (MC) dropout~\cite{gal2016dropout} by introducing a dropout layer with probability of 0.3 after convolutional layers in the ResNet blocks. We utilize 50 stochastic forward passes with dropout enabled during inference. We selected number of MC runs to be 50 in BALD method as the expected calibration error and accuracy saturated beyond 50 MC runs in our ablation study shown in Figure~\ref{fig:ablation_mcdropout}.

We report the results from 5 independent trials for each method at every active learning iteration.

\subsection{Query strategy}
\label{appdx:query}
\begin{itemize}
\item \textbf{Random}: Choose a random set of samples from unlabeled pool of data.
\item \textbf{Entropy}: Choose a set of samples from an unlabeled pool that yields higher predictive entropy~\cite{shannon1948mathematical}
\begin{equation}
\small
\label{eqn:pe_det}
\begin{aligned}
        \mathcal{H}\left[\mathrm{y}|\mathrm{x},\mathrm{D_{L}}\right]:= 
        -\sum_{k=1}^{K} p\left(\mathrm{y}={c}_{k} | \mathrm{x}, \mathrm{D_{L}}\right) \log \left( p\left(\mathrm{y}={c}_{k} | \mathrm{x}, \mathrm{D_{L}}\right)\right)
\end{aligned}
\end{equation}

\item \textbf{CoreSet}~\cite{sener2018active}: Choose the samples
such that a model learned over the subset is competitive over the whole dataset. The query function selects points that minimizes the maximum Euclidean distance of any point to a center using k-Center-Greedy algorithm.

\item \textbf{Learning Loss}~\cite{yoo2019learning}: Choose a set of samples based on the loss prediction module.

\item \textbf{BALD} (Bayesian Active Learning by Disagreement)~\cite{houlsby2011bayesian,gal2017deep}: Choose a set of samples from an unlabeled pool that yields higher model uncertainty. Monte Carlo dropout~\cite{gal2016dropout} based Bayesian neural network is used to obtain model uncertainty for the unlabeled samples. The model uncertainty is quantified by the mutual information between the posterior distribution of weights and predictive distribution as defined in Equation~\eqref{eqn:model_uncertainty}. BALD can be implemented with other robust approximate Bayesian inference methods such as variational inference~\cite{blundell2015weight,krishnan2020specifying} and stochastic gradient MCMC~\cite{zhang2019cyclical} to get well-calibrated uncertainty estimates, but we study with Monte Carlo Dropout in this paper.

\begin{equation}
\label{eqn:model_uncertainty}  
    \begin{aligned}
        \mathcal{I}[\mathrm{y}, \theta \mid \mathrm{x}, \mathcal{D}_{L}] &:=\mathcal{H}\left[\mathrm{y} \mid \mathrm{x}, \mathcal{D}_{L})\right] \\
        &-\mathbb{E}_{p(\theta \mid \mathcal{D}_{L})}[\mathcal{H}[p(\mathrm{y} \mid \mathrm{x}, \theta)]]
    \end{aligned}
\end{equation}
\item \textbf{SCAL (\textit{featuresim})} and \textbf{SCAL (\textit{fre})}: The sample selection strategy for our proposed methods is described in Section~\ref{subsec:scal}. 
\end{itemize}

\begin{figure*}
		\centering     
		\subfigure[Accuracy {$\uparrow$}]{\includegraphics[width=0.51\columnwidth]{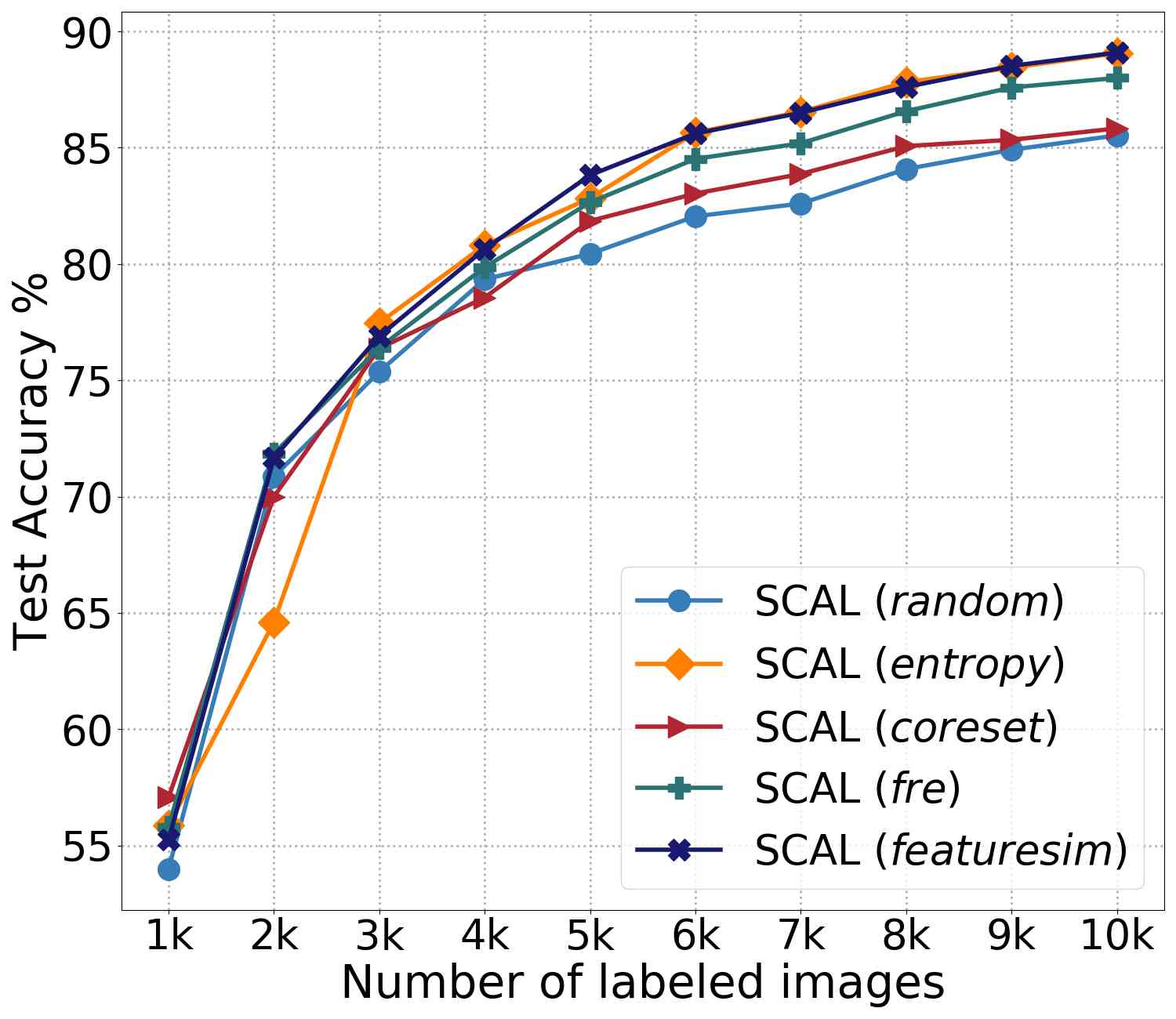}}
		\subfigure[Negative log-likelihood {$\downarrow$}]{\includegraphics[width=0.51\columnwidth]{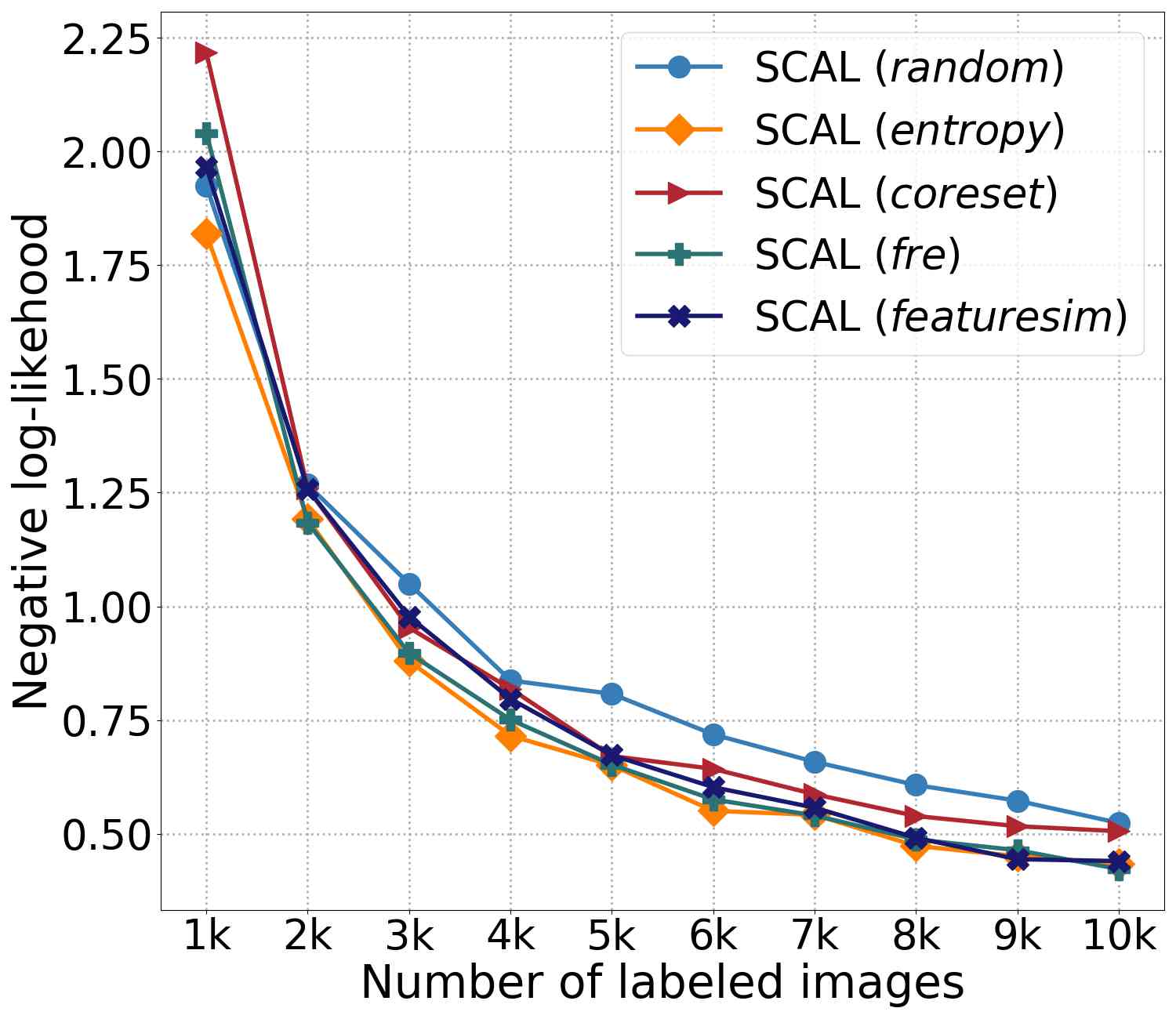}}
		\subfigure[Sampling Bias {$\downarrow$}]{\includegraphics[width=0.51\columnwidth]{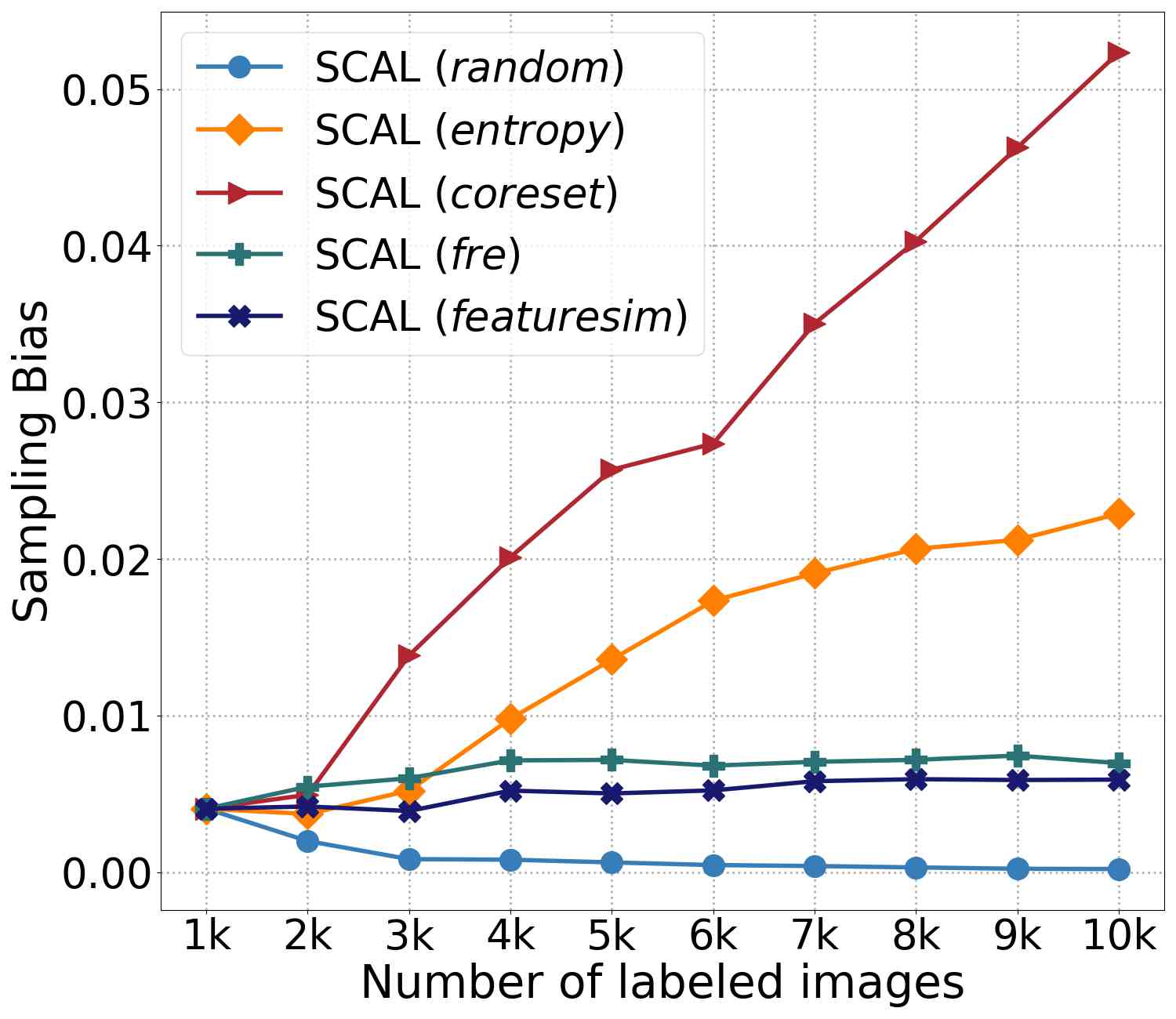}}
		\subfigure[AUROC OOD {$\uparrow$}  ]{\includegraphics[width=0.51\columnwidth]{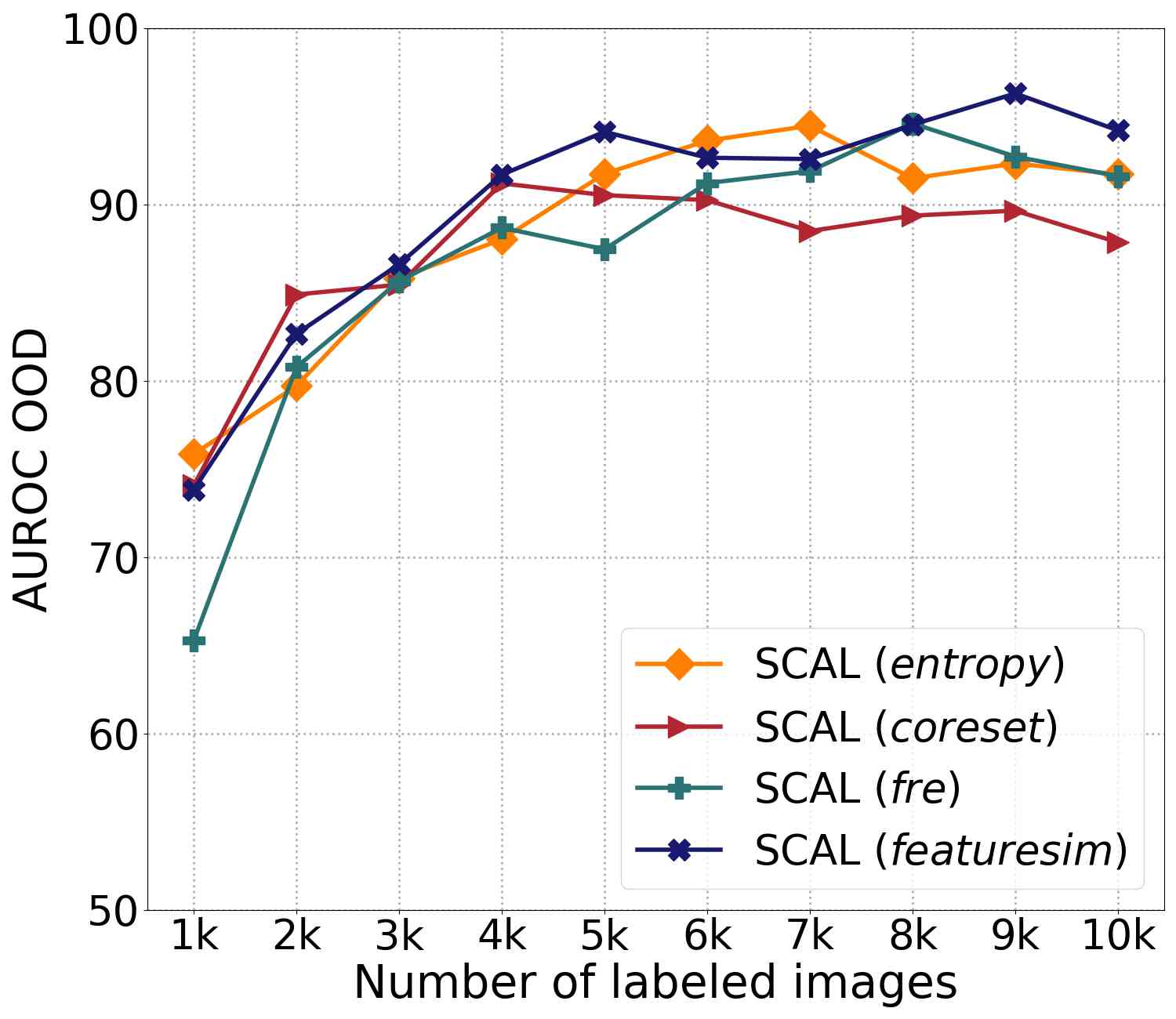}}
		\caption{\small CIFAR10/ResNet-18: Ablation study of different query functions in active learning with models trained with \textbf{contrastive loss}}
		\label{fig:ablation_cifar10}
\end{figure*}
\begin{figure*}
		\centering     
		\subfigure[Accuracy {$\uparrow$}]{\includegraphics[width=0.51\columnwidth]{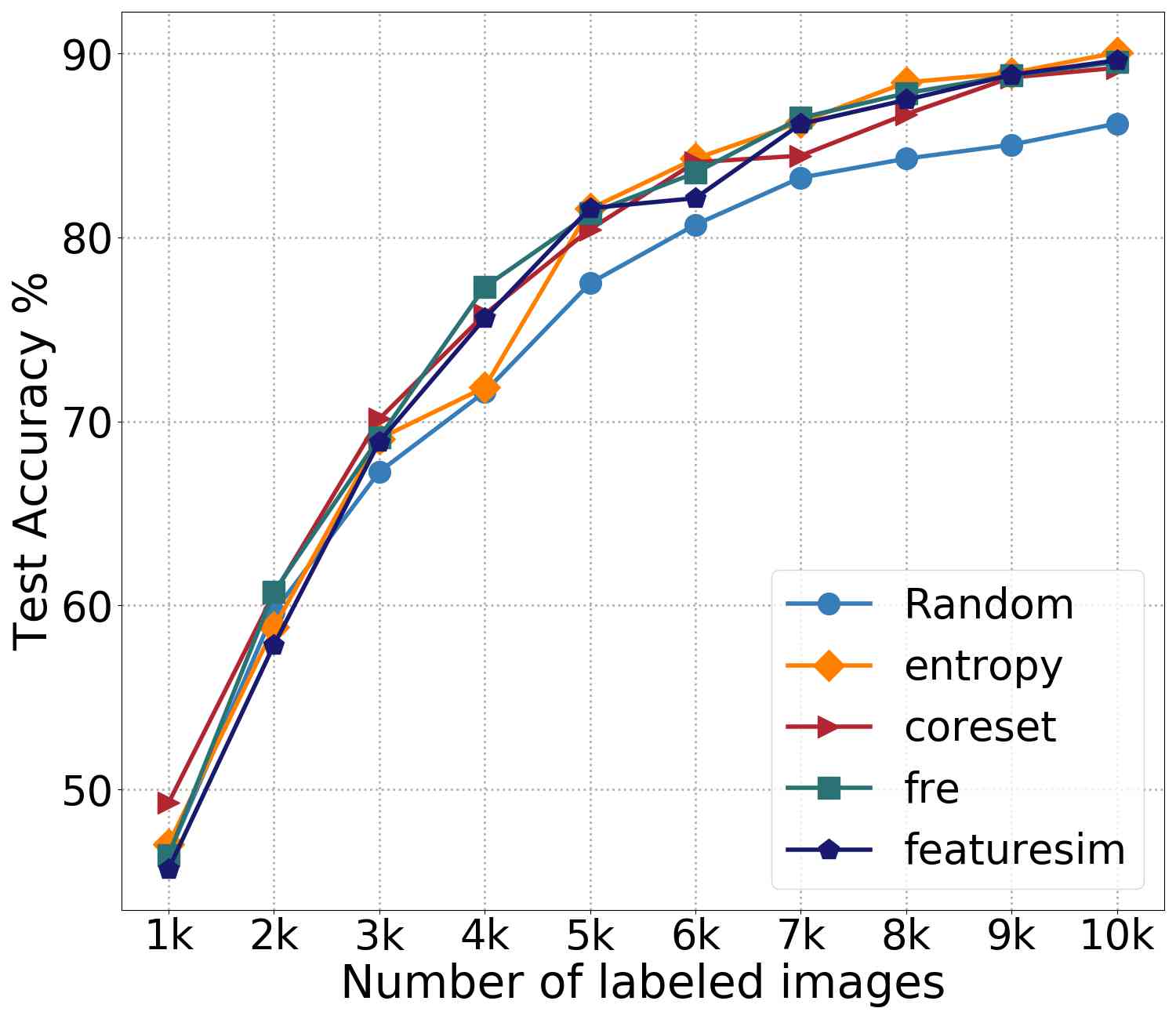}}
		\subfigure[Negative log-likelihood {$\downarrow$}]{\includegraphics[width=0.51\columnwidth]{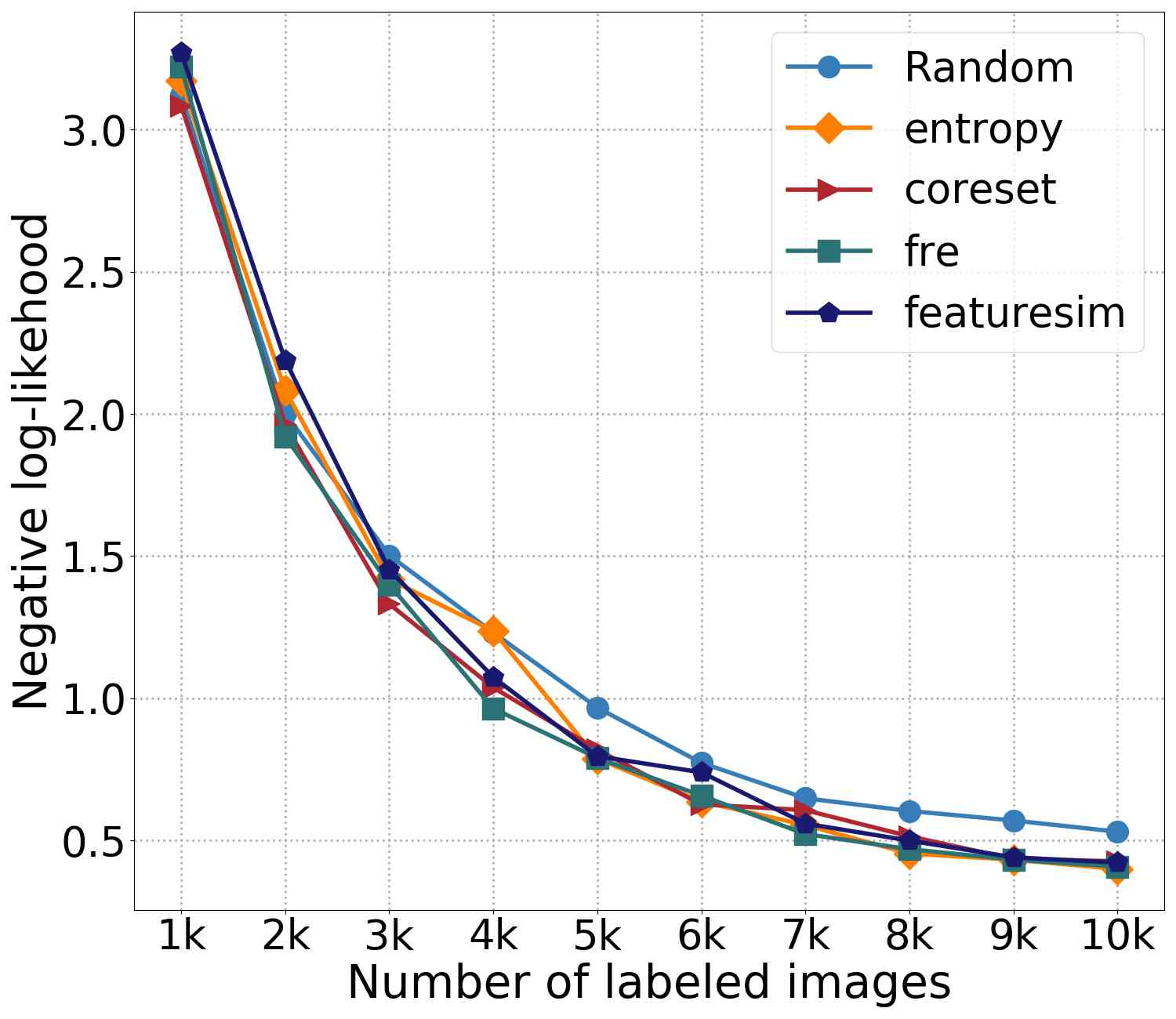}}
		\subfigure[Sampling Bias {$\downarrow$}]{\includegraphics[width=0.51\columnwidth]{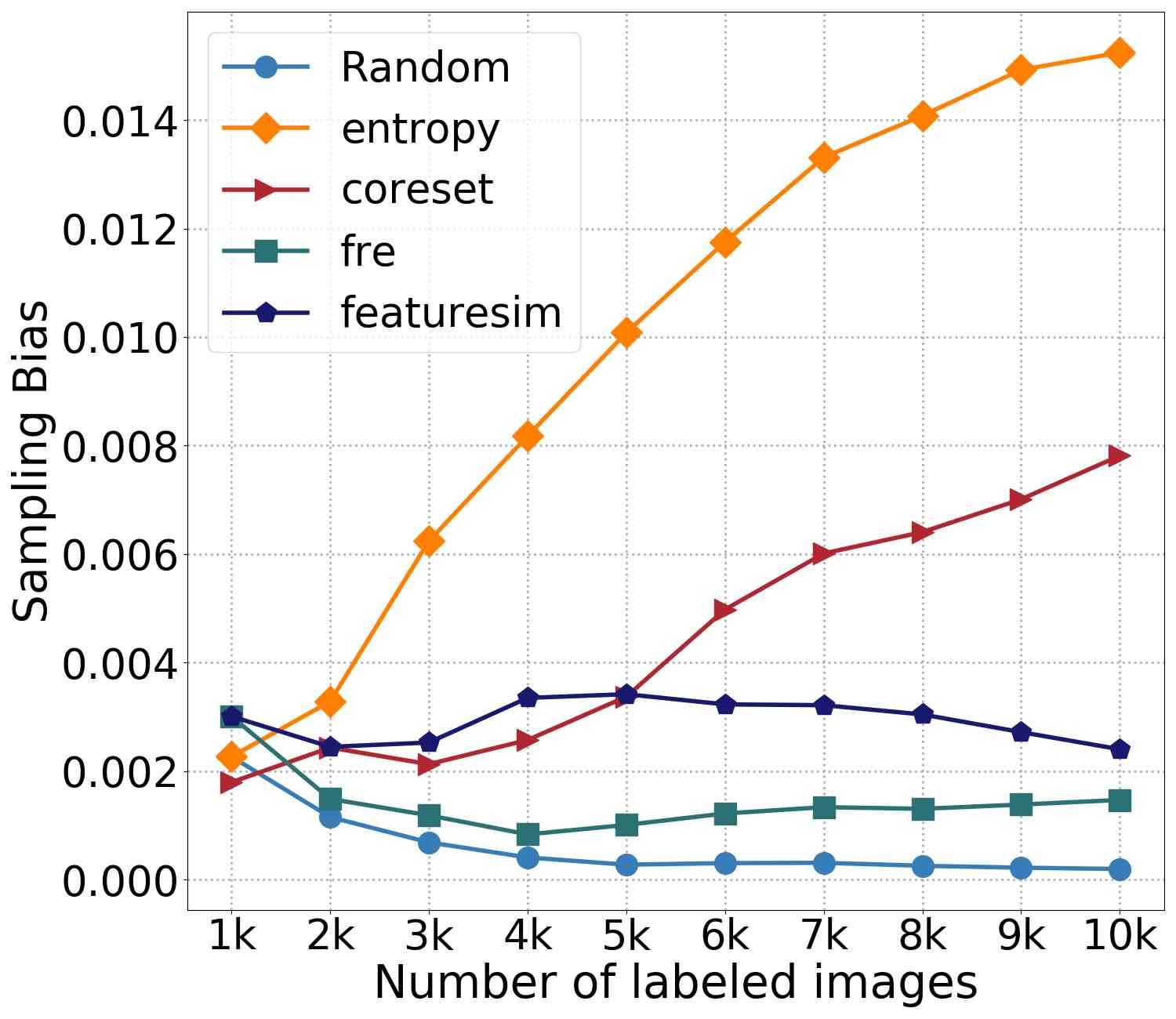}}
		\subfigure[AUROC OOD {$\uparrow$} ]{\includegraphics[width=0.51\columnwidth]{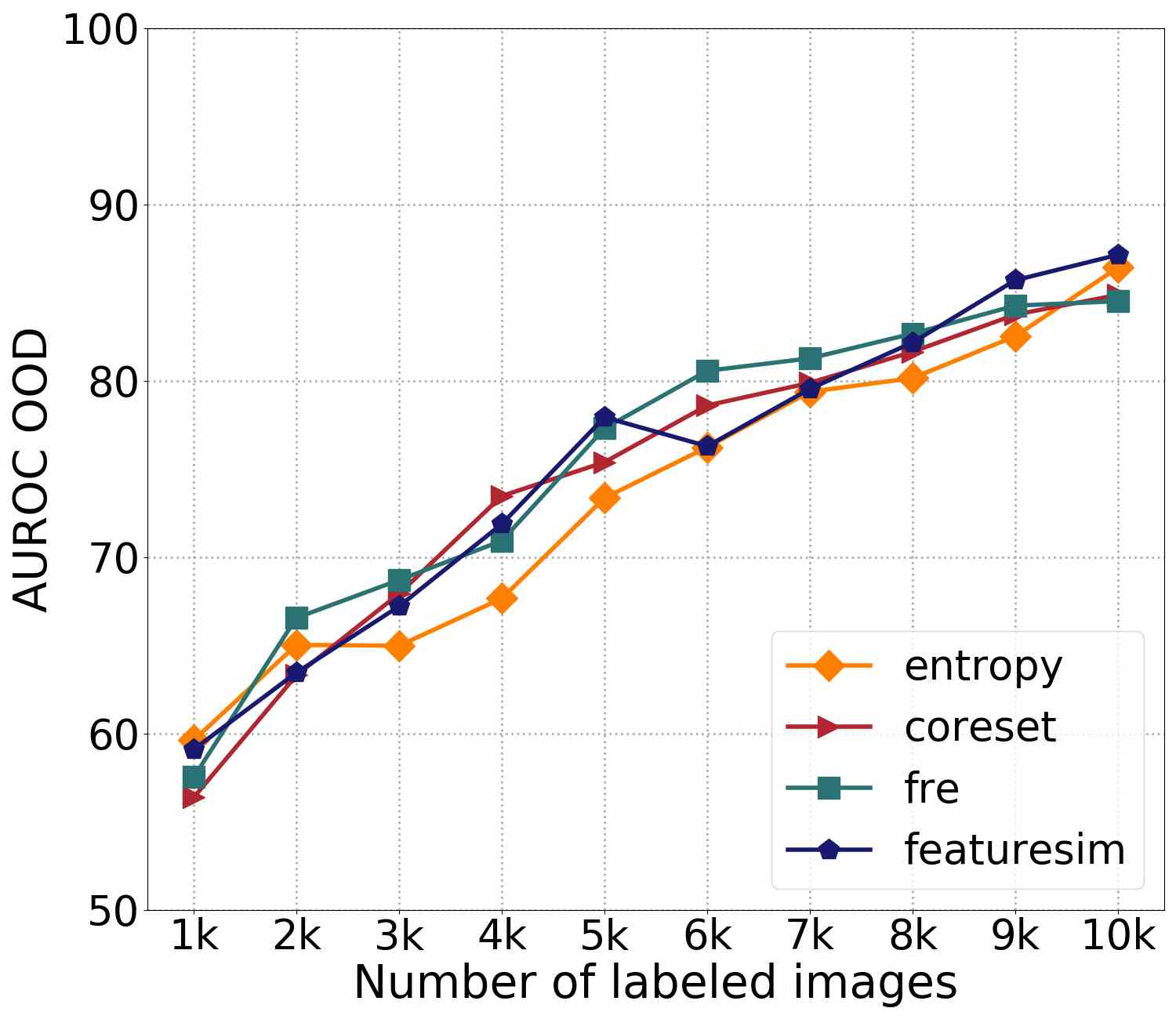}}
		\caption{\small CIFAR10/ResNet-18: Ablation study of different query functions in active learning with models trained with \textbf{cross-entropy loss}}
		\label{fig:ablation_cifar10_ce}
\end{figure*}
\begin{figure*}
		\centering     
		\subfigure[Accuracy {$\uparrow$}]{\includegraphics[width=0.51\columnwidth]{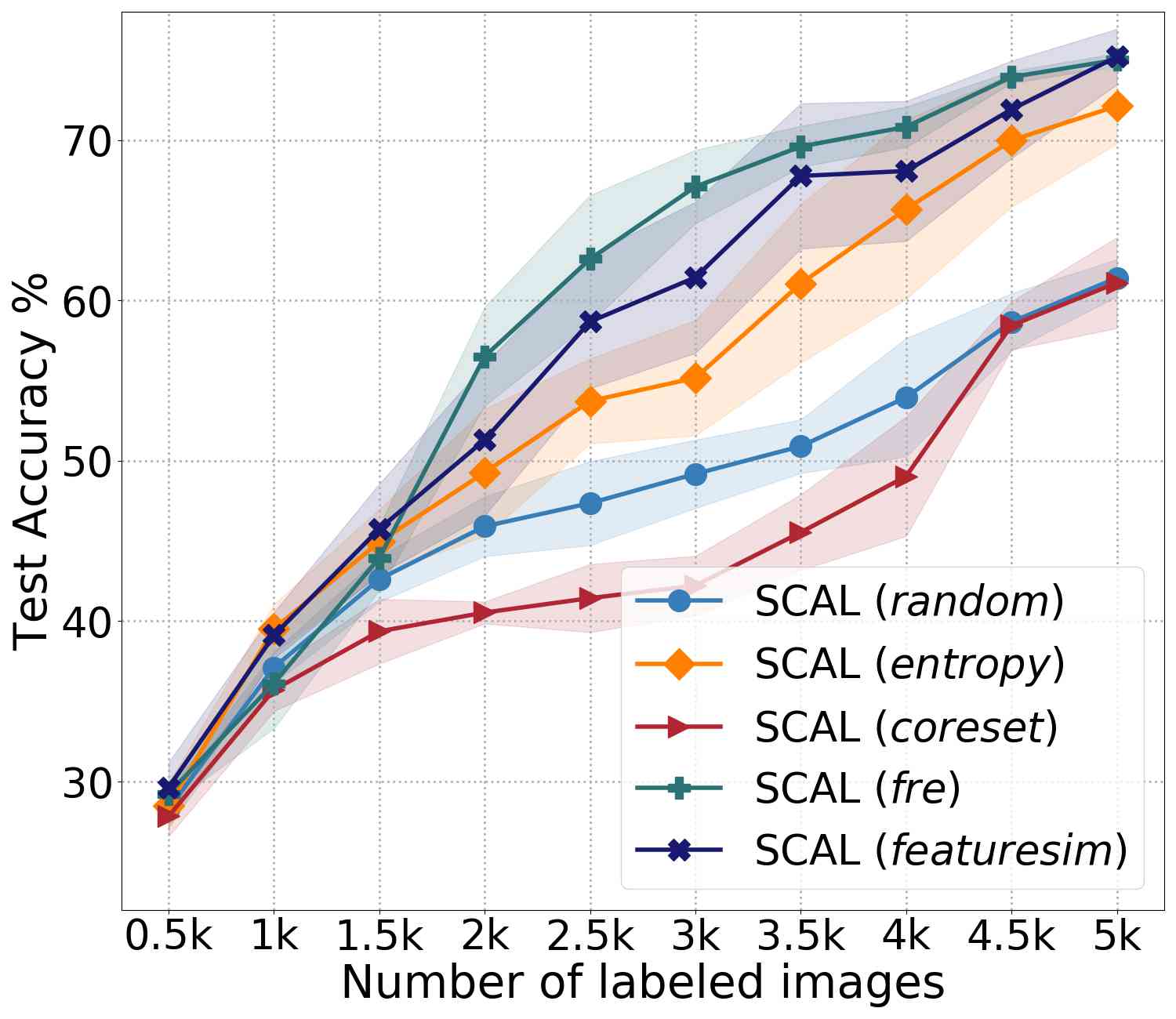}}
		\subfigure[Negative log-likelihood {$\downarrow$}]{\includegraphics[width=0.51\columnwidth]{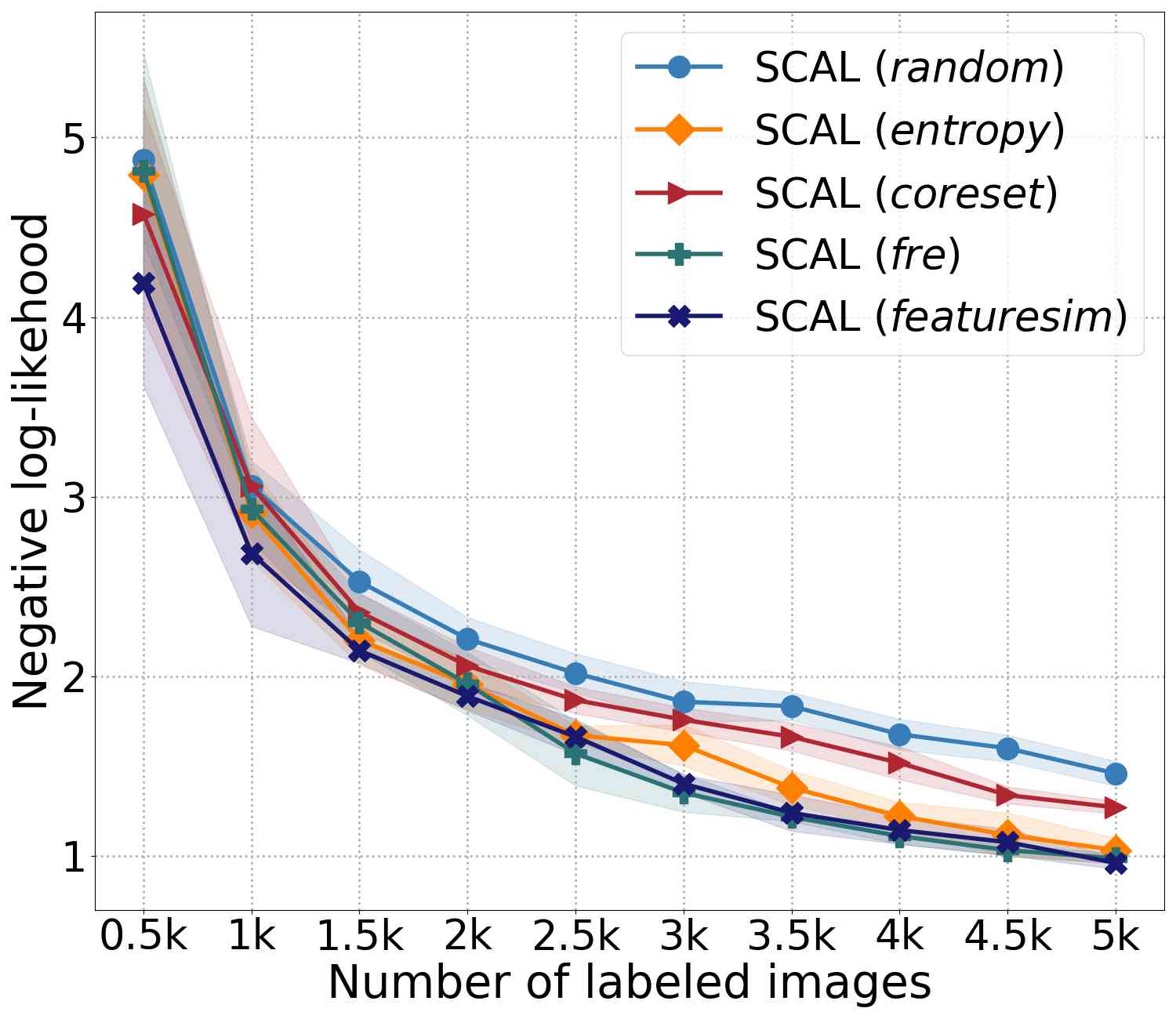}}
		\subfigure[Sampling Bias {$\downarrow$}]{\includegraphics[width=0.51\columnwidth]{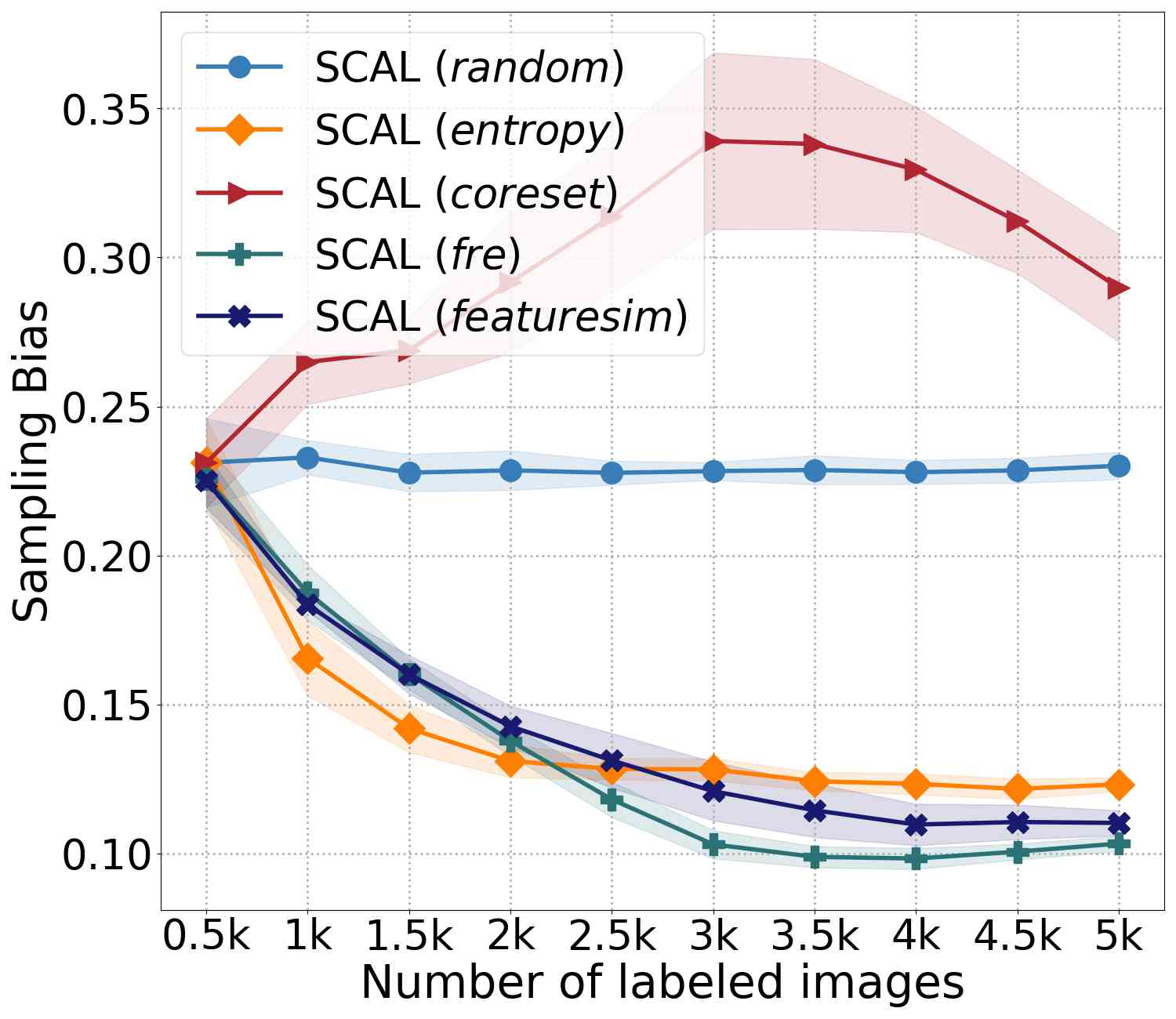}}
		\subfigure[ECE {$\downarrow$}]{\includegraphics[width=0.51\columnwidth]{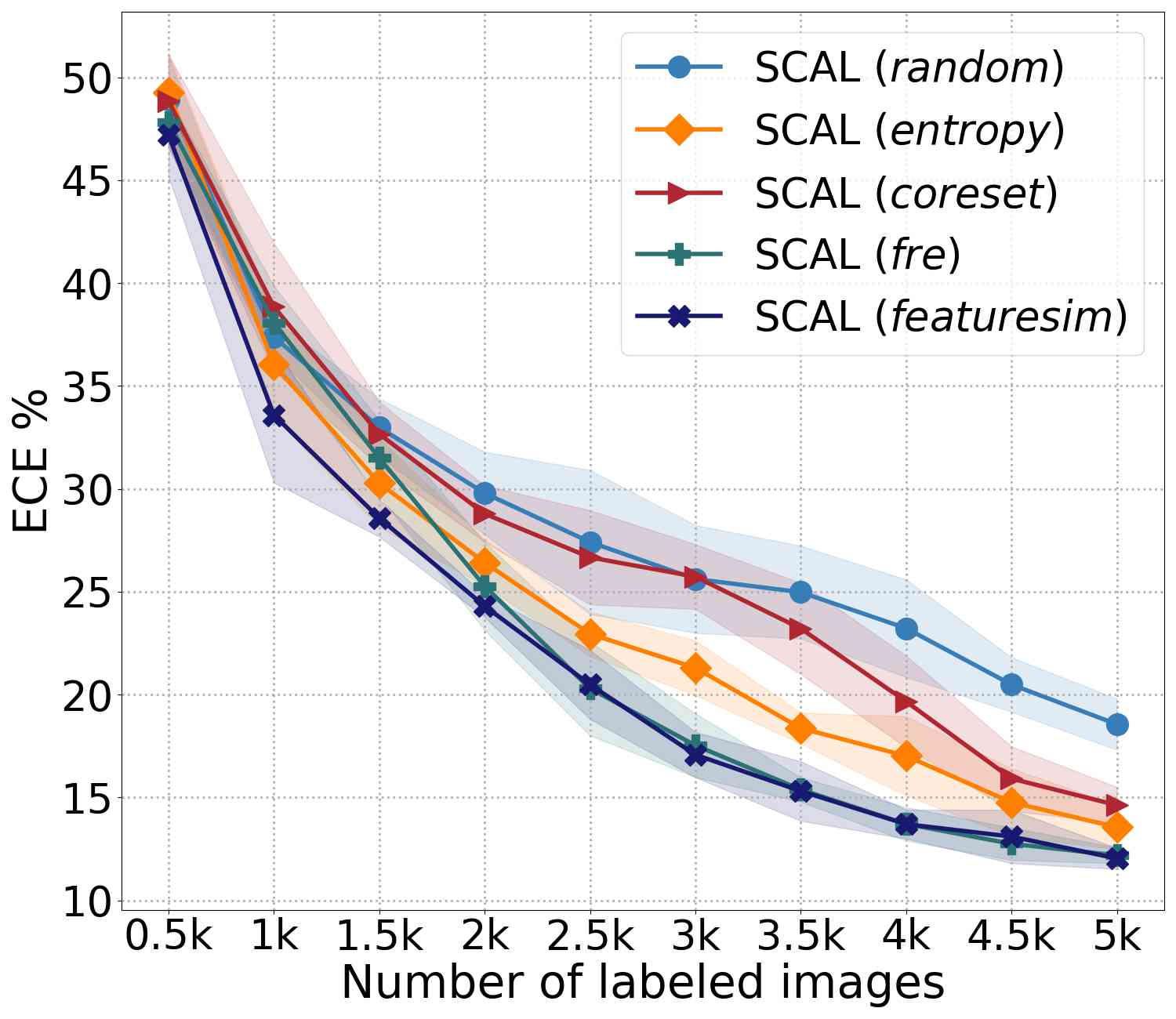}}
		\caption{\small Imbalanced-CIFAR10/ResNet-18: Ablation study of different query functions with models trained with \textbf{contrastive loss}}
		\label{fig:ablation_imbalanced_cifar10}
\end{figure*}
\begin{figure*}
		\centering     
		\subfigure[Accuracy {$\uparrow$}]{\includegraphics[width=0.51\columnwidth]{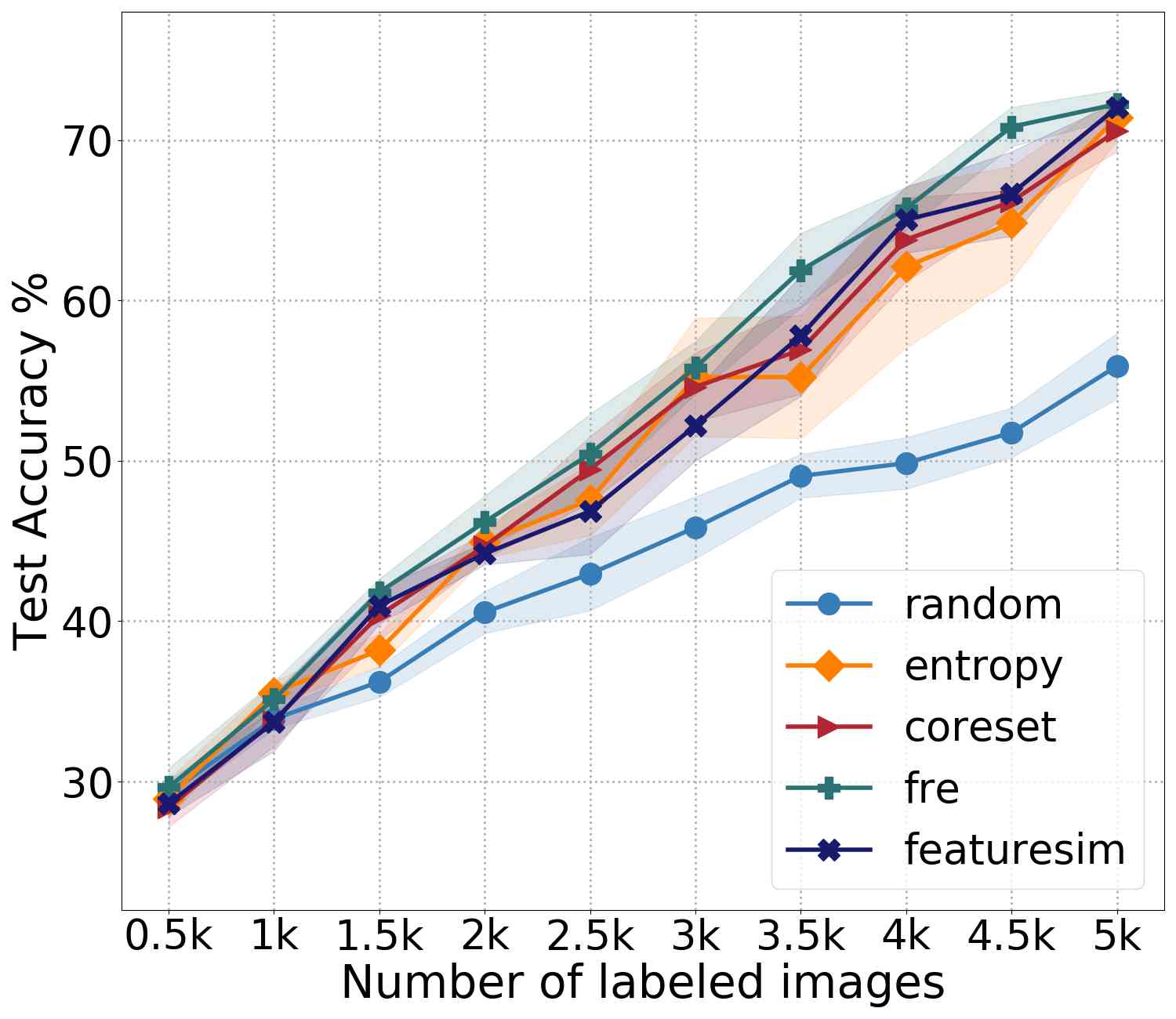}}
		\subfigure[Negative log-likelihood {$\downarrow$}]{\includegraphics[width=0.51\columnwidth]{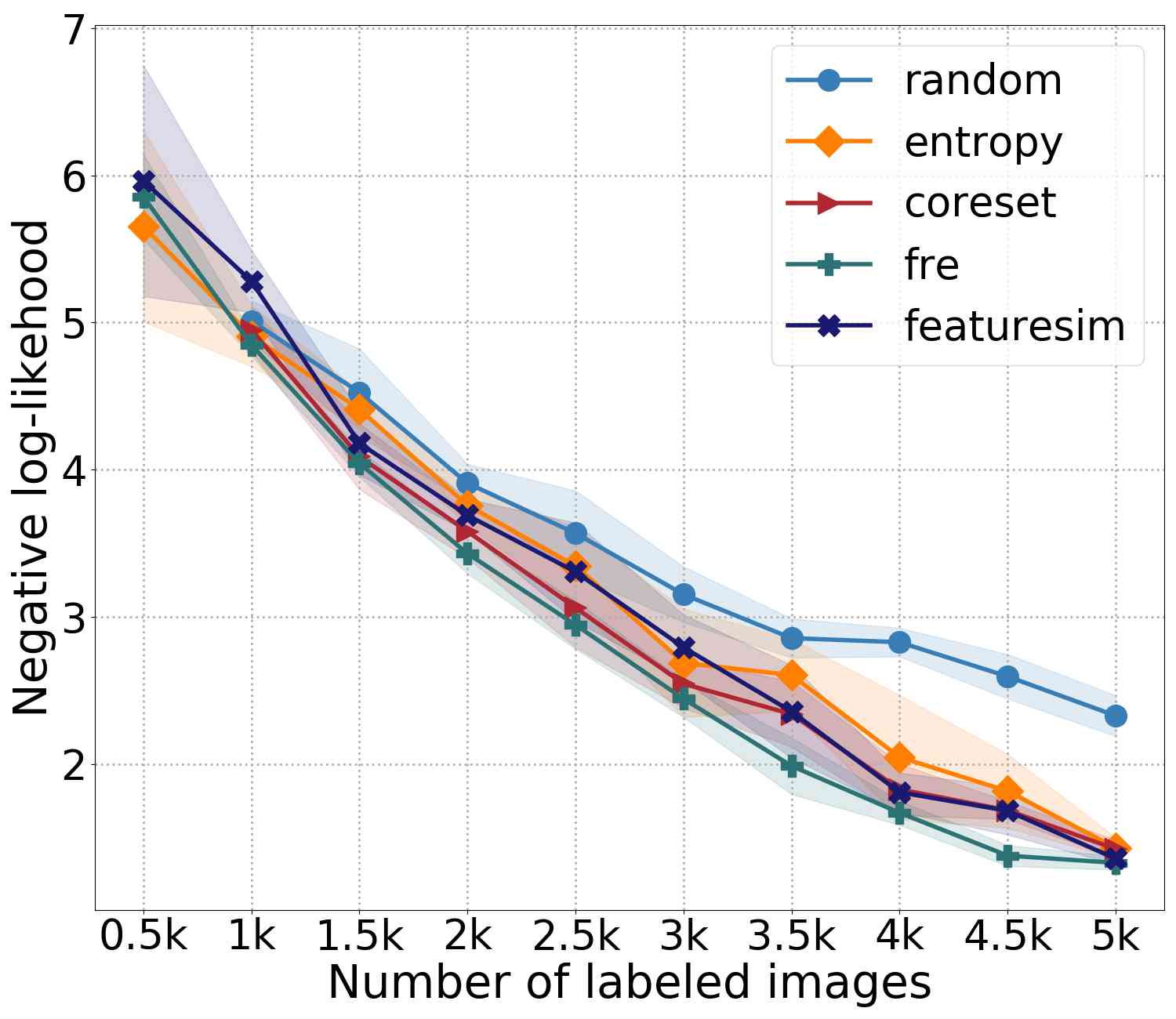}}
		\subfigure[Sampling Bias {$\downarrow$}]{\includegraphics[width=0.51\columnwidth]{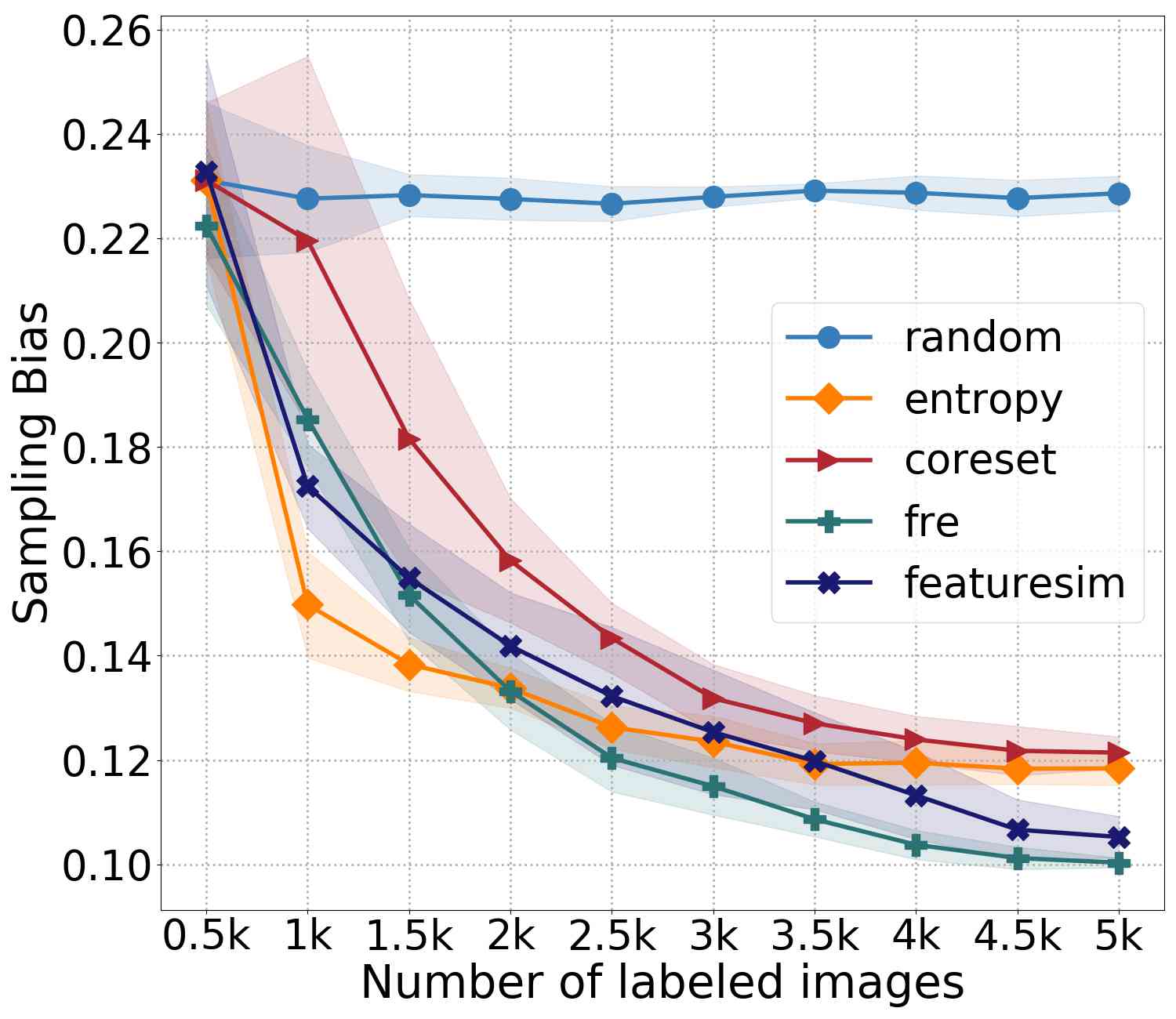}}
		\subfigure[ECE {$\downarrow$}]{\includegraphics[width=0.51\columnwidth]{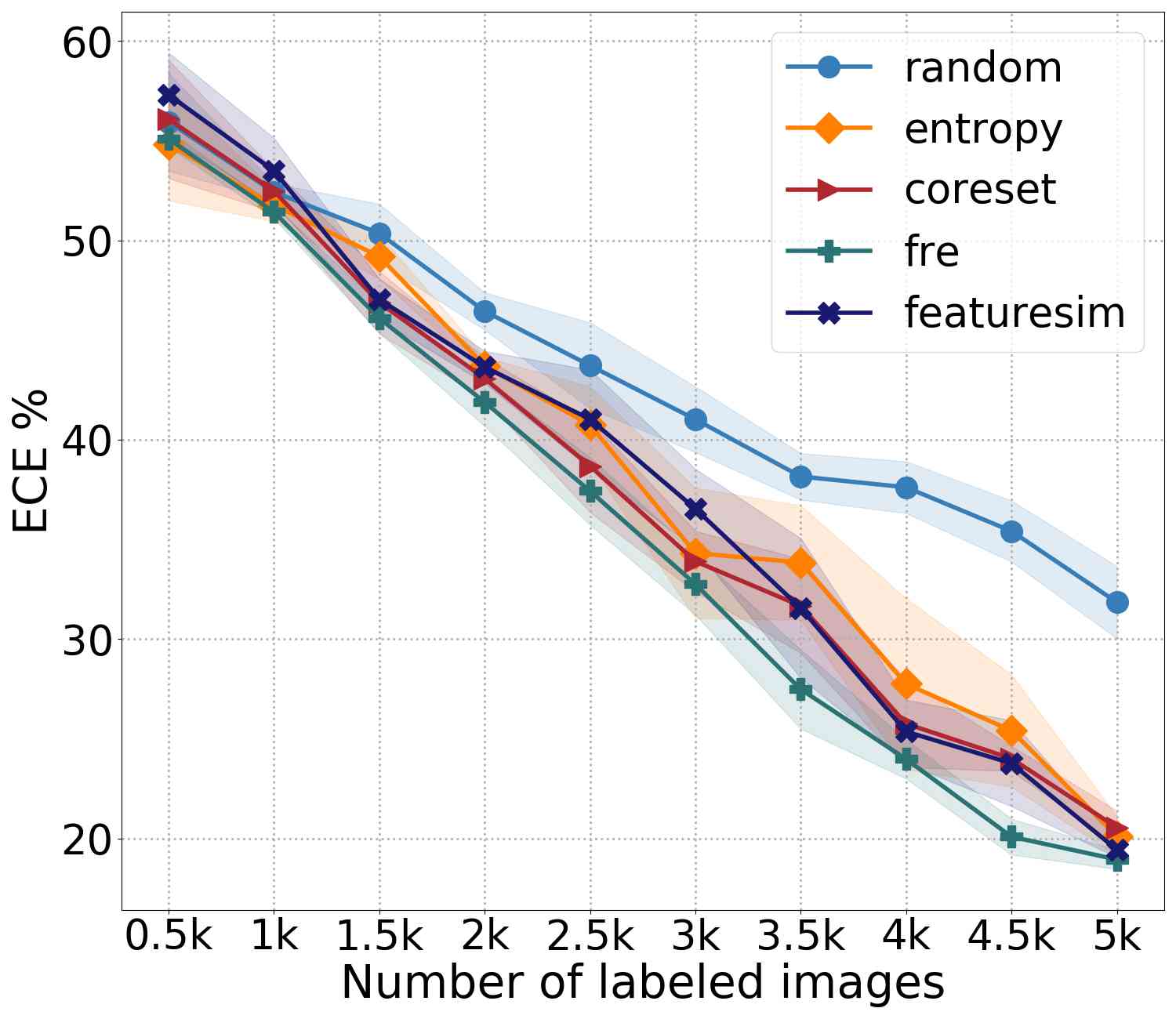}}
		\caption{\small Imbalanced-CIFAR10/ResNet-18: Ablation study of different query functions with models trained with \textbf{cross-entropy loss}}
		\label{fig:ablation_imbalanced_cifar10_ce}
\end{figure*}

\section{Additional Results}

\subsection{Ablation study: Effect of different query methods and loss functions}
\label{appdx:ablation}
We performed ablation study to understand the effect of proposed query functions $\textit{featuresim}$ and $\textit{fre}$, and training the model with contrastive versus cross-entropy loss in active learning setting. We compared the query methods including Random, Entropy, CoreSet, $\textit{featuresim}$ and $\textit{fre}$ while training the model with contrastive loss and cross-entropy loss separately. We performed these ablation studies on CIFAR10 and Imbalanced-CIFAR10 datasets.

We find that contrastive loss helps in learning better feature representation and improving robustness under distributional shift, while the proposed query functions help in selecting unbiased and diverse samples that guides the contrastive loss to learn feature representations from the most informative samples. Figures~\ref{fig:ablation_cifar10}-(c), \ref{fig:ablation_cifar10_ce}-(c), \ref{fig:ablation_imbalanced_cifar10}-(c) and \ref{fig:ablation_imbalanced_cifar10_ce}-(c) show that the proposed sample selection scoring functions $\textit{featuresim}$ and $\textit{fre}$ helps in reducing the sampling bias while learning from both balanced and imbalanced datasets, irrespective of contrastive or cross-entropy loss. Fig.~\ref{fig:ablation_cifar10}-(d) and Fig.~\ref{fig:ablation_cifar10_ce}-(d) show that contrastive loss helps in improved robustness to distributional shift (higher AUROC for out-of-distribution detection). We observe CoreSet approach that performs well with cross-entropy loss, suffers when trained with contrastive loss resulting in higher sampling bias (as seen in Fig.~\ref{fig:ablation_cifar10}-(c) and Fig.~\ref{fig:ablation_imbalanced_cifar10}-(c)) and lower accuracy (as seen in Fig.~\ref{fig:ablation_cifar10}-(a) and Fig.~\ref{fig:ablation_imbalanced_cifar10}-(a)). Figures~\ref{fig:ablation_cifar10}-(a),(b) and \ref{fig:ablation_imbalanced_cifar10}-(a),(b) show combining the contrastive loss and query functions $\textit{featuresim}$/$\textit{fre}$ benefits in active learning setting for imbalanced and balanced datasets, supporting the results presented in Section~\ref{sec:results}. 

\subsection{Qualitative analysis using tSNE (t-distributed Stochastic Neighbor Embedding)}
\label{appdx:tsne}
\begin{figure*}
		\centering     
		\subfigure[Random]{\includegraphics[width=0.4\columnwidth]{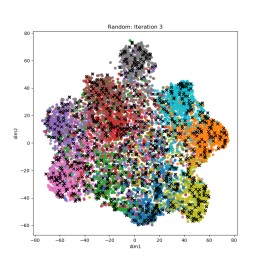}}
		\subfigure[Entropy]{\includegraphics[width=0.4\columnwidth]{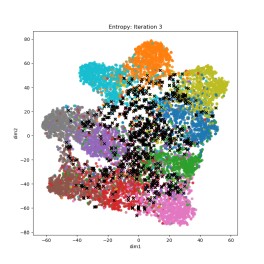}}
		\subfigure[CoreSet]{\includegraphics[width=0.4\columnwidth]{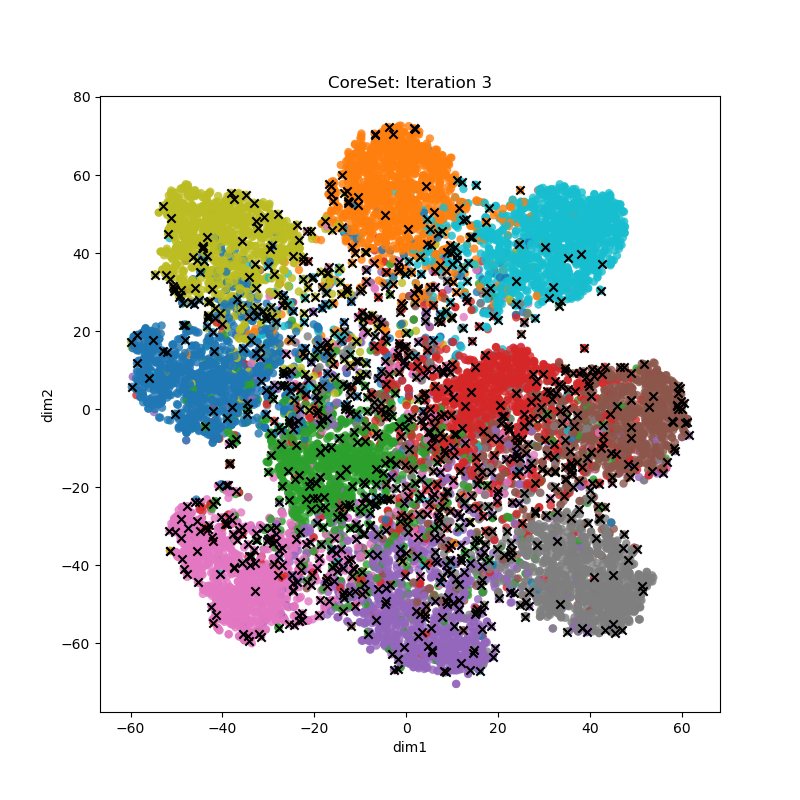}}
		\subfigure[BALD]{\includegraphics[width=0.41\columnwidth]{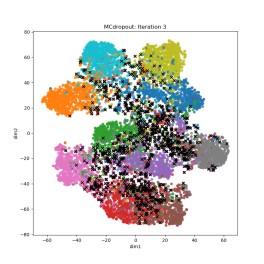}}
		\subfigure[SCAL]{\includegraphics[width=0.4\columnwidth]{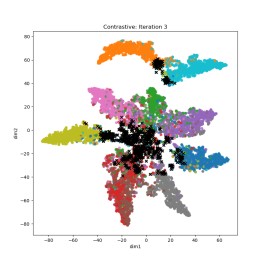}}
		\caption{Active learning Iteration 3: tSNE plots for CIFAR-10/ResNet-18}
		\label{fig:tsne_3}
\end{figure*}

\begin{figure*}
		\centering     
		\subfigure[Random]{\includegraphics[width=0.4\columnwidth]{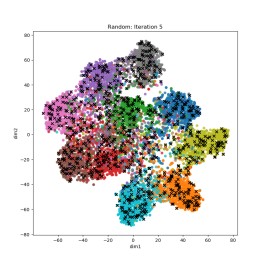}}
		\subfigure[Entropy]{\includegraphics[width=0.4\columnwidth]{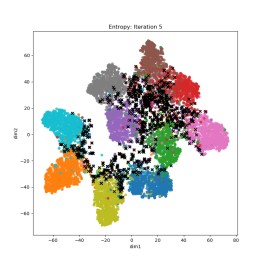}}
		\subfigure[CoreSet]{\includegraphics[width=0.4\columnwidth]{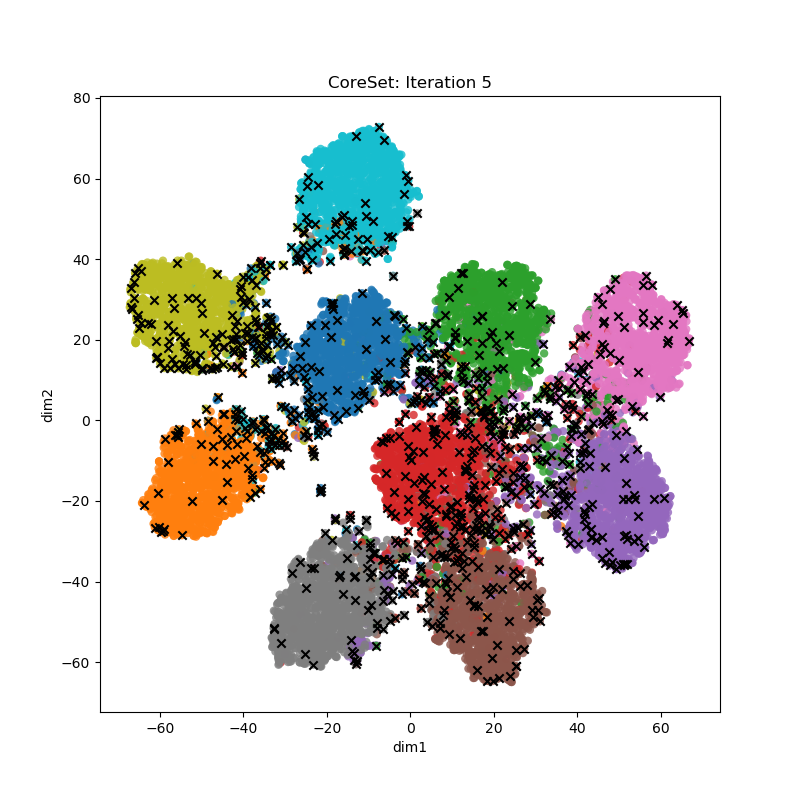}}
		\subfigure[BALD]{\includegraphics[width=0.41\columnwidth]{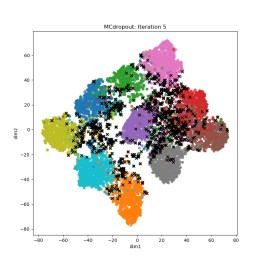}}
		\subfigure[SCAL]{\includegraphics[width=0.4\columnwidth]{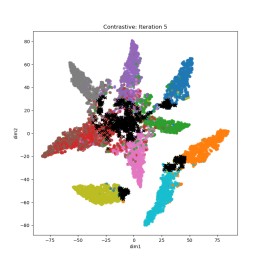}}
		\caption{Active learning Iteration 5: tSNE plots for CIFAR-10/ResNet-18}
		\label{fig:tsne_5}
\end{figure*}

\begin{figure*}
		\centering     
		\subfigure[Random]{\includegraphics[width=0.4\columnwidth]{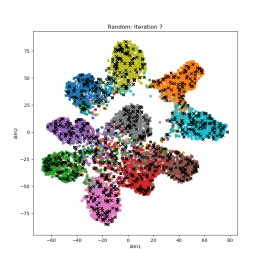}}
		\subfigure[Entropy]{\includegraphics[width=0.4\columnwidth]{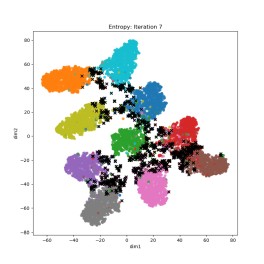}}
		\subfigure[CoreSet]{\includegraphics[width=0.4\columnwidth]{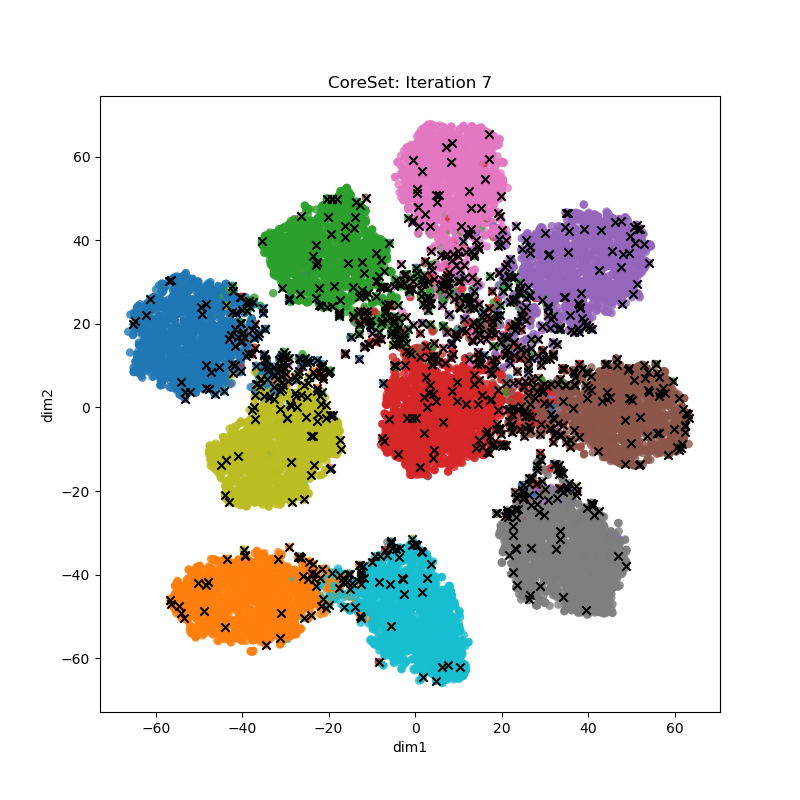}}
		\subfigure[BALD]{\includegraphics[width=0.41\columnwidth]{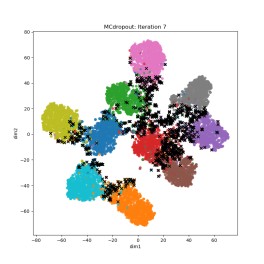}}
		\subfigure[SCAL]{\includegraphics[width=0.4\columnwidth]{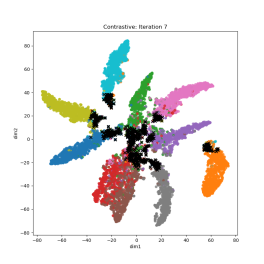}}
		\caption{Active learning Iteration 7: tSNE plots for CIFAR-10/ResNet-18}
		\label{fig:tsne_7}
\end{figure*}

\begin{figure*}
		\centering     
		\subfigure[Random]{\includegraphics[width=0.4\columnwidth]{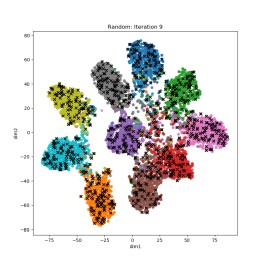}}
		\subfigure[Entropy]{\includegraphics[width=0.4\columnwidth]{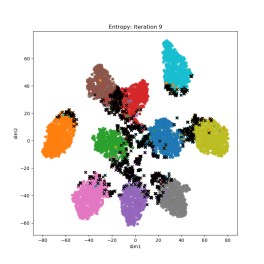}}
		\subfigure[CoreSet]{\includegraphics[width=0.4\columnwidth]{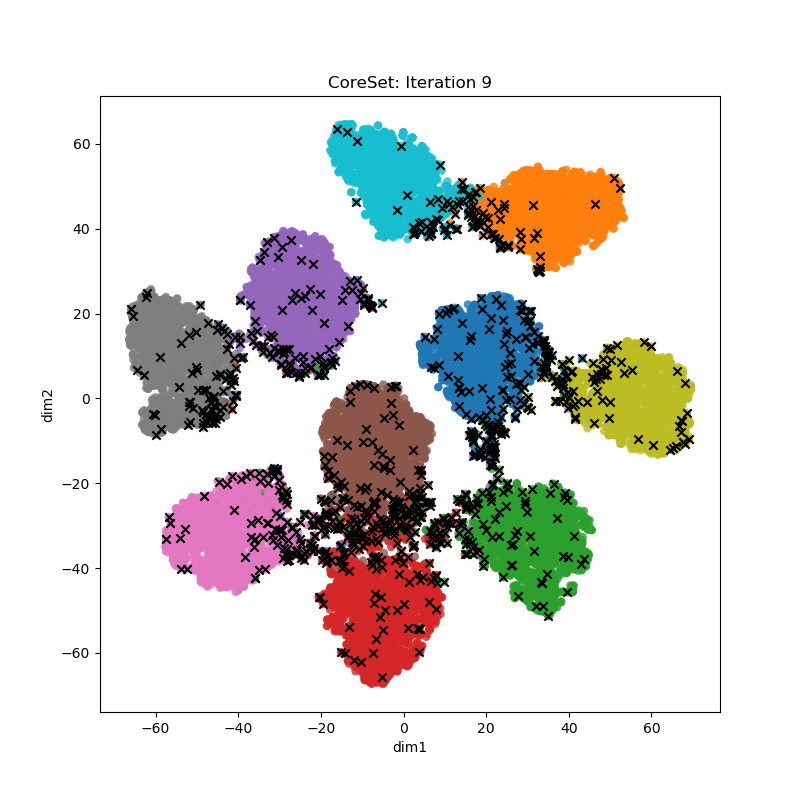}}
		\subfigure[BALD]{\includegraphics[width=0.41\columnwidth]{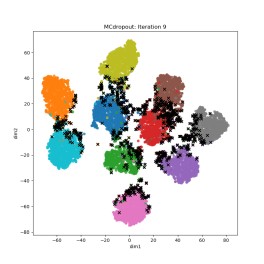}}
		\subfigure[SCAL]{\includegraphics[width=0.4\columnwidth]{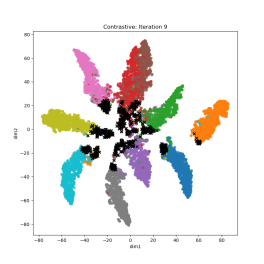}}
		\caption{Active learning Iteration 9: tSNE plots for CIFAR-10/ResNet-18}
		\label{fig:tsne_9}
\end{figure*}

We study the feature representations obtained from different methods with t-SNE~\cite{van2008visualizing} embeddings. Figures~\ref{fig:tsne_3}-\ref{fig:tsne_9} show the feature embeddings from different methods and sample selection exploration in active learning. The plots visualize the data samples corresponding to 10 classes with 10 different colors and the selected data with black markers, which is considered as most informative samples by the query strategies of corresponding methods. We notice that the features of SCAL method is well clustered even in the initial iterations of active learning as compared to other methods, this justifies the higher accuracy and lower expected calibration error for SCAL.

\begin{figure*}
		\centering     
		\subfigure[Entropy]{\includegraphics[width=0.45\columnwidth]{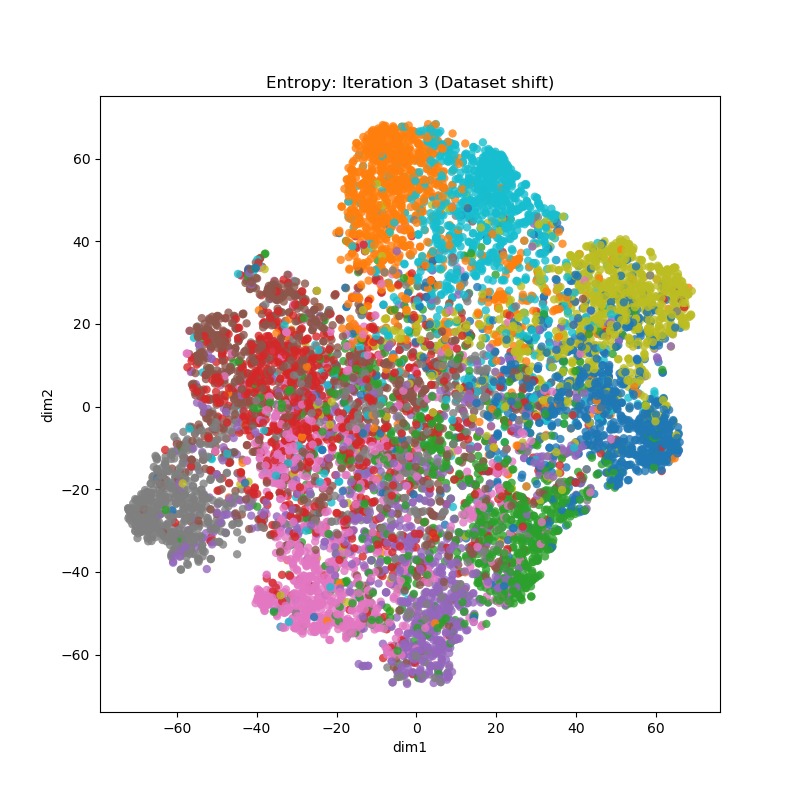}}
		\subfigure[Learning Loss]{\includegraphics[width=0.45\columnwidth]{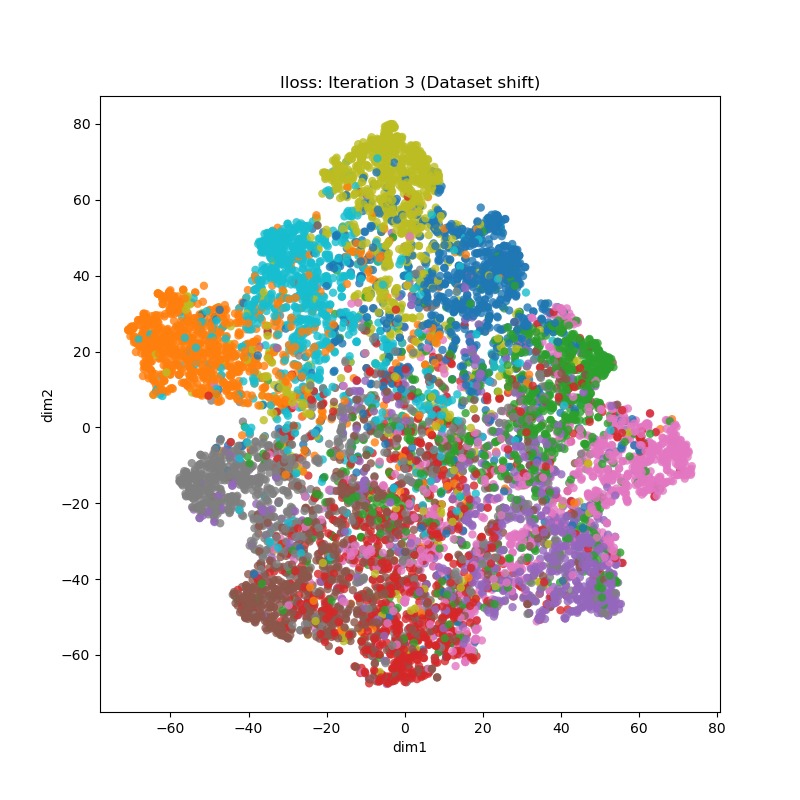}}
		\subfigure[CoreSet]{\includegraphics[width=0.45\columnwidth]{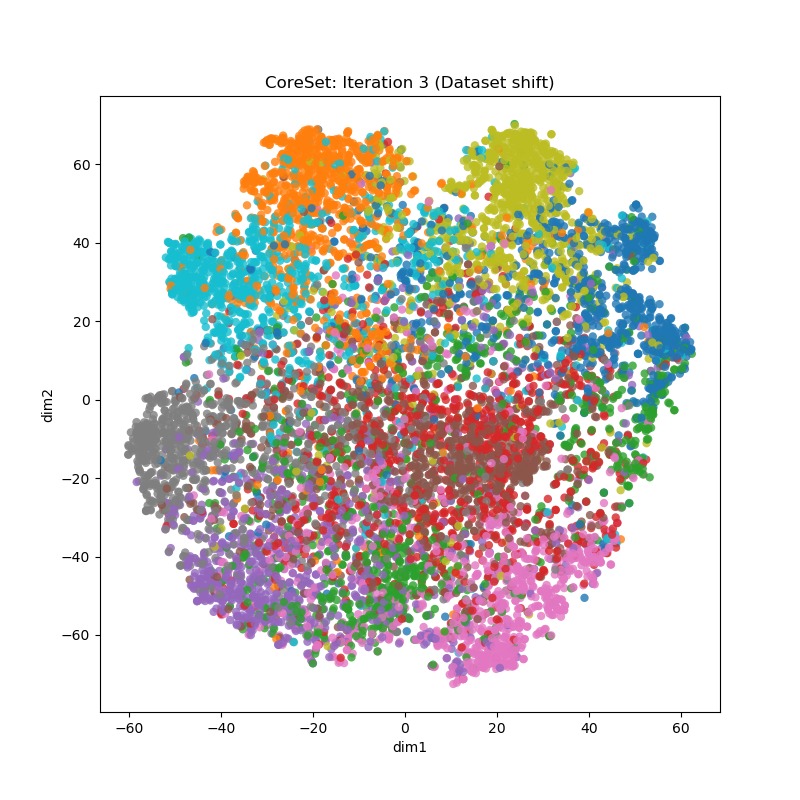}}
		\subfigure[SCAL]{\includegraphics[width=0.45\columnwidth]{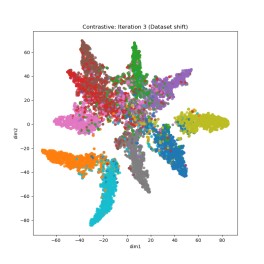}}
		\caption{tSNE plots for dataset shift (CIFAR10 corrupted with Gaussian blur) - Active learning Iteration 3}
		\label{fig:tsne_shift_3}
\end{figure*}

\begin{figure*}
		\centering     
		\subfigure[Entropy]{\includegraphics[width=0.45\columnwidth]{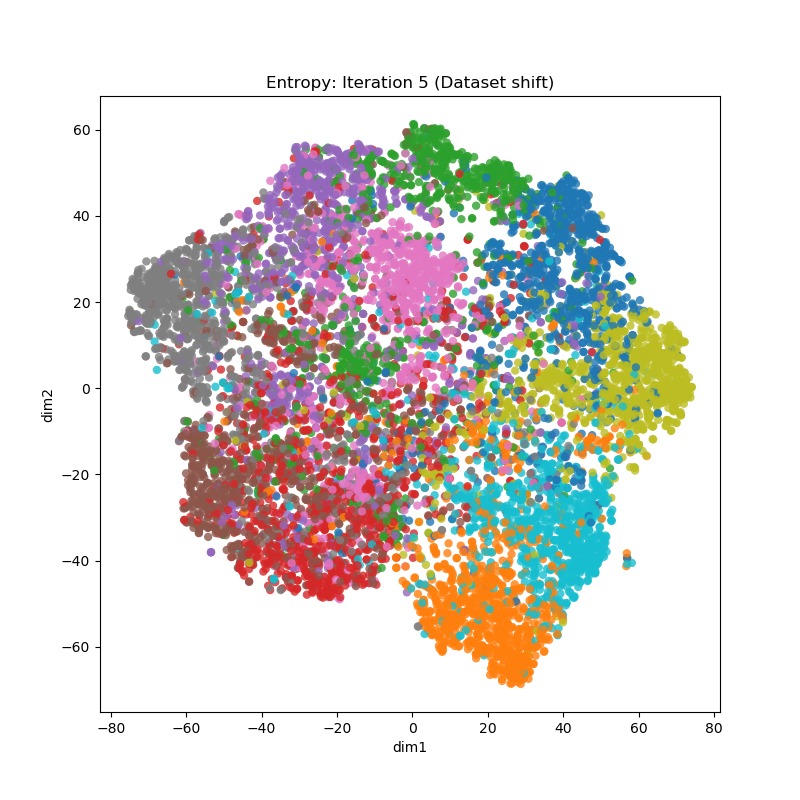}}
		\subfigure[Learning Loss]{\includegraphics[width=0.45\columnwidth]{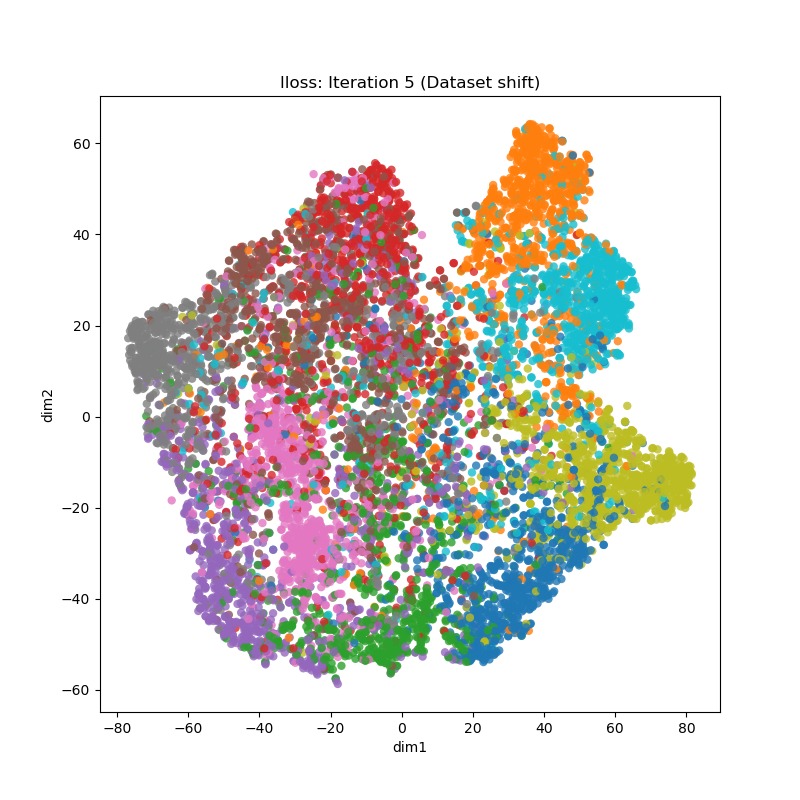}}
		\subfigure[CoreSet]{\includegraphics[width=0.45\columnwidth]{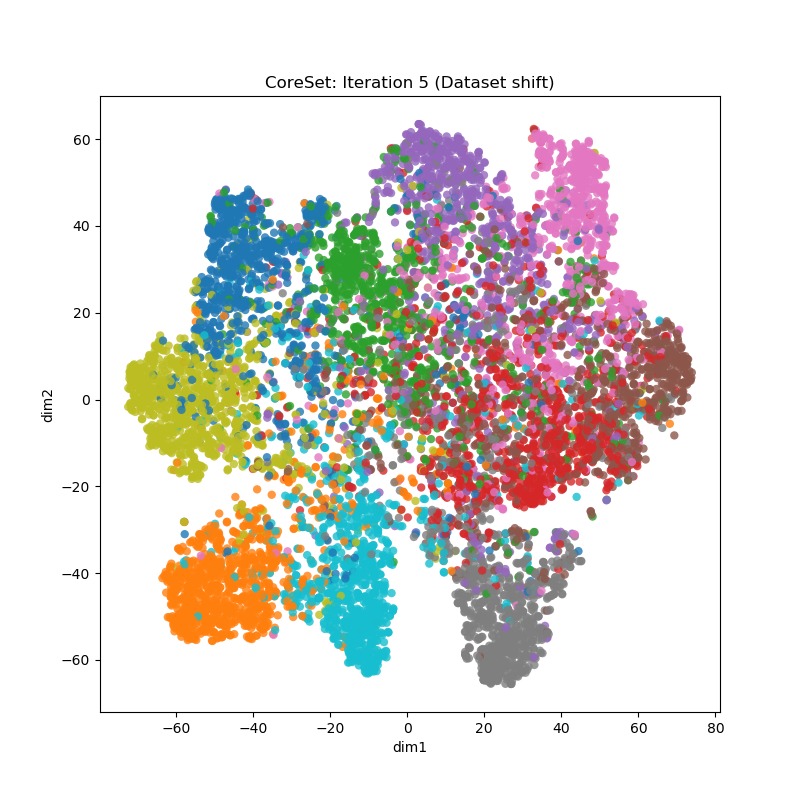}}
		\subfigure[SCAL]{\includegraphics[width=0.45\columnwidth]{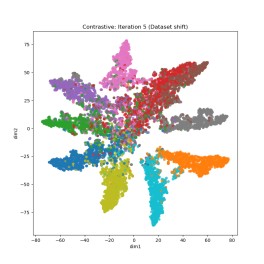}}
		\caption{tSNE plots for dataset shift (CIFAR10 corrupted with Gaussian blur) - Active learning Iteration 5}
		\label{fig:tsne_shift_5}
\end{figure*}

\begin{figure*}
		\centering     
		\subfigure[Entropy]{\includegraphics[width=0.45\columnwidth]{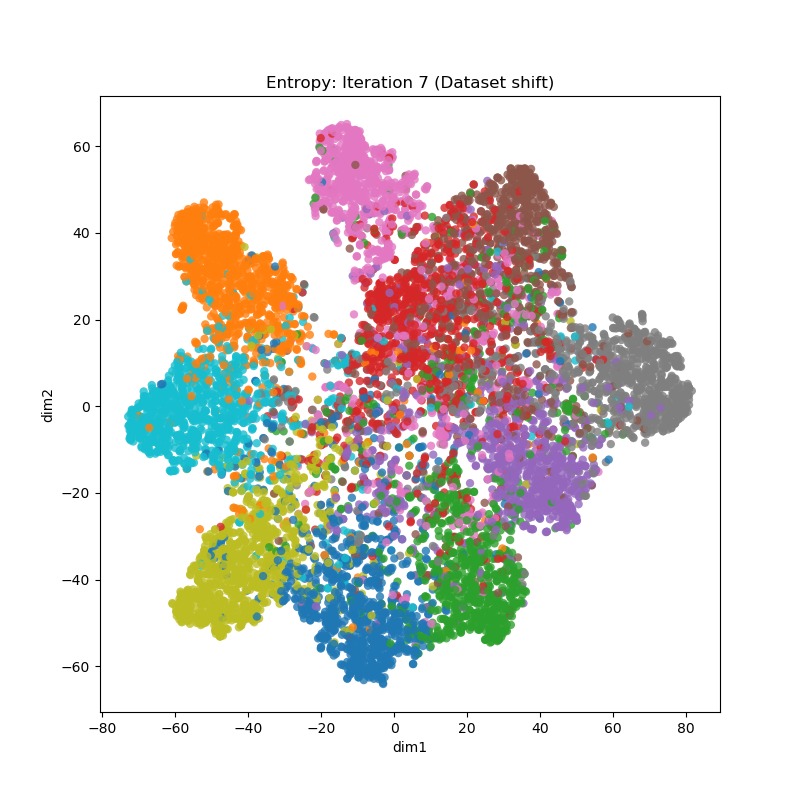}}
		\subfigure[Learning Loss]{\includegraphics[width=0.45\columnwidth]{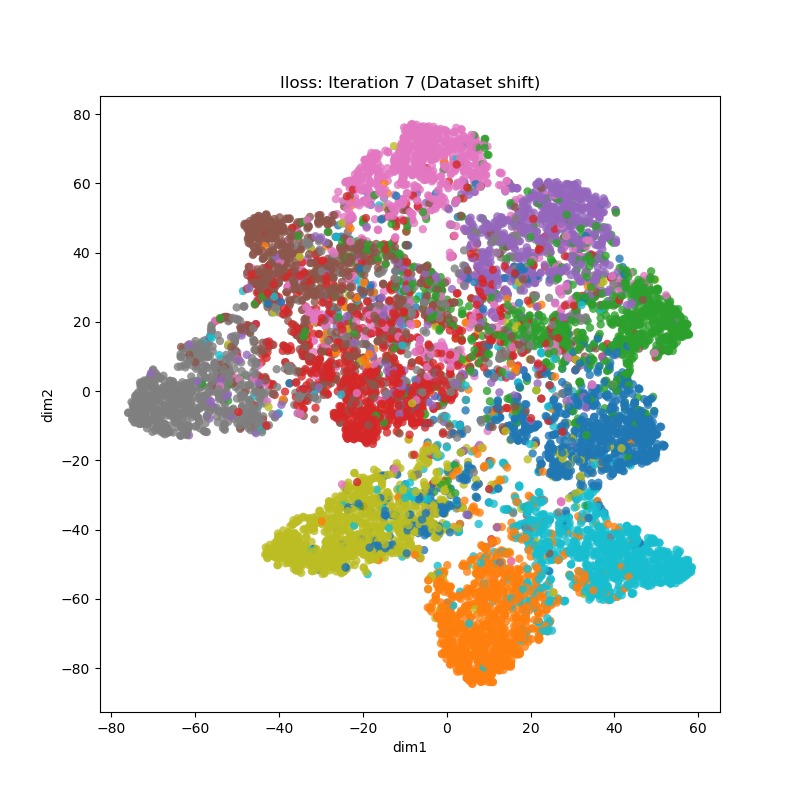}}
		\subfigure[CoreSet]{\includegraphics[width=0.45\columnwidth]{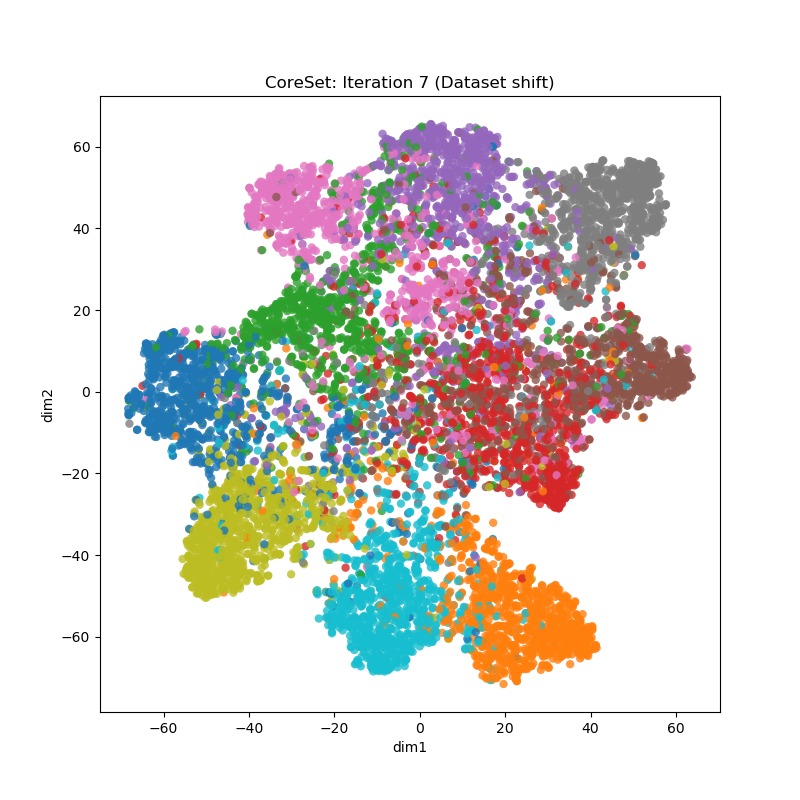}}
		\subfigure[SCAL]{\includegraphics[width=0.45\columnwidth]{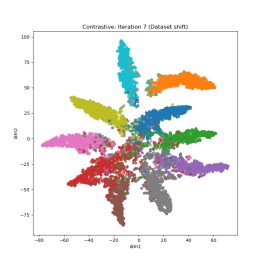}}
		\caption{tSNE plots for dataset shift (CIFAR10 corrupted with Gaussian blur) - Active learning Iteration 7}
		\label{fig:tsne_shift_7}
\end{figure*}

\begin{figure*}
		\centering     
		\subfigure[Entropy]{\includegraphics[width=0.45\columnwidth]{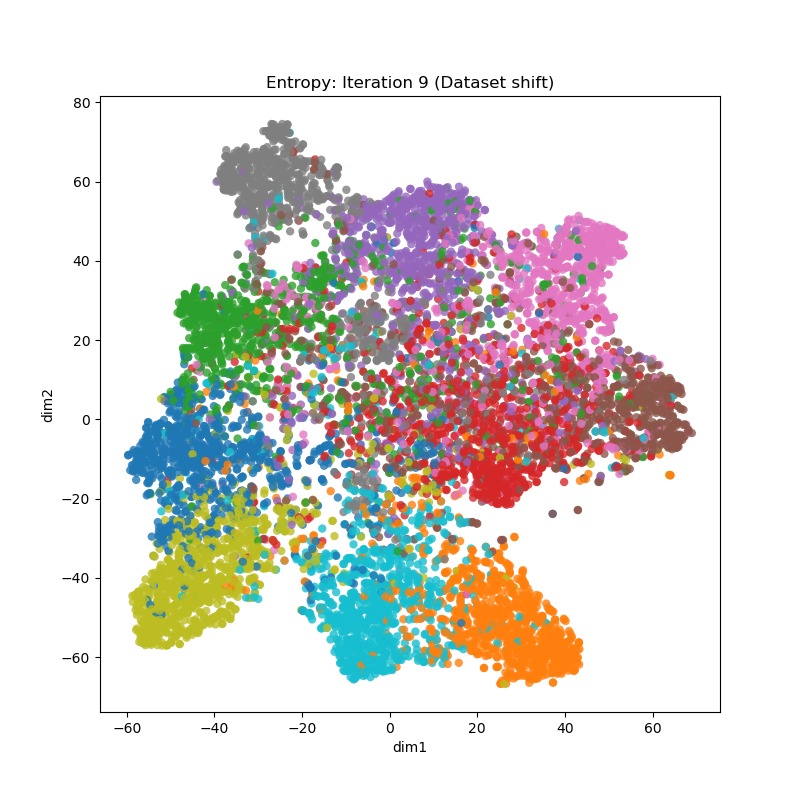}}
		\subfigure[Learning Loss]{\includegraphics[width=0.45\columnwidth]{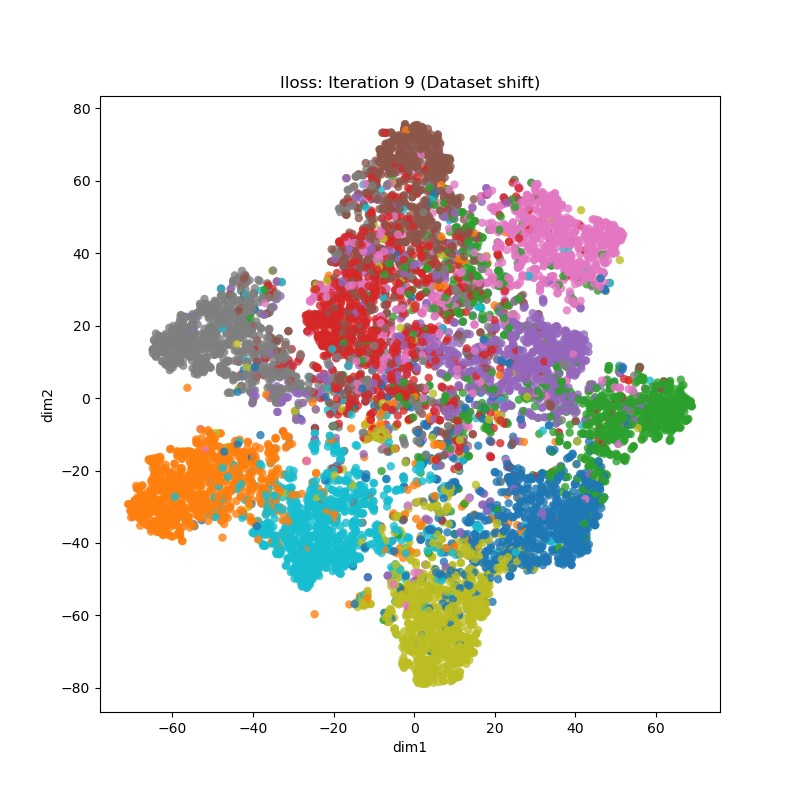}}
		\subfigure[CoreSet]{\includegraphics[width=0.45\columnwidth]{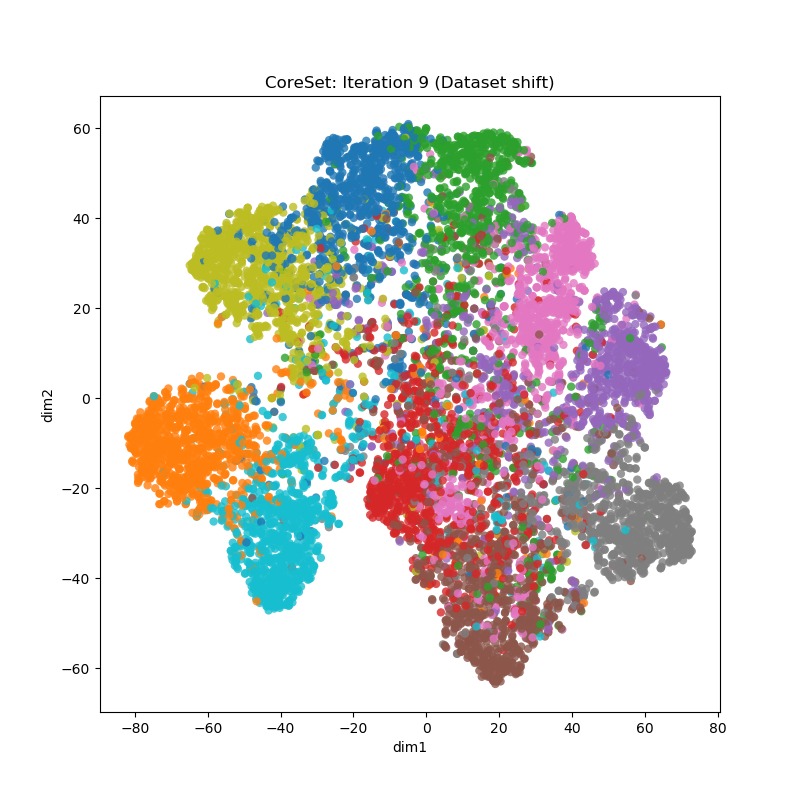}}
		\subfigure[SCAL]{\includegraphics[width=0.45\columnwidth]{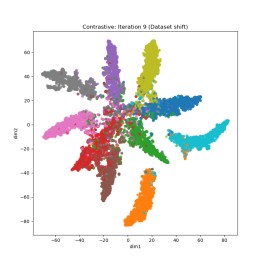}}
		\caption{tSNE plots for dataset shift (CIFAR10 corrupted with Gaussian blur) - Active learning Iteration 9}
		\label{fig:tsne_shift_9}
\end{figure*}

We further investigate the superior performance of SCAL under dataset shift (Fig.~\ref{fig:robustness}) with t-SNE embeddings. We compute the feature embeddings for the shifted data (CIFAR-10 corrupted with Gaussian Blur~\cite{hendrycks2019benchmarking}) at each active learning iteration after the models are trained with clean labeled set. Figures~\ref{fig:tsne_shift_3}-\ref{fig:tsne_shift_9} show the feature embeddings of SCAL method is well clustered and well separated between classes even under dataset shift, as compared to other methods. 

\subsection{Model calibration and reliability}
\label{appdx:brier}
\begin{figure*}
		\centering     
		\subfigure[Imbalanced-CIFAR10]{\includegraphics[width=0.65\columnwidth]{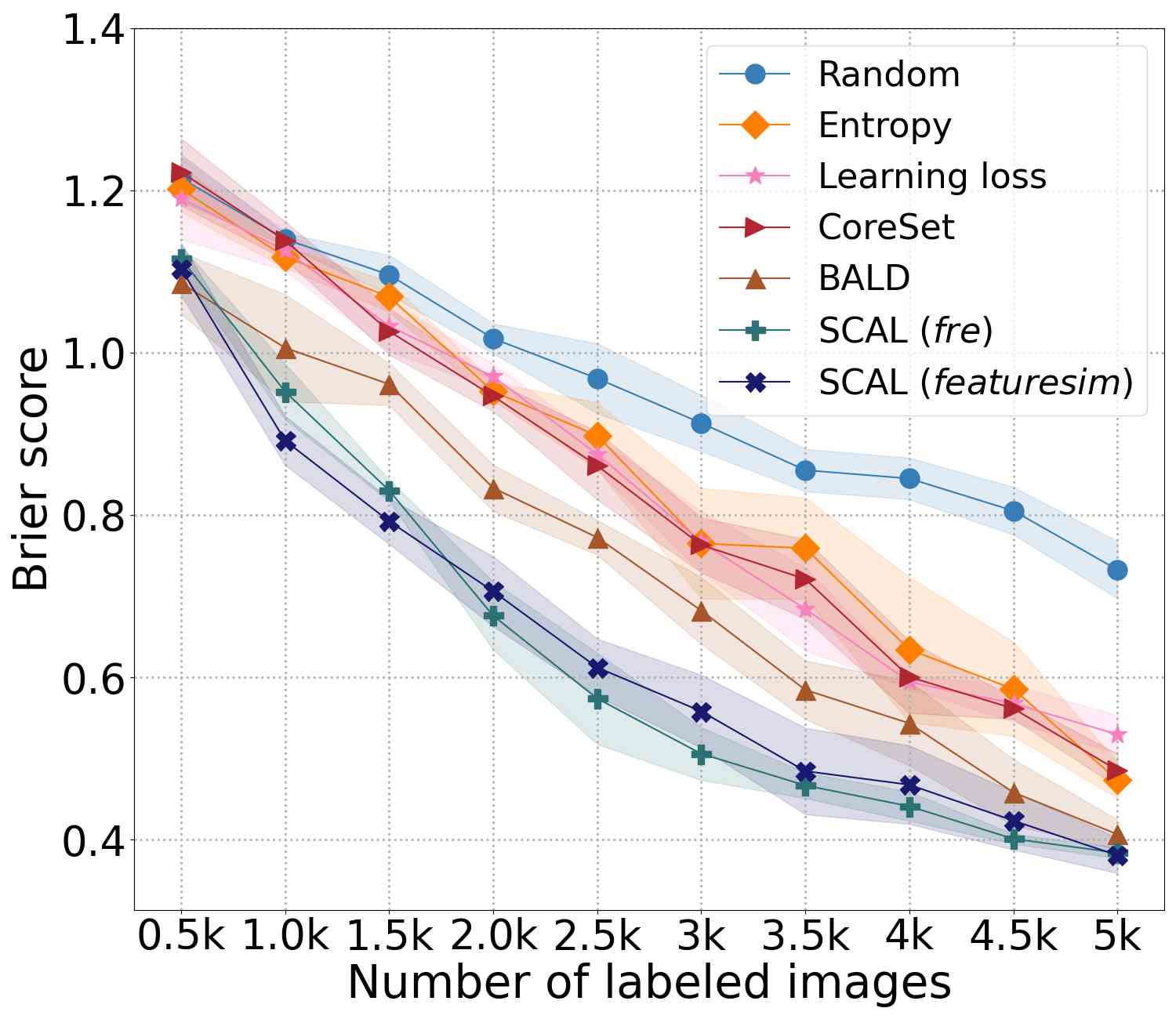}}
		\subfigure[CIFAR10]{\includegraphics[width=0.65\columnwidth]{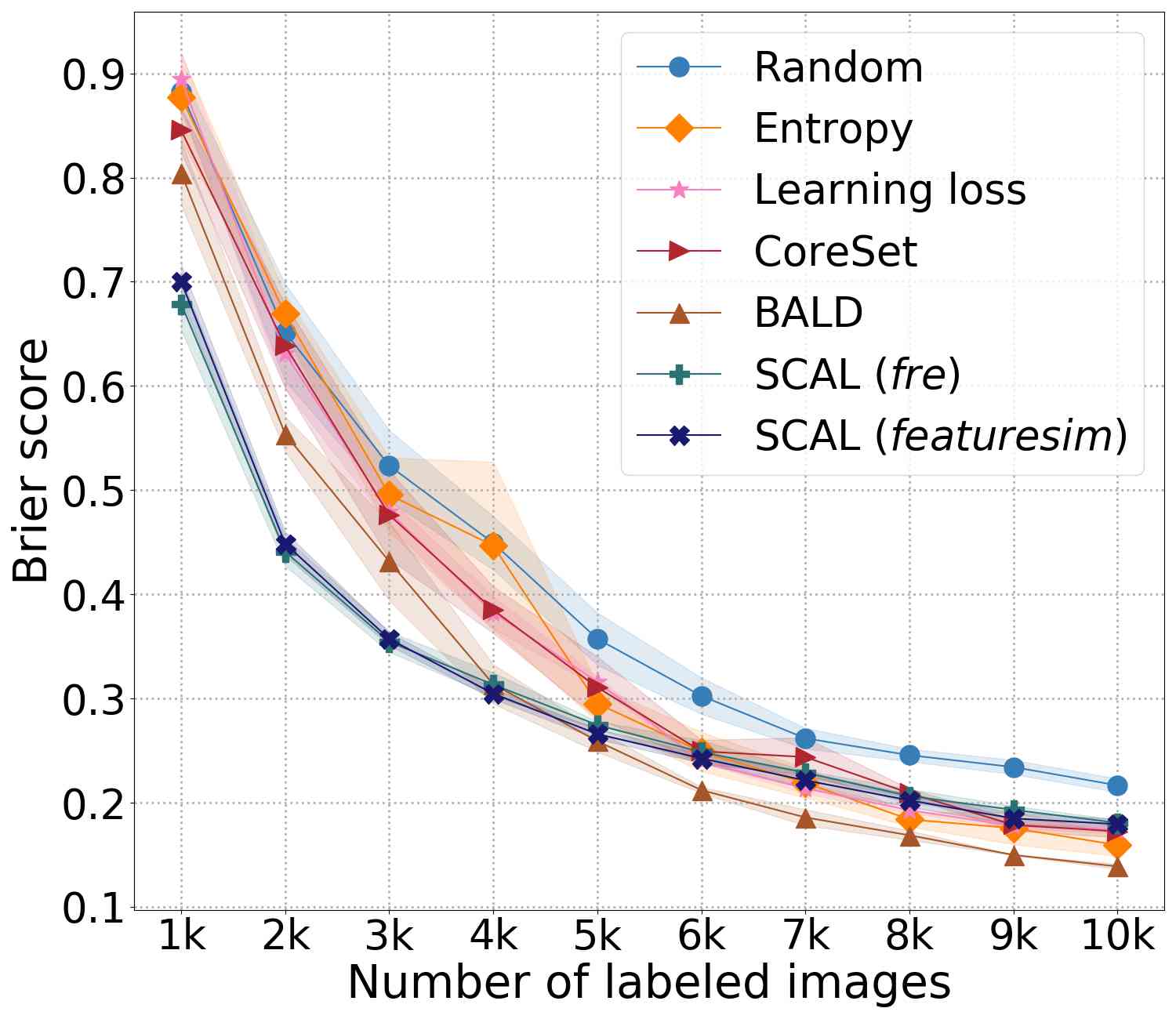}}
		\subfigure[CIFAR10-C ]{\includegraphics[width=0.65\columnwidth]{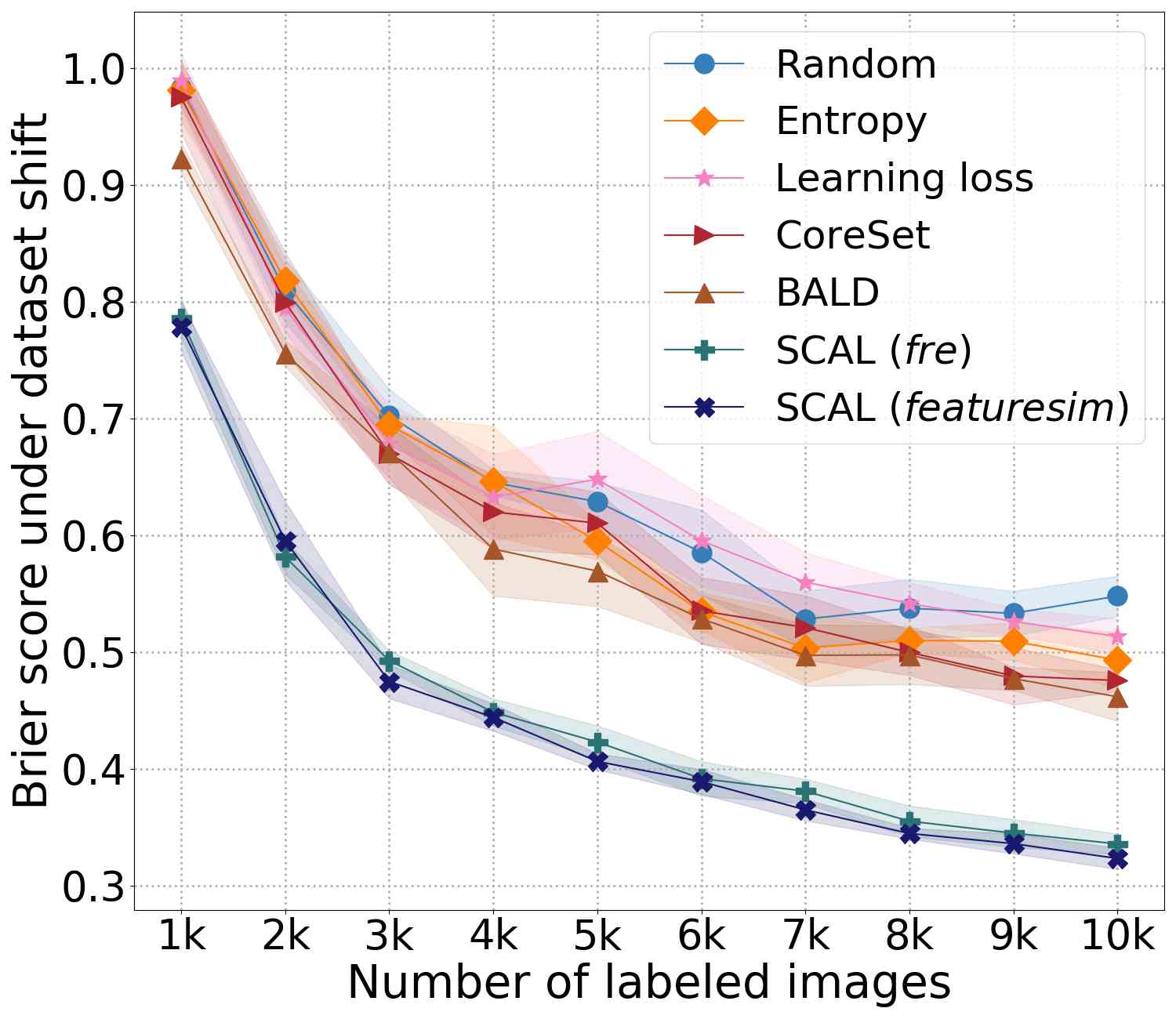}}
		\caption{Brier score (lower is better) evaluation of different query methods on Imbalanced-CIFAR10, CIFAR10 and CIFAR10-C datasets in active learning. The shading shows std-dev from 5 independent trials for each method. We present the results as (1+Brier) for easier readability. Lower Brier score indicates the model is well-calibrated.}
		\label{fig:brier}
\end{figure*}

\begin{figure*}
	\small
	\begin{subfigure}
		\centering
		\captionsetup{
			justification=centering}
		\includegraphics[scale=0.75]{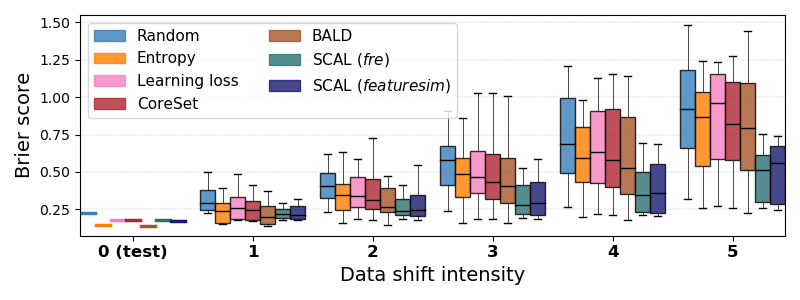}
	\end{subfigure}
	\caption{Brier score (lower is better) under dataset shift. Comparison of models derived with different query methods in active learning setting (10th iteration). At each shift intensity level, the boxplot summarizes the Brier score across 16 different datashift types showing the min, max and quartiles. Our proposed \textbf{SCAL} consistently yields lower Brier score even under increased dataset shift intensity, demonstrating better robustness compared to other state-of-the-art methods. }
	\label{fig:boxplots_appendix}
\end{figure*}

In addition to Expected Calibration Error (ECE)~\cite{naeini2015obtaining} presented in Fig.~\ref{fig:ece}, we evaluate model calibration and robustness using Brier score~\cite{brier1950verification}. Fig.~\ref{fig:brier} compares the active learning methods on Imbalanced-CIFAR10, CIFAR10 and CIFAR10-C with Brier score, which is a proper scoring rule~\cite{gneiting2007strictly} for evaluating the model calibration. We present the results with (1+Brier) for easier readability. Fig.~\ref{fig:boxplots_appendix} show the comparison under dataset shift at different intensity levels.

\subsection{Model accuracy when trained with full dataset}
The test accuracy when the model is trained using entire training set with cross entropy (Imbalanced-CIFAR10: 79.53\%, CIFAR10: 93.04\%, SVHN: 95.93\%), or contrastive (Imbalanced-CIFAR10: 79.36\%, CIFAR10: 92.69\%, SVHN: 95.54\%) loss indicates the upper bound on the accuracy that can be achieved in active learning setting without labeling 100\% of training set. The results are provided in Appendix Table~\ref{tab:totscores}.

\subsection{Full learning curve}
\begin{figure*}
  \centering
\subfigure[Test Accuracy ($\uparrow$)]{\includegraphics[width=0.7\textwidth]{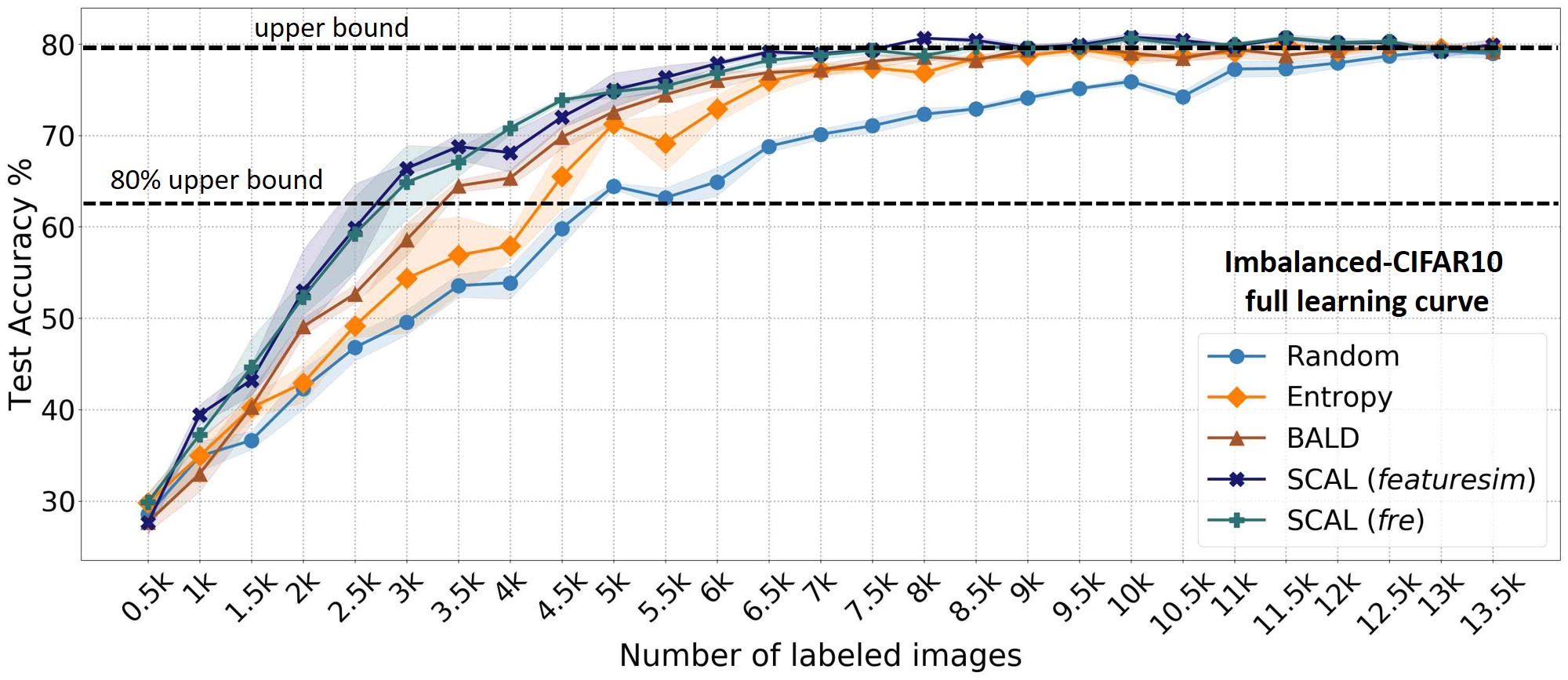}}
\subfigure[ECE ($\downarrow$)]{\includegraphics[width=0.7\textwidth]{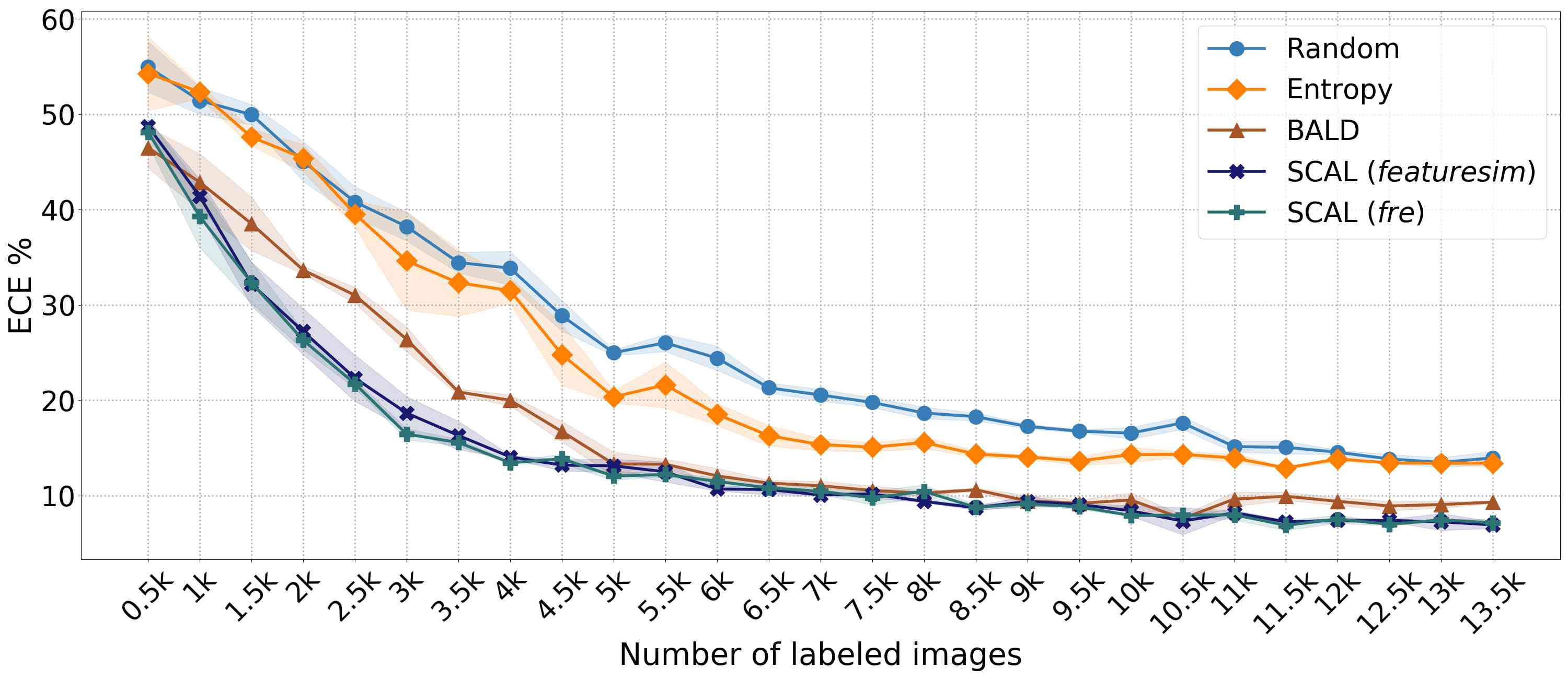}}
\subfigure[NLL ($\downarrow$)]{\includegraphics[width=0.7\textwidth]{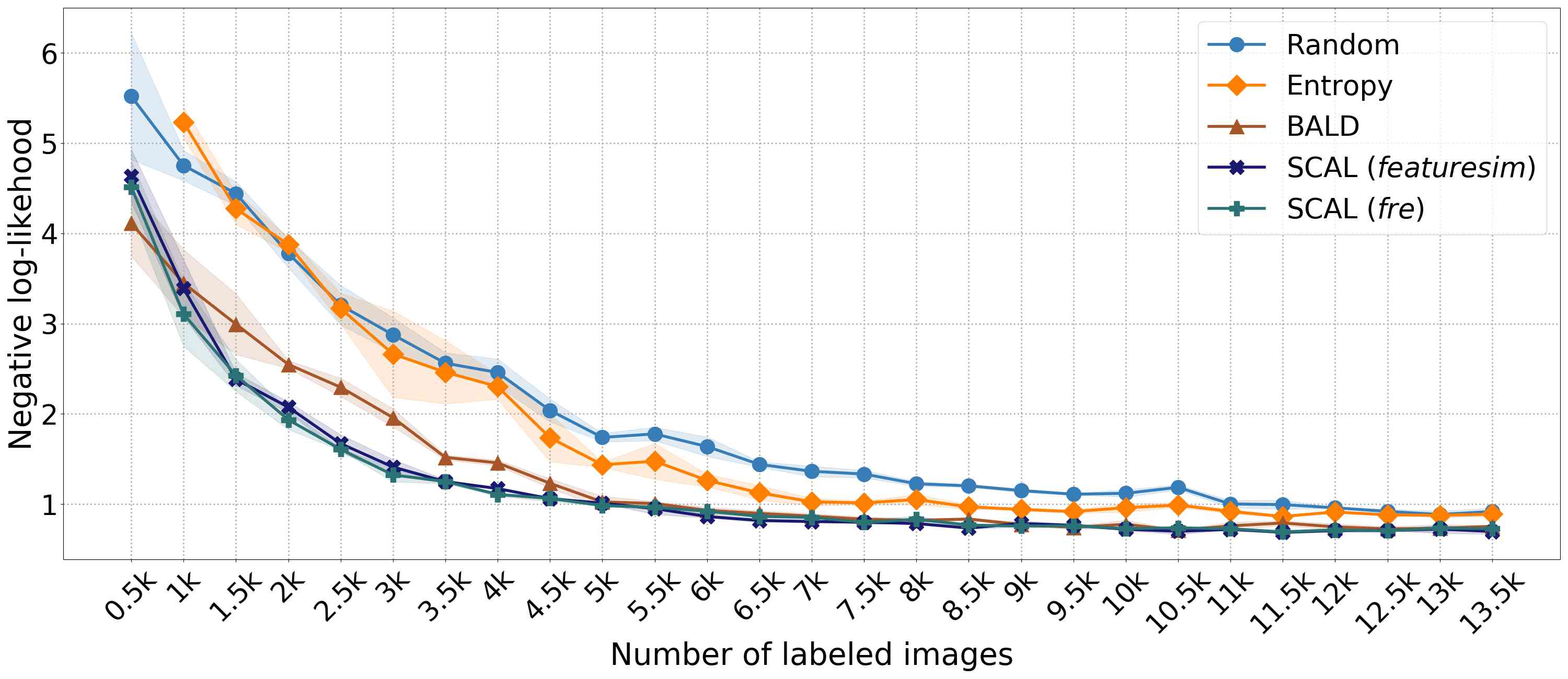}} 
\subfigure[Brier score ($\downarrow$)]{\includegraphics[width=0.7\textwidth]{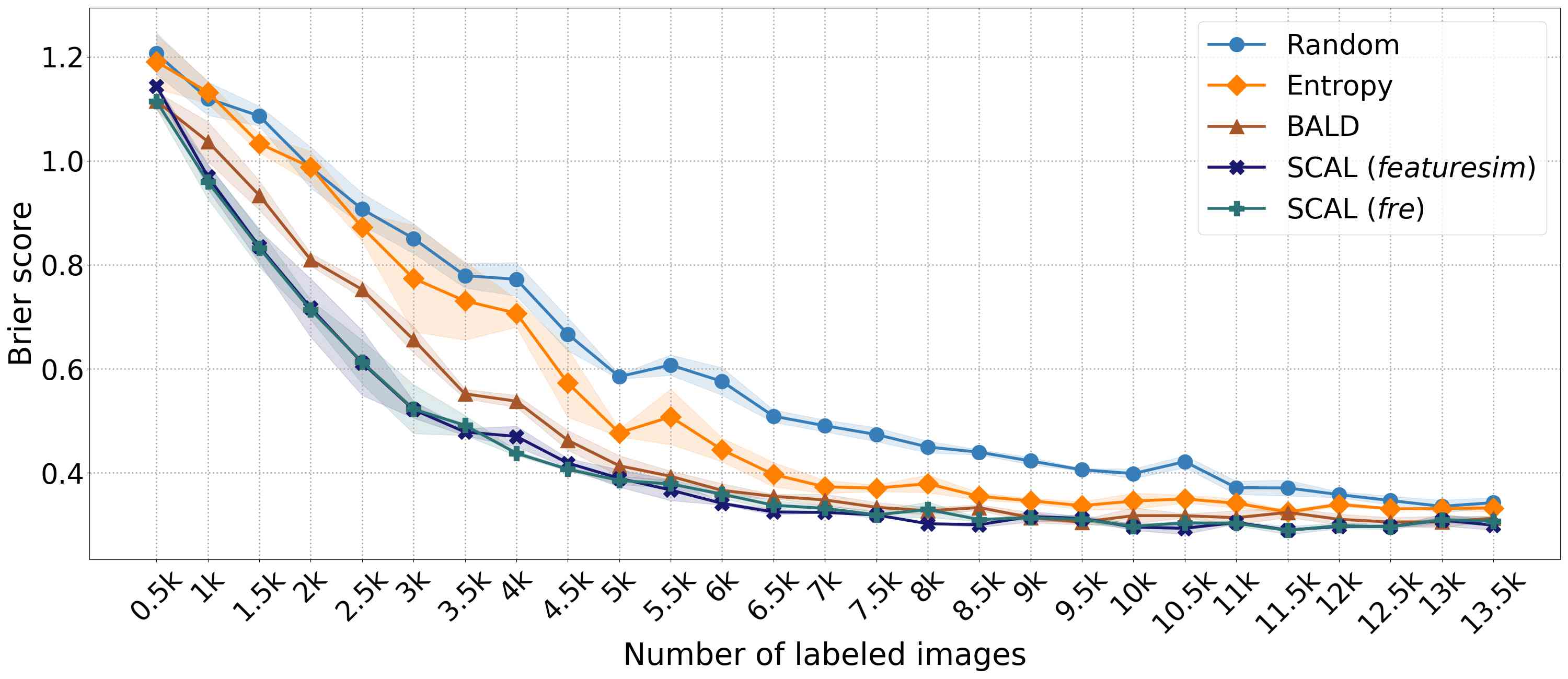}}  
\caption{Full learning curve for Imbalanced-CIFAR10.}
\label{fig:ablation_full_learning_curve}
\end{figure*}

We experiment repeating the active learning iterations until all the data samples in the entire unlabeled dataset are annotated. The full learning curve on Imabalanced-CIFAR10 is shown in Fig. \ref{fig:ablation_full_learning_curve}. SCAL method achieve upper bound accuracy with only 44\% of the total data samples, while Entropy and BALD require 70\% and 66\% of total data respectively.

Also, to achieve maximum upper bound accuracy SCAL required 48\% lesser labeled samples as compared to random selection and 27.8\% lesser labeled samples compared to BALD, requiring an oracle to label much fewer data samples to be annotated to achieve same accuracy.


\begin{table*}[!htp]	
	\caption{Test accuracy (Test Acc) for training with different loss functions using the entire dataset. This table indicates an upper bound on the achievable accuracy. } 
	\label{tab:metrics}
	\vskip 0.15in
	\begin{center}
		\begin{small}
			\begin{sc}
				\begin{tabular}{llcc}
					\toprule
					Dataset & Loss Function & \# of Train Samp & Test Acc (\%)  \\
					\midrule
\multirow{3}{*}{Imbalanced-CIFAR10} & CrossEntropy &   13996 & 79.53  \\
                              & CrossEntropy+Dropout  & 13996 & 78.51 \\
                              & Contrastive  & 13996 & 79.36  \\\hline
\multirow{3}{*}{CIFAR10} & CrossEntropy  & 50000 & 93.04 \\
                         & CrossEntropy+Dropout  & 50000 & 92.23  \\
                         & Contrastive  & 50000 & 92.69  \\\hline
\multirow{3}{*}{SVHN}    & CrossEntropy  & 73257 & 95.93  \\
                         & CrossEntropy+Dropout   & 73257 & 95.54  \\
                         & Contrastive  & 73257 & 96.06  \\\hline

					\bottomrule
				\end{tabular}
			\end{sc}
		\end{small}
	\end{center}
	\vskip -0.1in
	\label{tab:totscores}
\end{table*}

\end{document}